\pdfoutput=1

\documentclass[11pt]{article}

\usepackage{main}

\usepackage{times}
\usepackage{latexsym}
\usepackage{graphicx}
\usepackage{subcaption}
\usepackage{caption}
\usepackage{multirow}
\usepackage{array}
\usepackage{todonotes}
\usepackage{diagbox}
\usepackage{bm}
\usepackage{bm,bbm}
\newcommand{\bx}{\bm{x}}
\newcommand{\by}{\bm{y}}
\definecolor{SoftGreen}{HTML}{39ac73}
\usepackage{enumitem}
\usepackage[T1]{fontenc}

\usepackage[utf8]{inputenc}

\usepackage{microtype}

\title{Task Formulation Matters When Learning Continually: \\A Case Study in Visual Question Answering}

\author{Mavina Nikandrou$^{1}$, Lu Yu$^{2}$\thanks{\quad Work done at Heriot-Watt University.}, Alessandro Suglia$^{1}$, Ioannis Konstas$^{1}$, Verena Rieser$^{1}$ \\
  $^{1}$ Heriot-Watt University, Edinburgh, United Kingdom \\
  $^{2}$ Tianjin University of Technology, Tian jin, China\\
  {\tt\small \{mn2002,A.Suglia,I.Konstas,V.T.Rieser\}@hw.ac.uk,luyu@email.tjut.edu.cn}
  }

\date{}

\begin{document}
\maketitle
\begin{abstract}
Continual learning aims to train a model incrementally on a sequence of tasks without forgetting previous knowledge. 
Although continual learning has been widely studied in computer vision, its application to Vision+Language tasks is not that straightforward, as settings can be parameterized in multiple ways according to their input modalities. 
In this paper, we present a detailed study of how different settings affect performance for Visual Question Answering. 
We first propose three plausible task formulations and demonstrate their impact on the performance of continual learning algorithms.
We break down several factors of task similarity, showing that performance and sensitivity to task order highly depend on the shift of the output distribution.
We also investigate the potential of pretrained models and compare the robustness of transformer models with different visual embeddings. 
Finally, we provide an analysis interpreting model representations and their impact on forgetting. Our results highlight the importance of stabilizing visual representations in deeper layers.
\end{abstract}

\section{Introduction}\label{sec:intro}
The current paradigm to approach Vision+Language (V+L) tasks is to pretrain large-scale models, which are then finetuned and evaluated on independent and identically distributed (i.i.d.) data.
In practice, the i.i.d. assumption does not hold:
New data becomes available over time, which often results in a shift in data distribution.
One solution is to continuously adapt an existing model via finetuning. However, this will lead to catastrophic forgetting, i.e.\ significant performance degradation on previous data~\cite{mccloskey1989catastrophic,ratcliff1990connectionist}.
Continual learning provides a counterpart to i.i.d. learning: It defines a class of algorithms aiming at incremental learning with minimal forgetting.
This line of work becomes increasingly relevant given the financial and environmental costs of (re-)training large models~\cite{strubell-etal-2019-energy,bender2021dangers}, and the limited generalization of static models~\cite{lazaridou2021pitfalls}.

\begin{figure}
    \centering
    \includegraphics[width=\columnwidth]{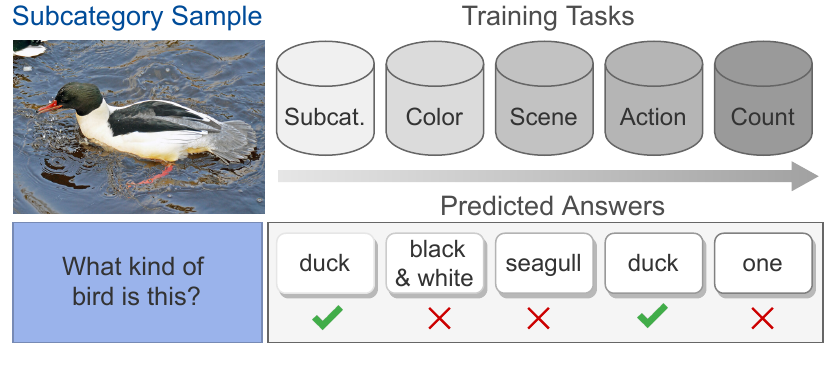}
    \caption{Predicted answers as the model continuously learns a sequence of tasks corresponding to different question types. Catastrophic forgetting causes incorrect predictions for preceding tasks.
    }
    \label{fig:exmaple_intro}
\end{figure}
 
While continual learning is widely studied in the computer vision community, its use within  V+L problems remains under-explored.
One challenge for applying continual learning to tasks like Visual Question Answering (VQA) is the lack of an agreed task definition:  
Computer vision tasks, like image classification, often fit in ``clear-cut" task, class, or domain incremental settings~\cite{van2019three}. This simplification is not always suitable. 
First, it rarely holds in real-world scenarios~\cite{mi2020generalized}. Second, it is unsuitable for V+L tasks, which can be parameterized in various ways according to their different modalities.
For example, task definitions for VQA can either be based on the language reasoning skills (as defined by the question type, cf.\ Figure~\ref{fig:exmaple_intro}) or the visual concepts in the images~\cite{whitehead2021separating}. Each of these perspectives reflects a different real-world requirement: New data might be collected with the intention of expanding the question types or domains to which a VQA system is applied. 
Similarly, output spaces are not exclusive. For example, counting questions are applicable to any visual domain, while binary questions can require different reasoning skills.

In this work, we provide an in-depth study of task design for Visual Question Answering (VQA) and its impact when combining pretrained V+L models with continual learning approaches. 
We introduce three continual learning settings based on the VQA-v2 dataset~\cite{vqav2}.  
Across these settings, we evaluate several regularization and memory-based continual learning methods.
Our results confirm that algorithmic performance is highly dependent on task design, order, and similarity, which is in-line with findings for image classification \cite{van2019three, Yoon2020Scalable, Delangesurvey}.
We also investigate the potential of pretrained models and their ability to generalize to unseen tasks in the CL setting. Our results show that although pretrained representations are more robust than when learning from scratch, they are still subject to catastrophic forgetting.

In addition, we perform a detailed analysis that relates the amount of forgetting to task similarity as measured by input embeddings and output distribution.
We find that incremental learning of new question types is the most challenging setting, as it shows a high divergence in answer distribution.
Figure~\ref{fig:exmaple_intro} provides an example where, given the question ``What kind of bird is this?", the last model in the CL task sequence predicts the incoherent answer ``one".
To measure more nuanced forgetting, we propose a novel evaluation metric based on semantic similarity. In the example in Figure~\ref{fig:exmaple_intro}, changing the answer from ``duck" to ``seagull" is penalized less.

Finally, we compare two transformer-based models, which use different visual representations.
We find that region features extracted from a fixed object detection model outperform representations based on image pixels. 
We track how representations from each modality change per layer, showing that visual representations from deeper layers are affected more prominently compared to language representations.

\section{Related Work}
To the best of our knowledge, this is the first work studying the impact of task formulation for continual learning in V+L models.
The vast majority of continual learning studies have focused on image classification settings.
For example, previous work has examined the relationship between catastrophic forgetting and different learning hyper-parameters, such as the activation function, dropout, and learning rate schedule~\cite{goodfellow2013empirical, mirzadeh2020understanding}.
Other work has highlighted the important
role of task similarity~\cite{ramasesh2020anatomy,lee2021continual} and which properties of task sequences amplify forgetting~\cite{nguyen-2019-understanding}.

Continual learning settings are typically categorized as task, class, or domain incremental~\cite{van2019three}.
In task and class-incremental settings, new classes are introduced over time, with the difference that task-incremental settings assume knowledge of the task identity during inference.
In domain-incremental learning, tasks differ in terms of their input distributions while sharing the same output space.

Previous work on V+L continual learning has studied these settings.
For example, \citet{del2020ratt} and~\citet{nguyen2019contcap} study continual learning for domain- and class-incremental image captioning,
while \citet{jin2020viscol} propose a more flexible setting of ``soft task boundaries" for the masked phrase prediction.
More recently, \citet{srinivasan2022climb} released a benchmark that combines V+L task-incremental learning with multimodal and unimodal transfer.
In contrast to these works, we examine the impact of task specification for V+L on performance and forgetting.

More closely related to our work,~\citet{greco2019psycholinguistics} explore the effect of forgetting in VQA with two question types (`Wh-' and binary questions). 
Consistent with our findings, they show that task order influences forgetting and that continual learning methods can alleviate forgetting.
However, their study is limited to only two tasks and does not test the impact of pretrained models, which has shown potential to mitigate forgetting~\cite{mehta2021empirical}.

\section{Settings for Continual VQA} \label{sec:settings}
\subsection{Problem formulation}
In continual learning, model parameters $\bm{\theta}$ are incrementally updated as new data become available.
We assume that samples from tasks $t=1\dots T$ arrive sequentially as $D_t=\{\bx_i, \by_i\}_{i=1}^{N_t}$, where $N_t$ is the number of data for task $t$.
Following previous work, VQA is formulated as a multi-label classification problem with soft targets $\by_i$~\cite{Anderson2017up-down}.
Starting from parameters $\bm{\theta}_{t-1}$ of the previous model, the updated parameters $\bm{\theta}_{t}$ are obtained by training on the new data $D_t$. 
Some approaches also use a memory $M_t$ containing a subset of samples from previous tasks, e.g. $D_{1}, \dots, D_{t-1}$.
In our setup, all tasks share a common output head which is extended with new classes from each task.
This allows inference to be task-agnostic but creates a more challenging setting than multi-head learning where separate heads are learned for each task~\cite{hussain2021towards}.
At the end of the training sequence, the objective is to achieve strong performance across all tasks observed so far. 
This objective encloses two challenges: 1) minimizing catastrophic forgetting of tasks seen earlier in training, 2) facilitating positive transfer to improve performance on new tasks~\cite{HADSELL20201028embracing}. 

\subsection{Task settings}

We define three continual learning settings for VQA based on different task definitions $D_t$, as summarized in Table~\ref{tab:stats}.
Two of these settings are based on visual object categories, see Subsection \ref{sec:vis_splits} and one setting is motivated by language capabilities, see  Subsection \ref{sec:q_splits}.
Concurrent work~\cite{lei2022symbolic} has followed a similar definition of continual learning settings for VQA. However, our work focuses on understanding how differences in task definitions affect the difficulty of the continual learning problem.
We study this problem from the point of view of both the downstream performance as well as the quality of the learned representations. 
This is in line with work on holistic evaluation frameworks for grounded language learning~\cite{suglia-etal-2020-compguesswhat}.

\begin{table}[t]
    \centering
    \resizebox{\columnwidth}{!}{%
    \begin{tabular}{cccccc} \hline
    {\bf Setting} & {\bf Task} & {\bf Train} & {\bf Val} & {\bf Test} & {\bf Classes} \\ \hline
    \multirow{5}{1em}{\rotatebox[origin=c]{90}{Diverse}}  
& Group 1 & 44254 &   11148    & 28315 &    2259  \\
& Group 2 & 39867 &   10202    & 22713 &    1929  \\
& Group 3 & 37477 &    9386    & 23095 &    1897  \\
& Group 4 & 35264 &    8871    & 21157 &    2165  \\
& Group 5 & 24454 &    6028    & 14490 &    1837  \\\hline
    \multirow{5}{1em}{\rotatebox[origin=c]{90}{Taxonomy}} 
&  Animals  & 37270 &    9237    & 22588 &    1378 \\
&   Food    & 26191 &    6612    & 15967 &    1419 \\
& Interior  & 43576 &   11038    & 26594 &    2143 \\
&  Sports   & 32885 &    8468    & 19205 &    1510 \\
& Transport & 41394 &   10280    & 25416 &    2009 \\\hline
    \multirow{5}{1em}{\rotatebox[origin=c]{90}{Question}} 
 &   Action     & 18730  &    4700     & 11008  &     233     \\
 &    Color     & 34588  &    8578     & 21559  &     92      \\
 &    Count     & 38857  &    9649     & 23261  &     42      \\
 &    Scene     & 25850  &    6417     & 14847  &     170     \\
 & Subcategory  & 22324  &    5419     & 13564  &     659     \\\hline
    \end{tabular}}
    \caption{Statistics per task within each setting.} 
    \label{tab:stats}
\end{table}
\subsubsection{Visual Settings} \label{sec:vis_splits}
We design two settings based on visual object categories, which correspond to expanding the domain on which the VQA system is applied to. We take advantage of the fact that images in the VQA-v2 dataset
originate from the COCO dataset~\cite{lin2014coco} which provides object-level image annotations.
Following previous work in image captioning~\cite{del2020ratt}, we organize $50$ object categories into five groups.
Images with objects from multiple groups are discarded in order to create clean task splits $D_t$ --
resulting in a total of 181K train, 45K validation, and 110K test samples. 

For the first setting, {\em Diverse Domains}, tasks are defined by grouping the object categories randomly.
Each task is assigned a balanced count of $10$ distinct objects resulting in five tasks. 
This type of setting corresponds to common practice of continual learning research within computer vision~\cite{rebuffi2017icarl,lomonaco2017core50}, and reflects a real-world scenario where sequential data do not necessarily follow a taxonomy.

The second setting, {\em Taxonomy Domains} groups objects based on their common super-category as in~\cite{del2020ratt}.
This results in five tasks: Animals, Food, Interior, Sports, and Transport. 
Note that the number of object classes per task under this definition is unbalanced since splits depend on the size of the super-category.
More details on each task can be found in Appendix~\ref{appendix:dataset}.

\subsubsection{Language Setting}\label{sec:q_splits}
We create a third setting {\em Question Types}, where each task corresponds to learning to answer a different category of questions.
We use a classification scheme developed by~\citet{whitehead2021separating}
to form a sequence of five tasks: 
Count, Color, Scene-level, Subcategory, and Action recognition. The splits for Count, Color, and Subcategory questions are obtained from~\citet{whitehead2021separating}. We create two additional tasks from the remaining questions.
In particular, we cluster question embeddings from Sentence-BERT~\cite{reimers-2019-sentence-bert} \footnote{We use the `all-MiniLM-L6-v2' model and Fast Clustering algorithm from the sentence-transformers package (\url{https://www.sbert.net/}).} so that each cluster has at least 15 questions and a minimum cosine similarity of 0.8 between all embeddings.
We annotate clusters as `scene', `action' or `irrelevant' question types.
Based on a seed of 10K annotated questions, we retrieve all other questions with similarity above 0.8 and label them using the K-nearest neighbor algorithm ($K=5$).  
Question Types have a total of 140K train, 35K validation and 84K test samples (cf.\ Table~\ref{tab:stats}). 
Common question words and answers per task are presented in the Appendix (Figure~\ref{fig:top_ans_q}).

\section{Experimental Framework}
\subsection{Models}\label{ssec:models}
In our experiments, we use two single-stream transformer models, UNITER-base~\cite{chen2020uniter} and  ViLT-base~\cite{pmlr-v139-kim21k-vilt} that differ in terms of how images are embedded at the input level.
UNITER relies on region features extracted from a frozen pretrained object detector, while ViLT 
directly embeds image patches.
Both models are pretrained on the same data that include among others in-domain images for VQA-v2, i.e.\ COCO captions~\cite{lin2014coco}.

\subsection{Continual Learning Methods}
We benchmark common continual learning algorithms, including regularization- and replay-based approaches. 
We investigate two regularization-based approaches:
{\em Learning without Forgetting} (LwF)~\cite{li2017lwf}, which uses knowledge distillation~\cite{hinton2015distilling} in order to retain knowledge from previous tasks, and  
{\em Elastic Weight Consolidation} (EWC)~\cite{kirkpatrick2017elastic}. The EWC regularization term discourages big changes of parameters that were important for previous tasks, where importance is approximated using the Fisher information matrix.

We apply three types of replay approaches that allow access to a memory of past samples.
{\em Experience Replay} (ER)~\cite{Chaudhry2019er} is the most straightforward approach, as it samples training data from both the current task and memory at each training step.
{\em Average Gradient Episodic Memory} (A-GEM)~\cite{lopez2017GEM, chaudhry2018agem} utilizes the memory of past data to ensure that gradient updates on past and new data are aligned. 

We also experiment with a baseline Pseudo-Replay method for the Question Types setting.
Instead of storing raw data from previous tasks, we use a data augmentation method, inspired by~\cite{kafle2017data,kil2021discovering}. 
When training on task $t$, we augment the data $D_t$ by retrieving past questions based on their shared detected objects classes. For example, if an elephant is detected on the current picture, we retrieve a past question about an elephant. We then use the previous model $f_{\theta_{t-1}}$ to generate a distribution $\tilde{ \by} = f_{\theta_{t-1}}(\tilde{\bx})$ which serves as soft targets for the new sample $\tilde{\bx}$.
By not storing the original answers, we address privacy and efficiency concerns of replay approaches~\cite{van2018generative,Delangesurvey}.

\subsection{Evaluation Metrics}
After training on task $t$, we compute the VQA accuracy $A_{t,i}$ on data from the previous task $i$. 
We report the macro-average accuracy at the end of the training sequence: $\mathrm{A} = \frac{1}{T} \sum_{i=1}^T A_{T,i}$.
Following~\citet{riemer2018learning}, we report the learned accuracy $\mathrm{LA} = \frac{1}{T} \sum_{i=1}^T A_{i,i}$, which measures the ability to learn the new task $i$.
We also compute backward transfer $\mathrm{BWT} = \frac{1}{T-1} \sum_{i=1}^{T-1} A_{T,i} - A_{i,i}$~\cite{lopez2017GEM}, that captures the impact of catastrophic forgetting.

In addition, we introduce a new metric, we term {\em semantic backward transfer} (SBWT), that weights backward transfer with the cosine distance of the predicted answer embeddings.
The motivation for this metric is simply that some incorrect answers are worse than others.
Consider the example in Figure~\ref{fig:exmaple_intro}, where the ground truth is `duck'. After training on subsequent tasks, the sample gets misclassified as `seagull' which might have a milder impact on the downstream application than completely unsuited answers such as `black and white' or `one'.
More detailed examples are provided in the Appendix Table~\ref{tab:sbwt_examples}.

For each sample $j=1\dots, N$ of task $i$, we measure the accuracy difference $\Delta^{Ti}_j$ of the answers predicted by the $T$-th and $i$-th models and
weigh it by cosine distance of the two answer embeddings $\bm{e_{Tj}}$ and $\bm{e_{ij}}$.
The final SBWT is computed as  :
 \begin{equation}
    \mathrm{SBWT} = \frac{1}{T-1} \sum_{i=1}^{T-1} S_{T,i}
 \end{equation}
 where $S_{T,i}$ is the average weighted accuracy difference for task $i$:
\begin{equation}
   S_{T,i} = \frac{1}{N} \sum_{j=1}^{N} (1-\mathrm{cos}(\bm{e_{Tj}}, \bm{e_{ij}}))\cdot \Delta^{Ti}_j
\end{equation}

In our implementation, we use averaged 300-dimensional GloVE embeddings~\cite{pennington2014glove}, since most answers are single words.

\begin{table*}[h!]
\centering
\small
\resizebox{\textwidth}{!}{%
\begin{tabular}{clcccc|cccc}
\hline
 & & & & & & & & & \\[-8pt]
& & \multicolumn{4}{c}{\textbf{w/o Pretraining}} & \multicolumn{4}{c}{\textbf{w/ Pretraining}}\\
\hline 
 & & & & & & & & & \\[-7pt]
\textbf{Split} & \textbf{Method} & \textbf{Accuracy} & \textbf{LA} & \textbf{BWT} & \textbf{SBWT} &\textbf{Accuracy} & \textbf{LA} & \textbf{BWT} & \textbf{SBWT} \\ \hline
 & & & & & & & & & \\[-7pt]
\multirow{7}{1em}{\rotatebox[origin=c]{90}{Diverse}} & Fixed Model & 41.60 \tiny{\textcolor{gray}{$\pm$ 0.84}} & - & - & - & 57.38 \tiny{\textcolor{gray}{$\pm$ 0.83}}
 & - & - & - \\
& Finetuning & 49.64 \tiny{\textcolor{gray}{$\pm$ 0.78}} & 56.69 \tiny{\textcolor{gray}{$\pm$ 0.28}} & -8.80 \tiny{\textcolor{gray}{$\pm$ 0.89}} & --5.35 \tiny{\textcolor{gray}{$\pm$ 0.61}} & 64.59 \tiny{\textcolor{gray}{$\pm$ 0.56}} & {\bf 67.77} \tiny{\textcolor{gray}{$\pm$ 0.22}} & -3.97 \tiny{\textcolor{gray}{$\pm$ 0.59}} & -1.93 \tiny{\textcolor{gray}{$\pm$ 0.39}} \\
& LwF & 50.70 \tiny{\textcolor{gray}{$\pm$ 0.56}} & 54.67 \tiny{\textcolor{gray}{$\pm$ 0.42}} & -4.96 \tiny{\textcolor{gray}{$\pm$ 0.29}} &  -2.89 \tiny{\textcolor{gray}{$\pm$ 0.17}} & 65.23 \tiny{\textcolor{gray}{$\pm$ 0.42}} & 67.62 \tiny{\textcolor{gray}{$\pm$ 0.25}} & -3.02 \tiny{\textcolor{gray}{$\pm$ 0.44}} & -1.50 \tiny{\textcolor{gray}{$\pm$ 0.28}}  \\
& AGEM & 51.56 \tiny{\textcolor{gray}{$\pm$ 0.78}} & {\bf 56.72} \tiny{\textcolor{gray}{$\pm$ 0.30}} & -6.45 \tiny{\textcolor{gray}{$\pm$ 0.87}} & -3.84 \tiny{\textcolor{gray}{$\pm$ 0.60}} &  65.65 \tiny{\textcolor{gray}{$\pm$ 0.85}} & 67.72 \tiny{\textcolor{gray}{$\pm$ 0.30}} & -2.60 \tiny{\textcolor{gray}{$\pm$ 0.71}} &  -1.22 \tiny{\textcolor{gray}{$\pm$ 0.38}} \\
& EWC & 52.05 \tiny{\textcolor{gray}{$\pm$ 0.30}} & 56.49 \tiny{\textcolor{gray}{$\pm$ 0.22}} & -5.55 \tiny{\textcolor{gray}{$\pm$ 0.60}} &  -3.12 \tiny{\textcolor{gray}{$\pm$ 0.40}}  & 66.26 \tiny{\textcolor{gray}{$\pm$ 0.55}} & 67.58 \tiny{\textcolor{gray}{$\pm$ 0.27}} & -1.65 \tiny{\textcolor{gray}{$\pm$ 0.45}} & -0.67 \tiny{\textcolor{gray}{$\pm$ 0.29}}  \\
& ER &  {\bf 54.36} \tiny{\textcolor{gray}{$\pm$ 0.33}} & 56.31 \tiny{\textcolor{gray}{$\pm$ 0.51}} & {\bf -2.45} \tiny{\textcolor{gray}{$\pm$ 0.49}} & {\bf -1.42} \tiny{\textcolor{gray}{$\pm$ 0.26}}  & {\bf 66.66} \tiny{\textcolor{gray}{$\pm$ 0.50}} & 67.55 \tiny{\textcolor{gray}{$\pm$ 0.23}} & {\bf -1.11} \tiny{\textcolor{gray}{$\pm$ 0.41}} & {\bf -0.51} \tiny{\textcolor{gray}{$\pm$ 0.27}}  \\
\rowcolor{gray!30}
& Joint &  60.41 \tiny{\textcolor{black}{$\pm$ 0.03}} & - & - & - &  69.76 \tiny{\textcolor{black}{$\pm$ 0.18}} & - & - & - \\\hline
 & & & & & & & & & \\[-7pt]
\multirow{7}{1em}{\rotatebox[origin=c]{90}{Taxonomy}} & Fixed Model & 39.96 \tiny{\textcolor{gray}{$\pm$ 1.05}} & - & - & - & 55.00 \tiny{\textcolor{gray}{$\pm$ 0.95}} & - & - & - \\
& Finetuning & 47.72 \tiny{\textcolor{gray}{$\pm$ 0.72}} & 57.75 \tiny{\textcolor{gray}{$\pm$ 0.24}} & -12.53 \tiny{\textcolor{gray}{$\pm$ 0.65}} &  -8.45 \tiny{\textcolor{gray}{$\pm$ 0.38}}  & 
 63.65 \tiny{\textcolor{gray}{$\pm$ 0.63}} & 68.77 \tiny{\textcolor{gray}{$\pm$ 0.12}} & -6.40 \tiny{\textcolor{gray}{$\pm$ 0.67}} & -3.89 \tiny{\textcolor{gray}{$\pm$ 0.53}} \\
& LwF & 48.05 \tiny{\textcolor{gray}{$\pm$ 0.24}} & 55.25 \tiny{\textcolor{gray}{$\pm$ 0.27}} & -9.00 \tiny{\textcolor{gray}{$\pm$ 0.38}} &  -6.13 \tiny{\textcolor{gray}{$\pm$ 0.44}}  & 64.83 \tiny{\textcolor{gray}{$\pm$ 0.50}} & 68.73 \tiny{\textcolor{gray}{$\pm$ 0.17}} & -4.88 \tiny{\textcolor{gray}{$\pm$ 0.69}} & -2.88 \tiny{\textcolor{gray}{$\pm$ 0.43}}\\
& AGEM & 50.51 \tiny{\textcolor{gray}{$\pm$ 0.66}} & {\bf 57.80} \tiny{\textcolor{gray}{$\pm$ 0.25}} & -9.10 \tiny{\textcolor{gray}{$\pm$ 0.79}} & -5.77 \tiny{\textcolor{gray}{$\pm$ 0.55}}  & 66.52 \tiny{\textcolor{gray}{$\pm$ 0.34}} & {\bf 68.86} \tiny{\textcolor{gray}{$\pm$ 0.12}} & -2.92 \tiny{\textcolor{gray}{$\pm$ 0.50}} &  -1.63 \tiny{\textcolor{gray}{$\pm$ 0.33}} \\
& EWC & 52.17 \tiny{\textcolor{gray}{$\pm$ 0.54}} & 57.49 \tiny{\textcolor{gray}{$\pm$ 0.19}} & -6.65 \tiny{\textcolor{gray}{$\pm$ 0.44}} & -4.33 \tiny{\textcolor{gray}{$\pm$ 0.28}} & {\bf 67.70} \tiny{\textcolor{gray}{$\pm$ 0.29}} & 68.57 \tiny{\textcolor{gray}{$\pm$ 0.16}} & {\bf -1.09} \tiny{\textcolor{gray}{$\pm$ 0.33}} & {\bf -0.62} \tiny{\textcolor{gray}{$\pm$ 0.19}}\\
& ER & {\bf 54.60} \tiny{\textcolor{gray}{$\pm$ 0.14}} & 57.67 \tiny{\textcolor{gray}{$\pm$ 0.28}} & {\bf -3.84} \tiny{\textcolor{gray}{$\pm$ 0.42}} & {\bf -2.38} \tiny{\textcolor{gray}{$\pm$ 0.27}}  & 66.76 \tiny{\textcolor{gray}{$\pm$ 0.16}} & 68.61 \tiny{\textcolor{gray}{$\pm$ 0.13}} & -2.32 \tiny{\textcolor{gray}{$\pm$ 0.16}} & -1.22 \tiny{\textcolor{gray}{$\pm$ 0.10}}\\
\rowcolor{gray!30}
& Joint & 60.82 \tiny{\textcolor{black}{$\pm$ 0.02}} & - & - & - & 70.08 \tiny{\textcolor{black}{$\pm$ 0.18}} & - & - & - \\\hline
 & & & & & & & & & \\[-7pt]
\multirow{7}{1em}{\rotatebox[origin=c]{90}{Questions}}  & Fixed Model & 18.81 \tiny{\textcolor{gray}{$\pm$ 5.90}} & - & - & - & 25.54 \tiny{\textcolor{gray}{$\pm$ 8.75}} & - & - & - \\
& Finetuning &  23.30 \tiny{\textcolor{gray}{$\pm$ 8.83}} & 65.24 \tiny{\textcolor{gray}{$\pm$ 0.42}} & -52.42 \tiny{\textcolor{gray}{$\pm$ 10.88}} & -39.86 \tiny{\textcolor{gray}{$\pm$ 12.08}} & 48.81 \tiny{\textcolor{gray}{$\pm$ 5.56}} & 72.94 \tiny{\textcolor{gray}{$\pm$ 0.20}} & -30.17 \tiny{\textcolor{gray}{$\pm$ 7.07}} & -22.43 \tiny{\textcolor{gray}{$\pm$ 7.02}} \\
& LwF & 26.23 \tiny{\textcolor{gray}{$\pm$ 8.56}} & 60.69 \tiny{\textcolor{gray}{$\pm$ 1.43}} & -43.08 \tiny{\textcolor{gray}{$\pm$ 11.22}} & -34.32 \tiny{\textcolor{gray}{$\pm$ 9.94}} &  46.61 \tiny{\textcolor{gray}{$\pm$ 3.95}} & 72.06 \tiny{\textcolor{gray}{$\pm$ 0.44}} & -31.82 \tiny{\textcolor{gray}{$\pm$ 5.42}} & -25.13 \tiny{\textcolor{gray}{$\pm$ 5.35}} \\
& AGEM & 50.73 \tiny{\textcolor{gray}{$\pm$ 1.92}} & {\bf 65.38} \tiny{\textcolor{gray}{$\pm$ 0.56}} & -18.31 \tiny{\textcolor{gray}{$\pm$ 3.04}} & -10.02 \tiny{\textcolor{gray}{$\pm$ 1.39}} & 68.30 \tiny{\textcolor{gray}{$\pm$ 0.74}} & 72.96 \tiny{\textcolor{gray}{$\pm$ 0.24}} & -5.83 \tiny{\textcolor{gray}{$\pm$ 1.08}} & -2.95 \tiny{\textcolor{gray}{$\pm$ 0.63}} \\
& EWC & 36.77 \tiny{\textcolor{gray}{$\pm$ 5.01}} & 49.05 \tiny{\textcolor{gray}{$\pm$ 3.82}} & -15.35 \tiny{\textcolor{gray}{$\pm$ 5.85}} & -11.76 \tiny{\textcolor{gray}{$\pm$ 5.41}} & 66.77 \tiny{\textcolor{gray}{$\pm$ 3.54}} & 70.03 \tiny{\textcolor{gray}{$\pm$ 1.03}} &  {\bf -4.08} \tiny{\textcolor{gray}{$\pm$ 3.58}} & -2.62 \tiny{\textcolor{gray}{$\pm$ 2.28}} \\
& ER  & {\bf 59.54} \tiny{\textcolor{gray}{$\pm$ 0.32}} & 65.09 \tiny{\textcolor{gray}{$\pm$ 0.52}} & {\bf -6.93} \tiny{\textcolor{gray}{$\pm$ 0.71}} & {\bf -3.50} \tiny{\textcolor{gray}{$\pm$ 0.35}} & {\bf 69.18} \tiny{\textcolor{gray}{$\pm$ 0.38}} & 72.82 \tiny{\textcolor{gray}{$\pm$ 0.22}} & -4.56 \tiny{\textcolor{gray}{$\pm$ 0.56}} &  {\bf -1.82} \tiny{\textcolor{gray}{$\pm$ 0.34}} \\
\rowcolor{gray!30}
& Joint &  66.35 \tiny{\textcolor{black}{$\pm$ 0.24}} & - & - & - & 72.54 \tiny{\textcolor{black}{$\pm$ 0.15}}  & - & - & -\\
\hline
\end{tabular}}
\caption{\label{vis1} Results from VQA Incremental
Learning. We report the average and standard deviation over five random task orders. LA: Learned Accuracy, BWT: Backward Transfer, SBWT: Semantic Backward Transfer.}
\label{tab:results}
\end{table*}

\subsection{Experimental Setup}
We follow a single head setting to allow for task-agnostic inference but assume knowledge of task boundaries during training.
Unless stated otherwise, memory-based approaches store 500 randomly selected samples per past task.
For further implementation details, please refer to Appendix~\ref{sec:implementation_details}.

We consider two baselines:
The \textit{Fix Model} baseline represents the generalization ability of the model across all tasks 
after being trained on only the first task $D_1$.
The vanilla \textit{Finetuning} baseline represents the performance degradation if no measures are taken to prevent forgetting.
We also report the performance of joint training on all the data simultaneously (\textit{Joint}) as an upper bound.

\section{Results}
\subsection{Main Results}

\paragraph{Task Settings.}
Table~\ref{tab:results} summarizes the results averaged over five task orders using the UNITER backbone.
The results show an increasing difficulty for the three incremental learning task definitions, i.e. Diversity Domains $<$ Taxonomy Domains $<$ Question Types, which we will further investigate in Section \ref{sec:task_relationships}. 

Although Question Types has the highest Joint accuracy, naive finetuning shows poor performance: it has the lowest final accuracy and large negative BWT.
The low Fixed Model accuracy suggests that tasks are dissimilar as a model trained on a single task fails to generalize.

\paragraph{Pretraining.}
Our results also confirm that pretraining leads to models that are more robust to forgetting~\cite{mehta2021empirical}: all metrics consistently improve starting from a pretrained model.
Pretraining combined with naive finetuning achieves on average 58$\%$ relative accuracy improvement over finetuning a model from scratch. 
Interestingly, the pretrained Fixed Model is able to generalize reasonably well to other domains for both image-based settings, and the final Pretraining+Finetuning accuracy exceeds the Joint accuracy without pretraining. 
These results indicate that learning generic V+L representations via pretraining has persistent benefits.
However, pretraining is insufficient for ensuring continual learning and additional strategies improve the final accuracy by 8.83\% on average. 

\paragraph{Continual Learning Methods.}
Among continual learning methods, LwF offers the smallest gains in terms of final accuracy and forgetting. 
For pretrained models in Question Types, it fails to improve the final accuracy.
This can be attributed to the pseudo-labels generated using the current data becoming too noisy when the answers for the current and previous tasks differ substantially.

Pretraining+EWC achieves the highest accuracy in the Taxonomy Domains. 
However, when dealing with heterogeneous tasks
(i.e.\ within Question Types) the high regularization weights, which are required to prevent forgetting, limit the model's ability to adapt to new tasks.
This is reflected in the low LA of EWC, indicating that the model struggles to learn new tasks.
On the other hand, memory-based approaches have consistently high LA.
AGEM performs reasonably well across settings, but is always outperformed by the straightforward ER, which shows the best performance with models trained from scratch and for the challenging setting of Question Types. 

\paragraph{Measuring Forgetting.}
We compare the SBWT metric, which takes semantic similarities into account, to the standard BWT, which measures absolute forgetting.
We observe some notable differences, which indicate that SBWT favors strong models that forget gradually.

For instance, EWC w/o pretraining shows lower performance and LA under the Question Types setting compared to, e.g.\ AGEM w/o pretraining. However, it receives a better BWT score. 
We make similar observations for LwF vs. AGEM in Taxonomy Domains w/o pretraining, and EWC vs. ER in Taxonomy Domains with pretraining.

Appendix \ref{tab:sbwt_examples} provides a qualitative analysis  with examples from the validation set.
In addition, Table  \ref{tab:sbwt_examples} in the Appendix provides an example-based analysis of our suggested metric, showing that semantically similar answers have higher SBWT scores.

\subsection{Experience Replay Ablation}
\begin{figure}[h]
    \includegraphics[width=0.45\textwidth]{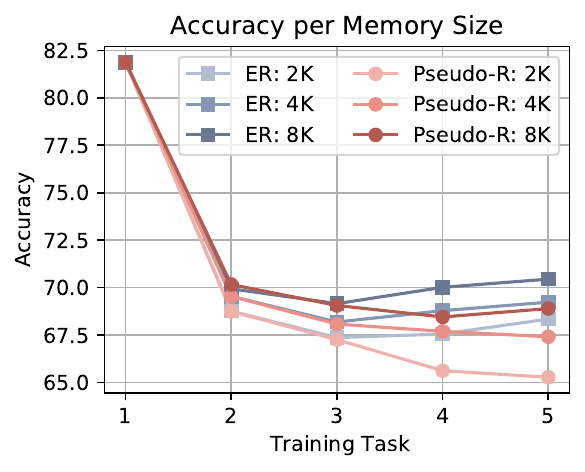}
    \caption{Average accuracy of seen tasks per memory size. Pseudo-Replay performs competitively up the the third task despite only storing questions.
    }
    \label{fig:mem_acc}
\end{figure}

The above strong performance of the straightforward replay methods suggests that more advanced strategies for selecting or generating samples representative of past tasks can yield further improvements.
One promising avenue is to make Experience Relay  more efficient. In general, more memory means less forgetting but at a higher computation and storage cost.
We experiment with a more efficient Pseudo-Replay  method which only stores past questions.
Figure~\ref{fig:mem_acc} shows the average accuracy  across training for three memory sizes. 
At each step, we compute the average accuracy of the experienced tasks up to that point.  
As expected, both methods benefit from access to a larger memory. 
Pseudo-Replay shows comparable performance for up to three tasks, while raw ER replay becomes more advantageous as more tasks are added. 
We attribute this convergence in performance to errors accumulated by pseudo-labeling~\cite{mean-teachers}.
Despite this limitation, Pseudo-Replay exceeds the performance of naive finetuning by $18\%$ when storing only 500 samples per task and without requiring access to any past images. 

\section{Task Similarity and Forgetting}\label{sec:task_relationships}

\subsection{Pairwise Task Characterization}
To gain further insight into which factors contribute to forgetting,
we measure the correlation between pairwise accuracy drop and task similarity. 
In the more widely studied task-incremental learning for image classification, task similarity refers to the semantic similarity of the old and new classes~\cite{ramasesh2020anatomy}. 
Here, we consider the similarity of the answer distributions, as well as the image, question and the joint pair representations.

\begin{table}[h]
    \centering
\resizebox{\columnwidth}{!}{%
\begin{tabular}{l|ccccc}
\multicolumn{6}{l}{{\bf Diverse Domains}} \\\hline
\diagbox[width=6em]{Task 1}{Task 2}
& Group 1 & Group 2 & Group 3 & Group 4 & Group 5 \\\hline
Group 1 & \cellcolor{gray!60} - & \cellcolor{red!5.47985781990521}  -6.58 & \cellcolor{red!4.344391785150075}  -5.21 & \cellcolor{SoftGreen!3.139810426540407}  -4.84 & \cellcolor{red!5.91182859399684}  -7.09 \\
Group 2 & \cellcolor{SoftGreen!9.010600706713689}  -4.55 & \cellcolor{gray!60} - & \cellcolor{red!4.674617196702005}  -5.61 & \cellcolor{SoftGreen!9.893992932862119}  -4.51 & \cellcolor{SoftGreen!0.17667844522966902}  -4.99 \\
Group 3 & \cellcolor{SoftGreen!7.1015106593253705}  -4.64 & \cellcolor{red!6.988564167725539}  -8.39 & \cellcolor{gray!60} - & \cellcolor{red!6.141465480728503}  -7.37 & \cellcolor{red!9.718104381382652}  -11.66 \\
Group 4 & \cellcolor{SoftGreen!6.197139781639095}  -4.69 & \cellcolor{red!5.9203444564047425}  -7.10 & \cellcolor{red!6.163821825823982}  -7.40 & \cellcolor{gray!60} - & \cellcolor{red!8.02193859244451}  -9.63 \\
Group 5 & \cellcolor{SoftGreen!14.294750158127727}  -4.29 & \cellcolor{red!4.849251528568416}  -5.82 & \cellcolor{red!5.073265865485981}  -6.09 & \cellcolor{SoftGreen!24.09867172675527}  -3.80 & \cellcolor{gray!60} - \\\hline
\multicolumn{6}{l}{}\\[2pt]

\multicolumn{6}{l}{{\bf Taxonomy Domains}} \\\hline
\diagbox[width=6em]{Task 1}{Task 2} 
& Animals & Food & Interior & Sports & Transport \\\hline
Animals & \cellcolor{gray!60} - & \cellcolor{red!6.719879929048997}  -8.06 & \cellcolor{SoftGreen!27.41165234001882}  -3.63 & \cellcolor{red!4.866512029835813}  -5.84 & \cellcolor{SoftGreen!12.948560513030117}  -4.35 \\
Food & \cellcolor{red!13.65079365079365}  -16.38 & \cellcolor{gray!60} - & \cellcolor{SoftGreen!14.285714285714192}  -4.29 & \cellcolor{red!14.232804232804229}  -17.08 & \cellcolor{red!9.947089947089951}  -11.94 \\
Interior & \cellcolor{red!4.788538155072028}  -5.75 & \cellcolor{red!4.328838492185116}  -5.19 & \cellcolor{gray!60} - & \cellcolor{red!6.3591786699356465}  -7.63 & \cellcolor{SoftGreen!43.30370824394703}  -2.83 \\
Sports & \cellcolor{red!9.689658306070521}  -11.63 & \cellcolor{red!15.164940021810253}  -18.20 & \cellcolor{red!7.997091966557623}  -9.60 & \cellcolor{gray!60} - & \cellcolor{red!7.894856415848786}  -9.47 \\
Transport & \cellcolor{SoftGreen!16.27906976744177}  -4.19 & \cellcolor{red!7.067183462532299}  -8.48 & \cellcolor{SoftGreen!47.596899224806265}  -2.62 & \cellcolor{SoftGreen!26.511627906976827}  -3.67 & \cellcolor{gray!60} - \\\hline
\multicolumn{6}{l}{}\\[2pt]
    
\multicolumn{6}{l}{{\bf Question Types}} \\\hline
\diagbox[width=6em]{Task 1}{Task 2}
& Action & Color & Count & Scene & Subcat. \\\hline
Action & \cellcolor{gray!60} - & \cellcolor{red!57.00123915737299}  -68.40 & \cellcolor{red!75.37495192923984}  -90.45 & \cellcolor{red!16.32269367175149}  -19.59 & \cellcolor{red!10.479425714651969}  -12.58 \\
Color & \cellcolor{red!74.07521705139284}  -88.89 & \cellcolor{gray!60} - & \cellcolor{red!83.04530304900631}  -99.65 & \cellcolor{red!23.12471711311361}  -27.75 & \cellcolor{red!52.05118709624326}  -62.46 \\
Count & \cellcolor{red!82.64429312581063}  -99.17 & \cellcolor{red!83.063121487246}  -99.68 & \cellcolor{gray!60} - & \cellcolor{red!81.26621271076525}  -97.52 & \cellcolor{red!72.4978383052313}  -87.00 \\
Scene & \cellcolor{red!9.091433848995615}  -10.91 & \cellcolor{red!28.66928346032479}  -34.40 & \cellcolor{red!64.7752636034788}  -77.73 & \cellcolor{gray!60} - & \cellcolor{red!12.679904563996002}  -15.22 \\
Subcat. & \cellcolor{red!26.44189628615437}  -31.73 & \cellcolor{red!71.21056534885048}  -85.45 & \cellcolor{red!80.12436533744082}  -96.15 & \cellcolor{red!25.457812767414}  -30.55 & \cellcolor{gray!60} - \\\hline
\multicolumn{6}{l}{}\\[2pt]
\end{tabular}}
    \caption{Task difficulty measured by forgetting in pairwise tasks.
 Non-diagonal elements show relative accuracy drop (\%) after finetuning on Task 2.}
    \label{tab:Tpairs}
\end{table}

\paragraph{Experimental Setup.} 
We first look into pairwise task relationships following studies in transfer~\cite{zamir2018taskonomy} and multitask learning ~\cite{pmlr-v119-standley20a, lu202012}.
In particular, we measure the extent to which each task is forgotten after training on a second task. 
We finetune a pretrained model on Task $T_{1}$ and compute the accuracy $A_{11}$ on its test set. Then, we finetune this model on another Task $T_{2}$ and compute the new accuracy $A_{12}$ on the test set of $T_{1}$.
Forgetting is measured as the relative accuracy drop: $(A_{12} - A_{11}) / A_{11}$.
Given the varying dataset sizes, we finetune on $T_{2}$ for a fixed number of 400 steps using a batch size of 512 and learning rate 5e-5.

Next, we compute the Spearman correlation between the relative accuracy drops and different factors of task dissimilarity.
Here, we consider the answer distributions , as well as average embeddings of the image, question and the joint pair.
Consider $P$, $Q$  the answer distributions of Tasks $T_1,T_2$ respectively.
Since some answers of $T_1$ do not appear in $T_2$, we measure the skew divergence~\cite{pmlr-vR3-lee01a} between $P$ and $Q$ as the KL divergence between $P$ and a mixture distribution \mbox{$(1-\alpha) P + \alpha Q$} with $\alpha=0.99$~\cite{ruder-plank-2017-learning}.
For the input embeddings, we measure the cosine distance between the average task representation. 
As image representations, we utilize Faster R-CNN features from~\cite{Anderson2017up-down}, while questions are embedded using Sentence-BERT.
Joint embeddings for image-question pairs are obtained using the final layer representation of the [CLS] token of UNITER~\footnote{[CLS] is the first token of the input sequence which aggregates multimodal information. Its representation from the final encoder layer is passed to the classifier to predict an answer.}.
The detailed similarity measures are shown in the Appendix Figure~\ref{fig:dissimilarity}.

\paragraph{Results.} 
Table~\ref{tab:Tpairs} shows the relative accuracy drop for all task pairs. 
Overall, we observe that each setting has a distinct pattern. 
Question Types is evidently a more challenging setting, where several task combinations show more than $90\%$ drop.
When comparing the visual settings, forgetting in Diverse Domains fluctuates less depending on the task pairing. 
This suggests that the task relationships in Taxonomy Domains might play a more important factor.
Although some relations make sense based on the expected similarity of the visual scenes, e.g., low forgetting between Food and Interior, others are less intuitive, e.g., low forgetting between Transport and Interior.
Moreover, certain second tasks seem to consistently affect the amount of forgetting after finetuning on them. 
Based on the total number of classes per task as shown in Table ~\ref{tab:stats}, we notice that the model is more robust against forgetting when Task $T_2$ has a wide range of possible answers (e.g., Interior); while $T_2$ with a narrow answer set (e.g., Food, Color, Count) lead to maximum forgetting.

\begin{table}[]
    \centering
    \resizebox{\columnwidth}{!}{%

    \begin{tabular}{lcccccc}\hline
     \textbf{Dissimilarity} & \textbf{Diverse} & \textbf{Taxonomy} & \textbf{Questions}  \\
      \textbf{Factor} & \textbf{Domains} & \textbf{Domains} & \textbf{Types} \\\hline
        Answer distribution  & 0.567* & 0.791* & 0.795* \\
        Image embedding &  0.248 & 0.492* & -0.640* \\
        Question embedding & 0.184 & 0.531* & 0.631* \\
        Joint embedding & 0.220 &   0.622*  & -0.223 
        \\\hline
    \end{tabular}}
    \caption{Spearman correlation of pairwise performance drop and task dissimilarity (* where $p < 0.05$).} 
    \label{tab:spearman}
\end{table}

The correlation results in Table~\ref{tab:spearman} indicate that the more similar two consecutive tasks are, the less forgetting occurs. 
The divergence of answer distributions consistently correlates with forgetting, but does not fully account for the performance drop.
For example, the divergence of Interior from Animals and Sports answer distributions is the same, however Sports
leads to 1.88$\%$ more forgetting.
Regarding the embedding distances, image embeddings show the highest correlation in Taxonomy Domain, meaning that the more visually similar two domains are, the less severe forgetting is. 
We observe the same relationship mirrored in Question Types for question embeddings.
We find no factor to correlate significantly with Diverse Domains, where tasks are relatively similar to each other (cf.\ Appendix~\ref{fig:dissimilarity}).
Looking across modalities, question and joint similarities in Taxonomy Domains correlate with forgetting, showing that the shift of the visual domains results in changes of the referred objects and types of questions per task.\footnote{We notice that the more similar images of two Question Types tasks are, the more forgetting occurs. 
A possible explanation is that new questions for similar images `overwrite' previous knowledge. However, all cosine distances of image embeddings are too low (<0.05) to lead to any conclusions.}

\subsection{Sensitivity to Task Order} \label{sec:task_order}
\begin{figure}
    \centering
    \includegraphics[width=0.44\textwidth]{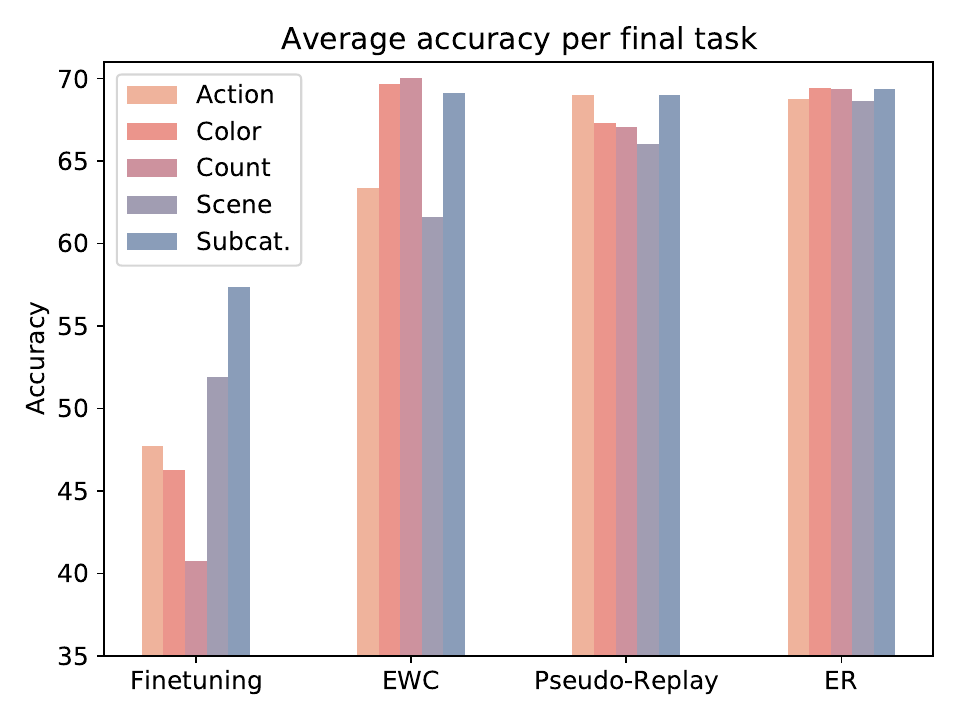}
    \caption{Sensitivity to task order as illustrated for Question Types. Each bar shows the accuracy of a task sequence ending with a different task.}
    \label{fig:qtask_order}
\end{figure}

\begin{figure*}
    \centering
         \includegraphics[width=0.8\textwidth]{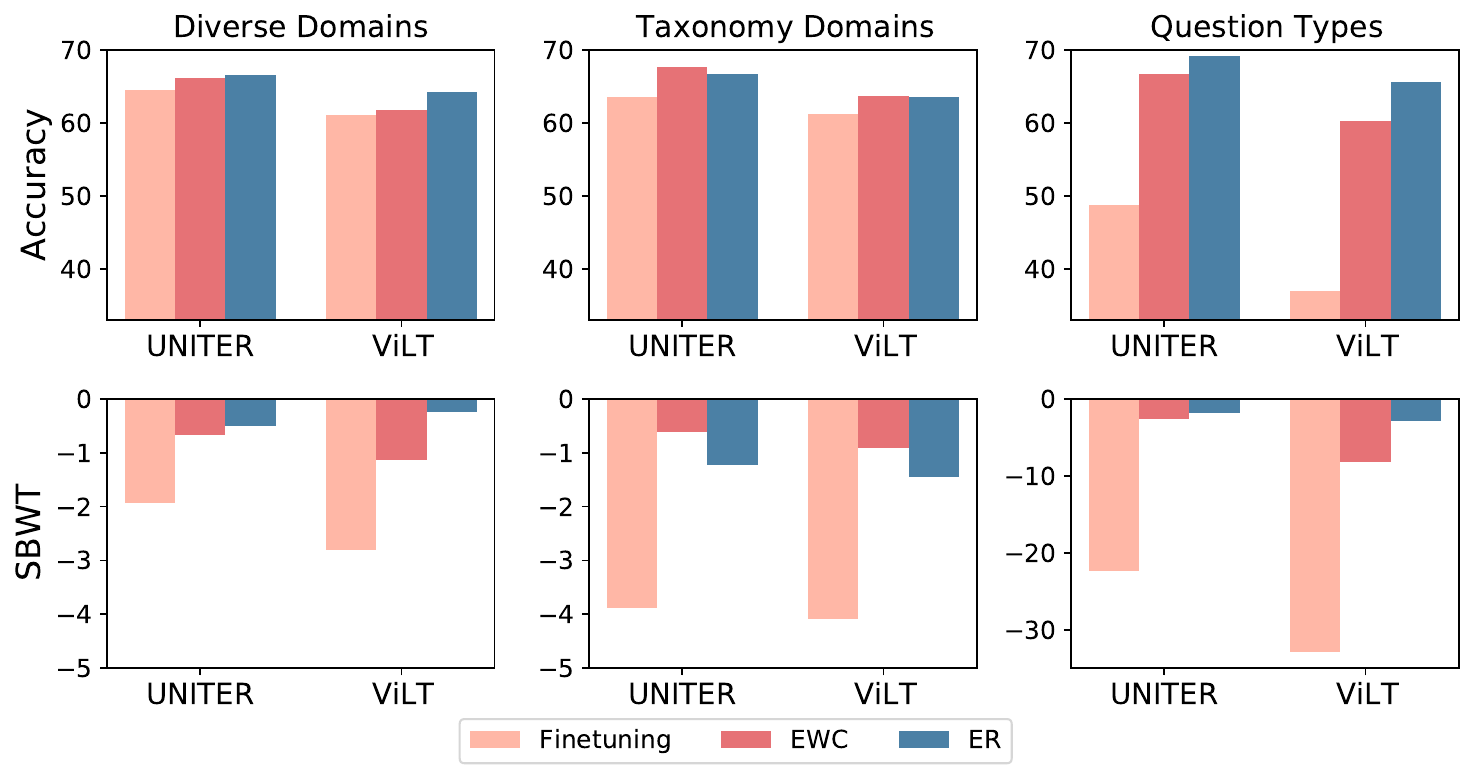}
    \caption{Comparison of Accuracy and Semantic Backward Transfer with UNITER vs ViLT.}
     \label{fig:uniter_vs_vilt}
\end{figure*}

\begin{table}
    \centering
    \resizebox{\columnwidth}{!}{%
    \begin{tabular}{lccc}\hline
    \multicolumn{4}{c}{{\bf w/o Pretraining}}  \\\hline
        \textbf{Method}  & \textbf{What animal} & \textbf{What room} & \textbf{What sport} \\\hline
        Finetuning &  33.09 \small{\textcolor{black}{$\pm$ 13.38}} & 54.38 \small{\textcolor{black}{$\pm$ 32.42}}  &  25.14 \small{\textcolor{black}{$\pm$ 32.11}} \\
        EWC & 48.18 \small{\textcolor{black}{$\pm$ 15.67}} & 83.48 \small{\textcolor{black}{$\pm$ 7.61}}  & 62.81 \small{\textcolor{black}{$\pm$ 13.67}} \\
        ER & 73.11 \small{\textcolor{black}{$\pm$ 0.70}} & 89.04 \small{\textcolor{black}{$\pm$ 2.80}}  & 87.20 \small{\textcolor{black}{$\pm$ 1.84}} \\\hline
    \multicolumn{4}{c}{{\bf w/ Pretraining}}  \\\hline
        \textbf{Method}  & \textbf{What animal} & \textbf{What room} & \textbf{What sport} \\\hline
        Finetuning &   75.07 \small{\textcolor{black}{$\pm$ 3.54}} & 83.26 \small{\textcolor{black}{$\pm$ 12.47}}  &  69.92 \small{\textcolor{black}{$\pm$ 14.14}}\\
        EWC & 81.75 \small{\textcolor{black}{$\pm$ 1.42}} & 94.32 \small{\textcolor{black}{$\pm$ 0.88}}  & 90.82 \small{\textcolor{black}{$\pm$ 1.36}}  \\
        ER & 80.73 \small{\textcolor{black}{$\pm$ 0.37}} & 94.10 \small{\textcolor{black}{$\pm$ 1.39}}  &  90.92 \small{\textcolor{black}{$\pm$ 0.71}}  \\\hline
    \end{tabular}}
    \caption{Accuracy and standard deviation of the best performing models on different sub-questions in Taxonomy Domains.}
    \label{tab:tax_task_order}
\end{table}

Previous work on task-incremental learning for image classification~\cite{Yoon2020Scalable} has discussed the impact of task order to final performance, especially when tasks are dissimilar.
Similarly, we observe a high standard deviation in the Question Types results of  Table~\ref{tab:results}. 
In order to investiagte this further, we plot the final accuracy of a pretrained model for five training sequences in Figure~\ref{fig:qtask_order}, each ending with a different task. 
Our results show that task order can lead to Finetuning accuracy that varies more than 15$\%$. Although EWC improves the average accuracy, there is still a 10$\%$ fluctuation depending on the order. However, replay-based methods are able to improve performance and mitigate the sensitivity to task order.

While Table~\ref{tab:results} shows low variance in Taxonomy Domains, we find high variance when examining the performance on specific questions. In particular, we find that certain question types, such as Animals, Interior, and Sports, have high variance. Table~\ref{tab:tax_task_order} reveals a standard deviation which is up to 30 times higher compared to the average results in Table~\ref{tab:results}.
High standard deviation across randomized task orders is problematic since models can have different behavior in practice despite similar (aggregated) performance.
In other words, the current task performance will highly depend on the previous task order, even though the overall  accuracy from the randomized trials appears similar.

\section{Model Representations} \label{sec:CKA}

As described in Section \ref{ssec:models}, we compare different input representations of two single-stream transformer models: UNITER-base~\cite{chen2020uniter}, which uses region features extracted from a frozen pretrained object detector; and  ViLT-base~\cite{pmlr-v139-kim21k-vilt}, which 
directly embeds image patches.

\subsection{Continual Learning Results}
Figure~\ref{fig:uniter_vs_vilt} shows the performance of ViLT against UNITER when using naive finetuning, EWC and ER.
The compared continual learning strategies perform similarly with both backbones.
However, ViLT shows more forgetting especially in the case of question types. 
Although UNITER's region-based features are more robust to forgetting, they rely on a frozen pretrained object detector model.
This could limit the model's applicability to domains with larger visual distribution shifts.
Future work should focus on developing methods that perform well with V+L models that take image pixels as inputs.

\begin{figure*}
    \centering
    \begin{subfigure}[b]{0.32\textwidth}
        \centering
         \includegraphics[width=\textwidth]{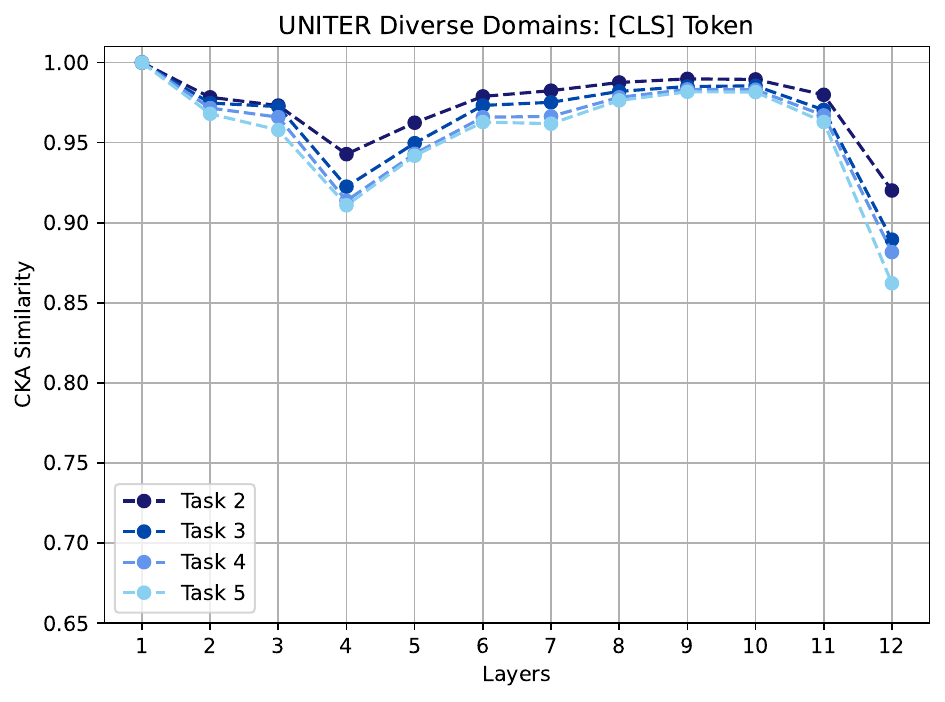}
        \end{subfigure}
     \hfill
     \begin{subfigure}[b]{0.32\textwidth}
        \centering
        \includegraphics[width=\textwidth]{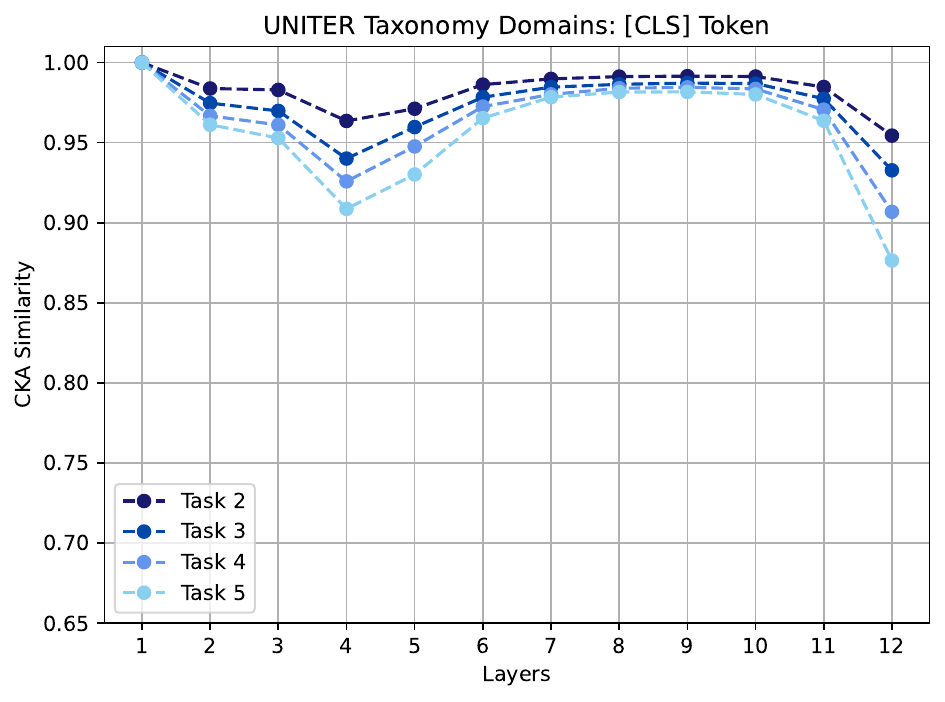}
    \end{subfigure}
    \hfill
    \begin{subfigure}[b]{0.32\textwidth}
    \centering
    \includegraphics[width=\textwidth]{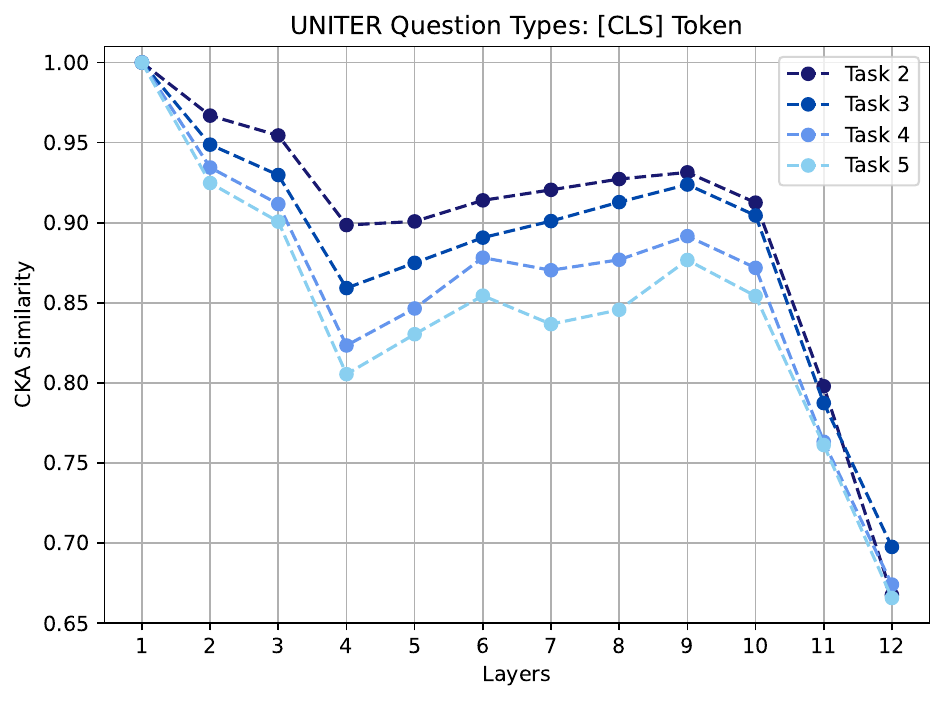}
    \end{subfigure}
    \hfill
     \\
     \begin{subfigure}[b]{0.32\textwidth}
        \centering
        \includegraphics[width=\textwidth]{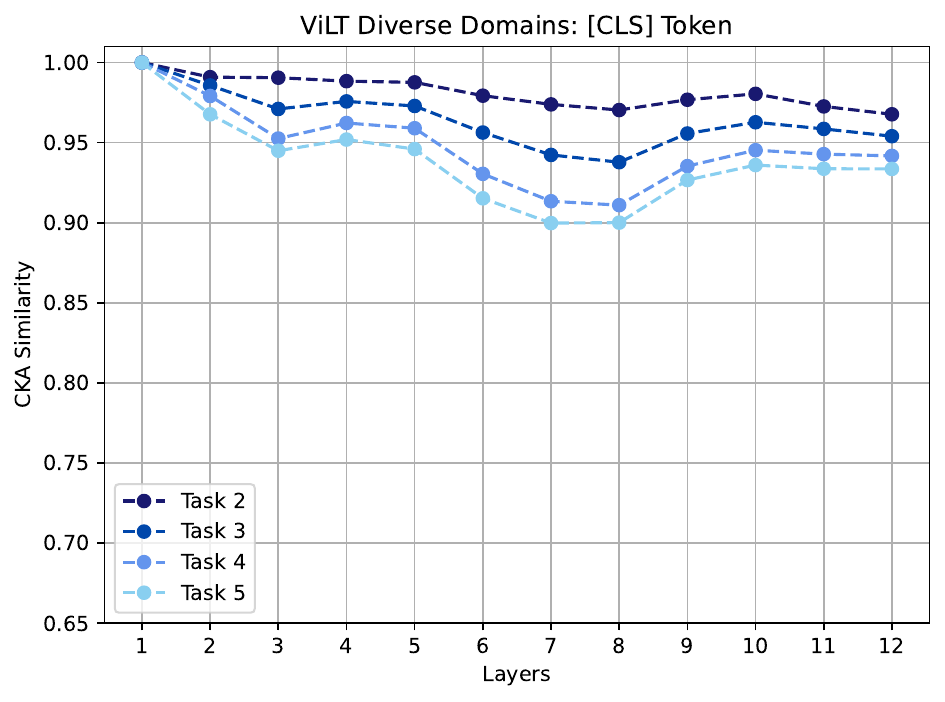}
    \end{subfigure}
    \hfill
    \begin{subfigure}[b]{0.32\textwidth}
        \centering
        \includegraphics[width=\textwidth]{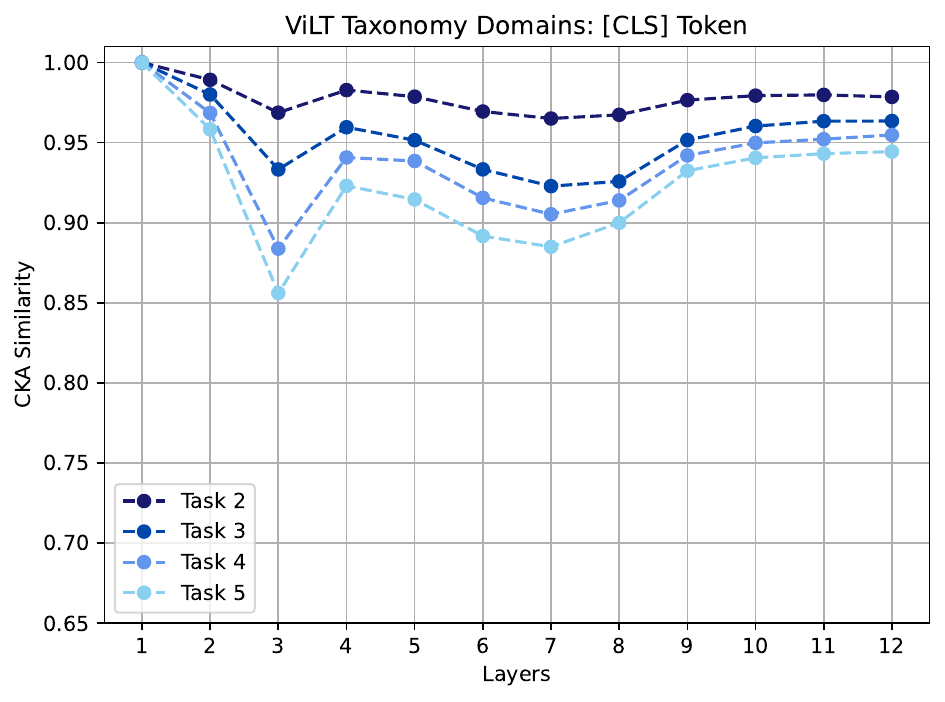}
    \end{subfigure}
    \hfill
    \begin{subfigure}[b]{0.32\textwidth}
    \centering
    \includegraphics[width=\textwidth]{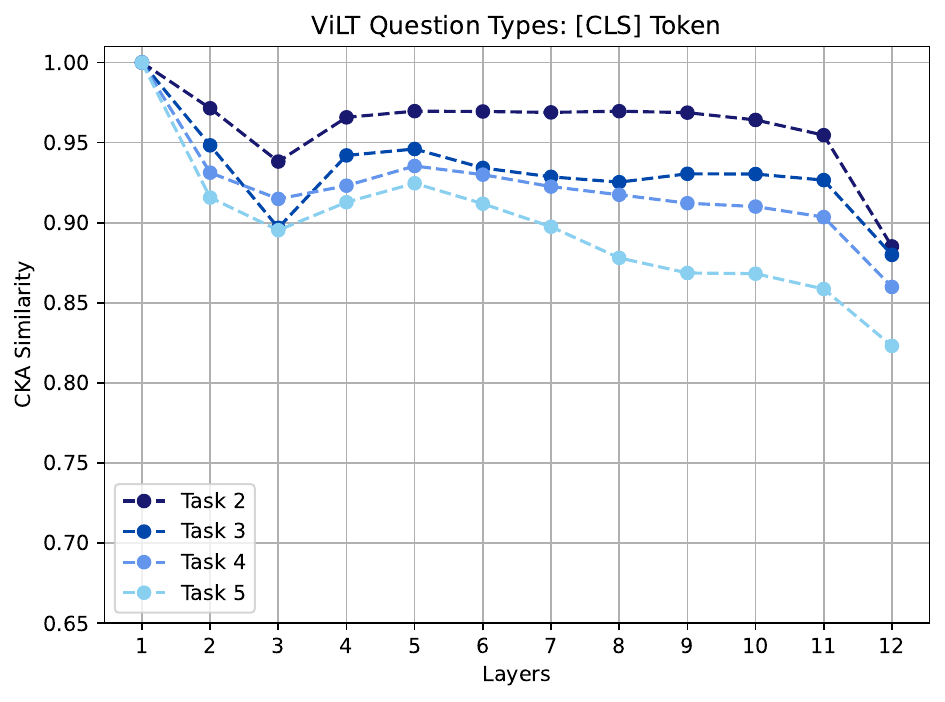}
   \end{subfigure}
    \hfill
    \caption{Representation similarity for the [CLS] token of samples from the first task after each training task. The first row corresponds to UNITER and the second row to ViLT representations. The columns from left to right refer to Diverse Domains, Taxonomy Domains, and Question Types. The layers that are affected are the final layers but also earlier layers (UNITER layer 4 and ViLT layer 3).}
    \label{fig:cka_cls}
\end{figure*} 
\begin{figure*}
    \begin{subfigure}[b]{0.32\textwidth}
        \centering
        \includegraphics[width=\textwidth]{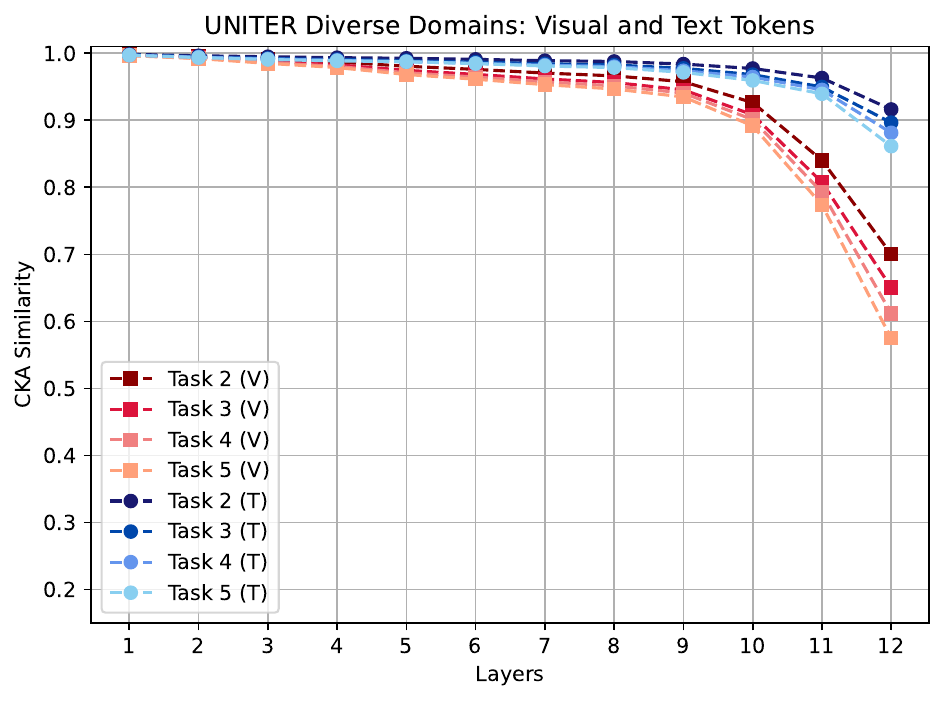}
    \end{subfigure}\hfill 
     \begin{subfigure}[b]{0.32\textwidth}
        \centering
        \includegraphics[width=\textwidth]{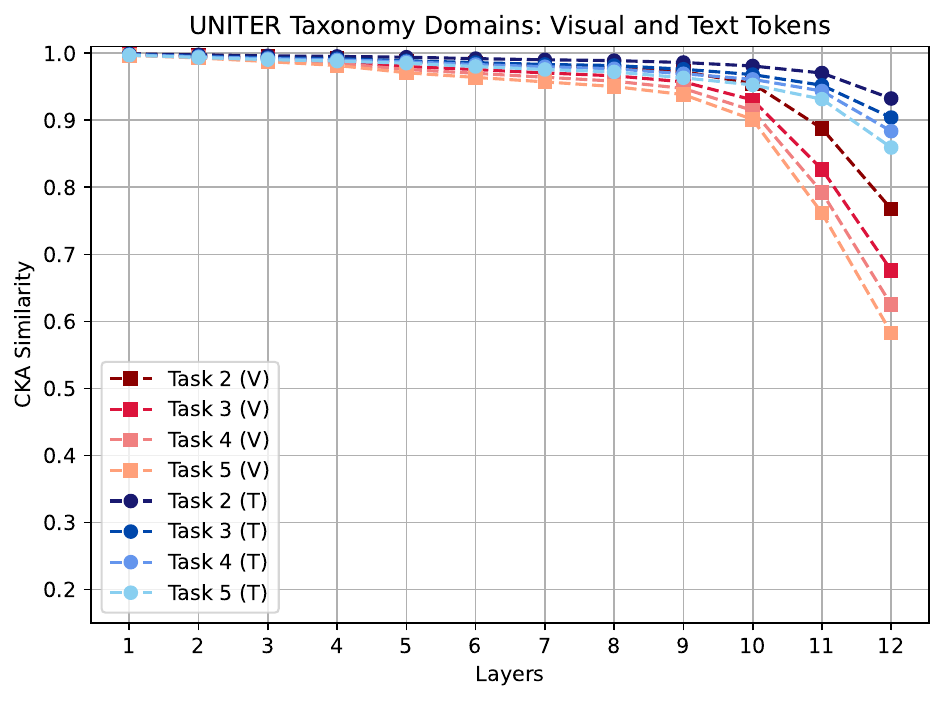}
    \end{subfigure}\hfill 
    \begin{subfigure}[b]{0.32\textwidth}
    \centering
    \includegraphics[width=\textwidth]{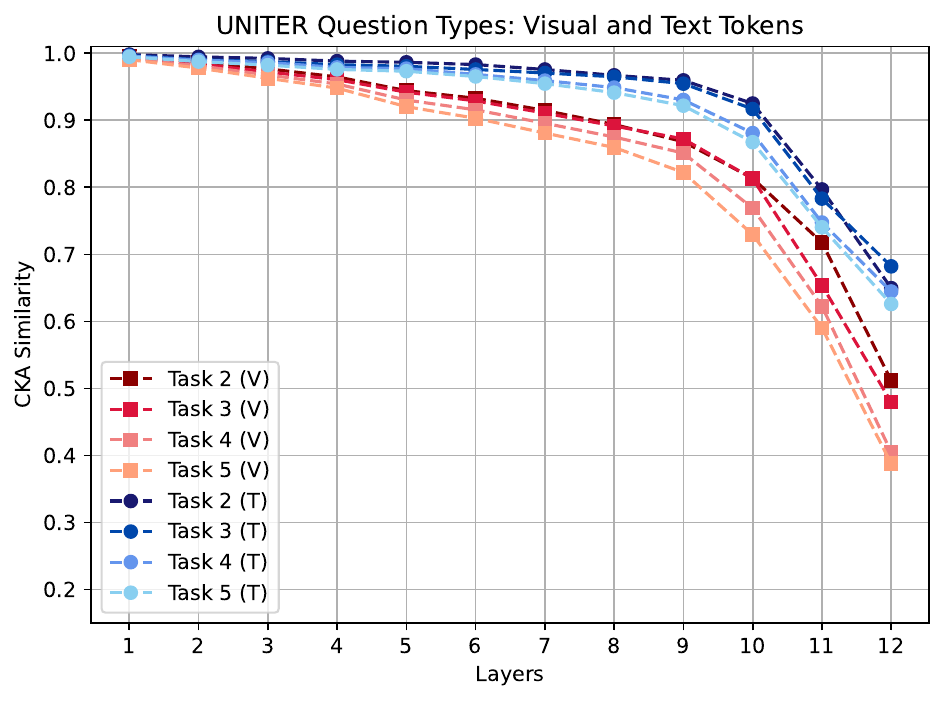}
\end{subfigure}\hfill 
     \\
    \begin{subfigure}[b]{0.32\textwidth}
         \centering
         \includegraphics[width=\textwidth]{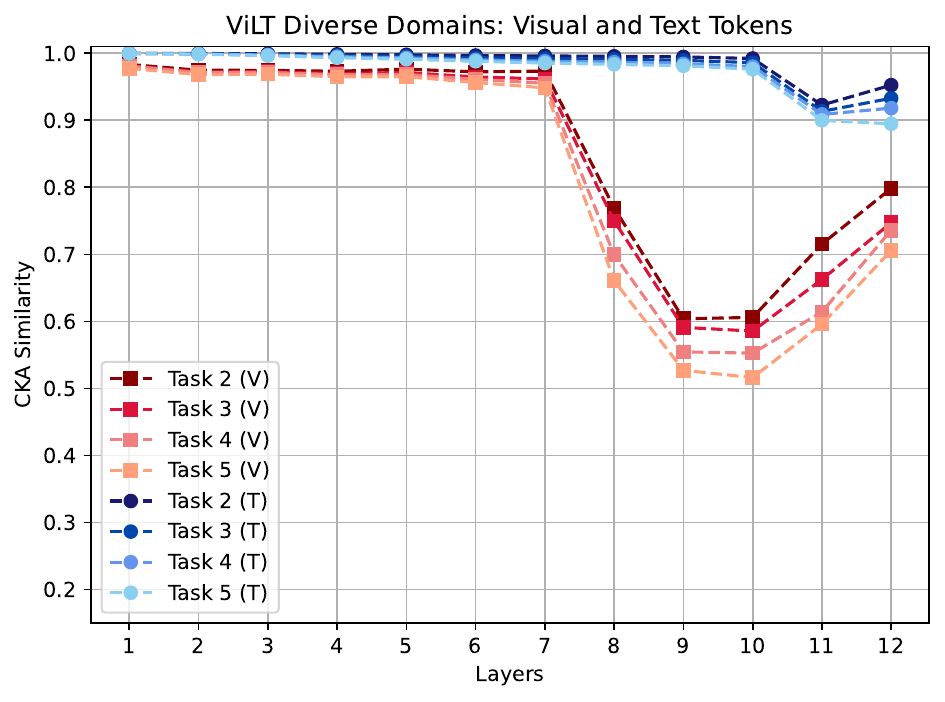}
     \end{subfigure}
     \hfill
    \begin{subfigure}[b]{0.32\textwidth}
         \centering
         \includegraphics[width=\textwidth]{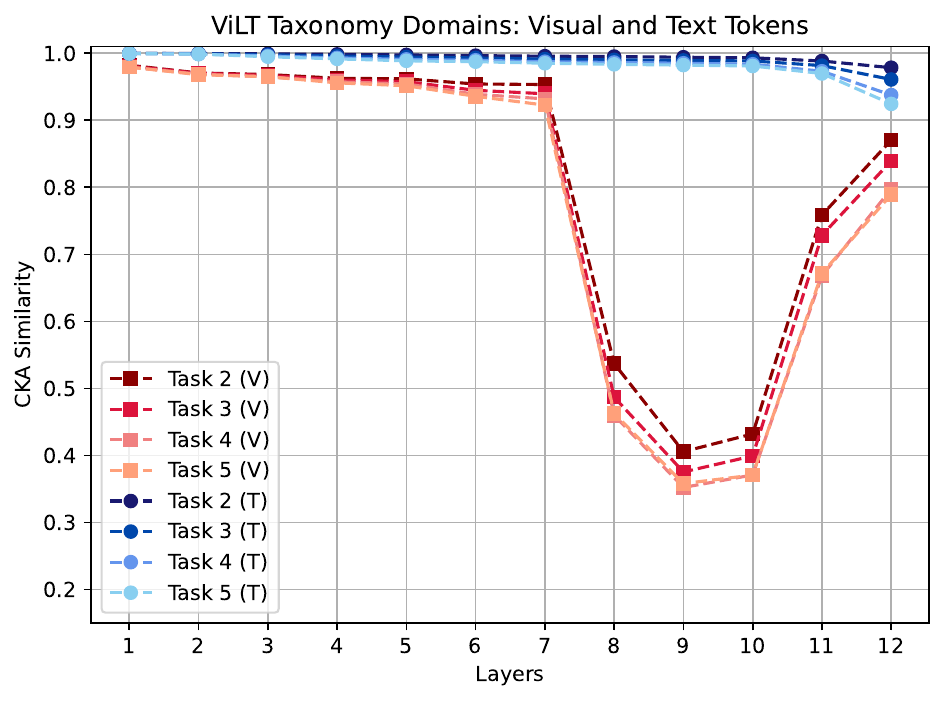}
     \end{subfigure}
     \begin{subfigure}[b]{0.32\textwidth}
     \hfill
         \centering
         \includegraphics[width=\textwidth]{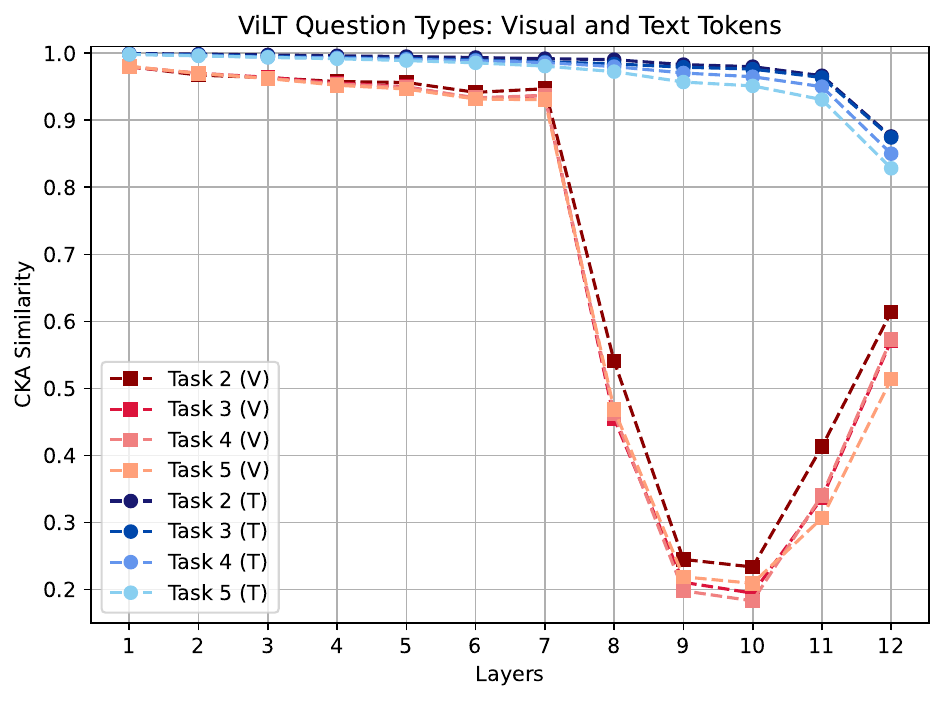}
     \end{subfigure}
     \hfill
     \\
    \caption{Representation similarity for the average visual (V) and text (T) tokens of of samples from the first task. The first row corresponds to UNITER and the second row to ViLT representations. The columns from left to right refer to Diverse Domains, Taxonomy Domains, and Question Types. The text representations from both models show decreased representation similarity for deeper layers. The visual representations of UNITER follow the same trend, while ViLT visual representations show a large similarity drop for layer 8-11.}
     \label{fig:cka_image_text}
\end{figure*}

\subsection{Representation Analysis}
Finally, we ask how representations from each modality evolve throughout the training sequence and compare this evolution across our continual learning settings. 
We use centered kernel alignment (CKA)~\cite{kornblith2019cka} to track the representation similarity of sequentially finetuned models. 
We extract representations $X^1_t$ of the validation data of the first task after training for each task $t=1\cdots T$, and measure the CKA similarity of $X^1_{t>1}$ to the original representations $X^1_1$. 

Figure~\ref{fig:cka_cls} shows the evolution of the representation similarity of the sentence-level representation [CLS] token per layer. 
Across the three settings, the representations of different layers change following a similar pattern but at different magnitudes which agree with the measured amount of forgetting.
Our results echo previous findings~\cite{wu2022pretrainedLM} showing that representations from deeper layers are more affected during continual learning, but there are also fragile earlier layers (UNITER layer 4, ViLT layer 3). 

Figure~\ref{fig:cka_image_text} shows the evolution of the average visual and text token representations per layer.
The representations of question tokens from both models retain higher similarity than image and [CLS] tokens.
In particular, ViLT visual representations show a large drop in representation similarity for layers 8-11. 
Since ViLT uses image patches instead of features extracted from a separate vision module, it needs to perform both feature extraction and multimodal alignment.
These results suggest that the features extracted from the visual inputs for VQA are more task-dependent and highlight the importance of stabilizing visual representations in deeper layers.

\section{Conclusion}
In this work, we provide an in-depth study of
task design for VQA and its impact when combining pretrained V+L models with continual learning approaches.
We empirically investigate the impact of task formulation, i.e.\ task design, order and similarity, by evaluating two transformer-based models and benchmarking several baseline methods.
We also propose SBWT as a new evaluation metric  that utilizes the semantic distance of answers.
Our results show that both task order and similarity, especially from the viewpoint of the answer distribution, highly influence performance.

These results are important for designing continual learning experiments for real-world settings that take into account how data become available over time. For example, the Taxonomy Domains resembles applications where data is continuously collected in different visual surroundings, whereas Question Types corresponds to `teaching' the system new reasoning capabilities. Our results suggest that the latter is the most challenging.
Our results also suggest that the easiest and thus `best-case' scenario is the `Diverse' data collection setup, where the system incrementally learns to recognize new objects which are randomly sampled from different domains.


In terms of model architectures, we investigated single-stream backbones with region features and image patches as inputs. 
Our representation analysis shows that image and text representations change at different scales. This implies that regularization-based approaches might be more suited for models with separate visual and text parameters where different regularization strengths are applied to each modality. 



\newpage



\newpage
\bibliographystyle{natbib}
\bibliography{anthology,main}

\begin{thebibliography}{52}
\expandafter\ifx\csname natexlab\endcsname\relax\def\natexlab#1{#1}\fi

\bibitem[{Anderson et~al.(2018)Anderson, He, Buehler, Teney, Johnson, Gould,
  and Zhang}]{Anderson2017up-down}
Peter Anderson, Xiaodong He, Chris Buehler, Damien Teney, Mark Johnson, Stephen
  Gould, and Lei Zhang. 2018.
\newblock \href
  {https://openaccess.thecvf.com/content_cvpr_2018/html/Anderson_Bottom-Up_and_Top-Down_CVPR_2018_paper.html}
  {Bottom-up and top-down attention for image captioning and visual question
  answering}.
\newblock In \emph{Proceedings of the IEEE Conference on Computer Vision and
  Pattern Recognition (CVPR)}.

\bibitem[{Bender et~al.(2021)Bender, Gebru, McMillan-Major, and
  Shmitchell}]{bender2021dangers}
Emily~M. Bender, Timnit Gebru, Angelina McMillan-Major, and Shmargaret
  Shmitchell. 2021.
\newblock \href {https://doi.org/10.1145/3442188.3445922} {On the dangers of
  stochastic parrots: Can language models be too big?}
\newblock In \emph{Proceedings of the 2021 ACM Conference on Fairness,
  Accountability, and Transparency}, FAccT '21, page 610–623, New York, NY,
  USA. Association for Computing Machinery.

\bibitem[{Chaudhry et~al.(2019{\natexlab{a}})Chaudhry, Ranzato, Rohrbach, and
  Elhoseiny}]{chaudhry2018agem}
Arslan Chaudhry, Marc’Aurelio Ranzato, Marcus Rohrbach, and Mohamed
  Elhoseiny. 2019{\natexlab{a}}.
\newblock \href {https://openreview.net/forum?id=Hkf2_sC5FX} {Efficient
  lifelong learning with a-{GEM}}.
\newblock In \emph{International Conference on Learning Representations}.

\bibitem[{Chaudhry et~al.(2019{\natexlab{b}})Chaudhry, Rohrbach, Elhoseiny,
  Ajanthan, Dokania, Torr, and Ranzato}]{Chaudhry2019er}
Arslan Chaudhry, Marcus Rohrbach, Mohamed Elhoseiny, Thalaiyasingam Ajanthan,
  Puneet~Kumar Dokania, Philip H.~S. Torr, and Marc'Aurelio Ranzato.
  2019{\natexlab{b}}.
\newblock \href {http://arxiv.org/abs/1902.10486} {Continual learning with tiny
  episodic memories.}
\newblock \emph{CoRR}, abs/1902.10486.

\bibitem[{Chen et~al.(2020)Chen, Li, Yu, El~Kholy, Ahmed, Gan, Cheng, and
  Liu}]{chen2020uniter}
Yen-Chun Chen, Linjie Li, Licheng Yu, Ahmed El~Kholy, Faisal Ahmed, Zhe Gan,
  Yu~Cheng, and Jingjing Liu. 2020.
\newblock \href {https://link.springer.com/chapter/10.1007/978-3-030-58577-8_7}
  {Uniter: Universal image-text representation learning}.
\newblock In \emph{Computer Vision -- ECCV 2020}, pages 104--120, Cham.
  Springer International Publishing.

\bibitem[{Del~Chiaro et~al.(2020)Del~Chiaro, Twardowski, Bagdanov, and van~de
  Weijer}]{del2020ratt}
Riccardo Del~Chiaro, Bart\l~omiej Twardowski, Andrew Bagdanov, and Joost van~de
  Weijer. 2020.
\newblock \href
  {https://proceedings.neurips.cc/paper/2020/file/c2964caac096f26db222cb325aa267cb-Paper.pdf}
  {Ratt: Recurrent attention to transient tasks for continual image
  captioning}.
\newblock In \emph{Advances in Neural Information Processing Systems},
  volume~33, pages 16736--16748. Curran Associates, Inc.

\bibitem[{Delange et~al.(2021)Delange, Aljundi, Masana, Parisot, Jia,
  Leonardis, Slabaugh, and Tuytelaars}]{Delangesurvey}
Matthias Delange, Rahaf Aljundi, Marc Masana, Sarah Parisot, Xu~Jia, Ales
  Leonardis, Greg Slabaugh, and Tinne Tuytelaars. 2021.
\newblock \href {https://doi.org/10.1109/TPAMI.2021.3057446} {A continual
  learning survey: Defying forgetting in classification tasks}.
\newblock \emph{IEEE Transactions on Pattern Analysis and Machine
  Intelligence}, pages 1--1.

\bibitem[{Goodfellow et~al.(2013)Goodfellow, Mirza, Xiao, Courville, and
  Bengio}]{goodfellow2013empirical}
Ian~J Goodfellow, Mehdi Mirza, Da~Xiao, Aaron Courville, and Yoshua Bengio.
  2013.
\newblock \href {https://arxiv.org/abs/1312.6211} {An empirical investigation
  of catastrophic forgetting in gradient-based neural networks}.
\newblock \emph{arXiv preprint arXiv:1312.6211}.

\bibitem[{Goyal et~al.(2017)Goyal, Khot, Summers-Stay, Batra, and
  Parikh}]{vqav2}
Yash Goyal, Tejas Khot, Douglas Summers-Stay, Dhruv Batra, and Devi Parikh.
  2017.
\newblock \href
  {https://openaccess.thecvf.com/content_cvpr_2017/html/Goyal_Making_the_v_CVPR_2017_paper.html}
  {Making the v in vqa matter: Elevating the role of image understanding in
  visual question answering}.
\newblock In \emph{Proceedings of the IEEE Conference on Computer Vision and
  Pattern Recognition (CVPR)}.

\bibitem[{Greco et~al.(2019)Greco, Plank, Fern{\'a}ndez, and
  Bernardi}]{greco2019psycholinguistics}
Claudio Greco, Barbara Plank, Raquel Fern{\'a}ndez, and Raffaella Bernardi.
  2019.
\newblock \href {https://doi.org/10.18653/v1/P19-1350} {Psycholinguistics meets
  continual learning: Measuring catastrophic forgetting in visual question
  answering}.
\newblock In \emph{Proceedings of the 57th Annual Meeting of the Association
  for Computational Linguistics}, pages 3601--3605, Florence, Italy.
  Association for Computational Linguistics.

\bibitem[{Hadsell et~al.(2020)Hadsell, Rao, Rusu, and
  Pascanu}]{HADSELL20201028embracing}
Raia Hadsell, Dushyant Rao, Andrei~A. Rusu, and Razvan Pascanu. 2020.
\newblock \href {https://doi.org/https://doi.org/10.1016/j.tics.2020.09.004}
  {Embracing change: Continual learning in deep neural networks}.
\newblock \emph{Trends in Cognitive Sciences}, 24(12):1028--1040.

\bibitem[{Hinton et~al.(2015)Hinton, Vinyals, and Dean}]{hinton2015distilling}
Geoffrey Hinton, Oriol Vinyals, and Jeff Dean. 2015.
\newblock \href {https://arxiv.org/abs/1503.02531} {Distilling the knowledge in
  a neural network}.
\newblock \emph{arXiv preprint arXiv:1503.02531}.

\bibitem[{Hussain et~al.(2021)Hussain, Holla, Mishra, Yannakoudakis, and
  Shutova}]{hussain2021towards}
Aman Hussain, Nithin Holla, Pushkar Mishra, Helen Yannakoudakis, and Ekaterina
  Shutova. 2021.
\newblock \href {https://openreview.net/forum?id=yJyIjWyPJgs} {Towards a robust
  experimental framework and benchmark for lifelong language learning}.
\newblock In \emph{Thirty-fifth Conference on Neural Information Processing
  Systems Datasets and Benchmarks Track (Round 1)}.

\bibitem[{Jin et~al.(2020)Jin, Du, Sadhu, Nevatia, and Ren}]{jin2020viscol}
Xisen Jin, Junyi Du, Arka Sadhu, Ram Nevatia, and Xiang Ren. 2020.
\newblock \href {https://doi.org/10.18653/v1/2020.emnlp-main.158} {Visually
  grounded continual learning of compositional phrases}.
\newblock In \emph{Proceedings of the 2020 Conference on Empirical Methods in
  Natural Language Processing (EMNLP)}, pages 2018--2029, Online. Association
  for Computational Linguistics.

\bibitem[{Kafle et~al.(2017)Kafle, Yousefhussien, and Kanan}]{kafle2017data}
Kushal Kafle, Mohammed Yousefhussien, and Christopher Kanan. 2017.
\newblock \href {https://doi.org/10.18653/v1/W17-3529} {Data augmentation for
  visual question answering}.
\newblock In \emph{Proceedings of the 10th International Conference on Natural
  Language Generation}, pages 198--202, Santiago de Compostela, Spain.
  Association for Computational Linguistics.

\bibitem[{Kil et~al.(2021)Kil, Zhang, Xuan, and Chao}]{kil2021discovering}
Jihyung Kil, Cheng Zhang, Dong Xuan, and Wei-Lun Chao. 2021.
\newblock \href {https://arxiv.org/abs/2109.06122} {Discovering the unknown
  knowns: Turning implicit knowledge in the dataset into explicit training
  examples for visual question answering}.
\newblock \emph{arXiv preprint arXiv:2109.06122}.

\bibitem[{Kim et~al.(2021)Kim, Son, and Kim}]{pmlr-v139-kim21k-vilt}
Wonjae Kim, Bokyung Son, and Ildoo Kim. 2021.
\newblock \href {http://proceedings.mlr.press/v139/kim21k.html} {Vilt:
  Vision-and-language transformer without convolution or region supervision}.
\newblock In \emph{Proceedings of the 38th International Conference on Machine
  Learning}, volume 139 of \emph{Proceedings of Machine Learning Research},
  pages 5583--5594. PMLR.

\bibitem[{Kirkpatrick et~al.(2017)Kirkpatrick, Pascanu, Rabinowitz, Veness,
  Desjardins, Rusu, Milan, Quan, Ramalho, Grabska-Barwinska
  et~al.}]{kirkpatrick2017elastic}
James Kirkpatrick, Razvan Pascanu, Neil Rabinowitz, Joel Veness, Guillaume
  Desjardins, Andrei~A Rusu, Kieran Milan, John Quan, Tiago Ramalho, Agnieszka
  Grabska-Barwinska, et~al. 2017.
\newblock \href {https://www.pnas.org/content/114/13/3521} {Overcoming
  catastrophic forgetting in neural networks}.
\newblock \emph{Proceedings of the national academy of sciences},
  114(13):3521--3526.

\bibitem[{Kornblith et~al.(2019)Kornblith, Norouzi, Lee, and
  Hinton}]{kornblith2019cka}
Simon Kornblith, Mohammad Norouzi, Honglak Lee, and Geoffrey Hinton. 2019.
\newblock \href {https://proceedings.mlr.press/v97/kornblith19a.html}
  {Similarity of neural network representations revisited}.
\newblock In \emph{Proceedings of the 36th International Conference on Machine
  Learning}, volume~97 of \emph{Proceedings of Machine Learning Research},
  pages 3519--3529. PMLR.

\bibitem[{Lazaridou et~al.(2021)Lazaridou, Kuncoro, Gribovskaya, Agrawal,
  Liska, Terzi, Gimenez, de~Masson~d'Autume, Ko{\v{c}}isk{\'y}, Ruder,
  Yogatama, Cao, Young, and Blunsom}]{lazaridou2021pitfalls}
Angeliki Lazaridou, Adhiguna Kuncoro, Elena Gribovskaya, Devang Agrawal, Adam
  Liska, Tayfun Terzi, Mai Gimenez, Cyprien de~Masson~d'Autume, Tom{\'a}{\v{s}}
  Ko{\v{c}}isk{\'y}, Sebastian Ruder, Dani Yogatama, Kris Cao, Susannah Young,
  and Phil Blunsom. 2021.
\newblock \href {https://openreview.net/forum?id=73OmmrCfSyy} {Mind the gap:
  Assessing temporal generalization in neural language models}.
\newblock In \emph{Advances in Neural Information Processing Systems}.

\bibitem[{Lee(2001)}]{pmlr-vR3-lee01a}
Lillian Lee. 2001.
\newblock \href {https://proceedings.mlr.press/r3/lee01a.html} {On the
  effectiveness of the skew divergence for statistical language analysis}.
\newblock In \emph{Proceedings of the Eighth International Workshop on
  Artificial Intelligence and Statistics}, volume~R3 of \emph{Proceedings of
  Machine Learning Research}, pages 176--183. PMLR.
\newblock Reissued by PMLR on 31 March 2021.

\bibitem[{Lee et~al.(2021)Lee, Goldt, and Saxe}]{lee2021continual}
Sebastian Lee, Sebastian Goldt, and Andrew Saxe. 2021.
\newblock \href {http://proceedings.mlr.press/v139/lee21e/lee21e.pdf}
  {Continual learning in the teacher-student setup: Impact of task similarity}.
\newblock In \emph{2021 International Conference on Machine Learning}.

\bibitem[{Lei et~al.(2022)Lei, Gao, Wu, Wang, Liu, Zhang, and
  Shou}]{lei2022symbolic}
Stan~Weixian Lei, Difei Gao, Jay~Zhangjie Wu, Yuxuan Wang, Wei Liu, Mengmi
  Zhang, and Mike~Zheng Shou. 2022.
\newblock Symbolic replay: Scene graph as prompt for continual learning on vqa
  task.
\newblock \emph{arXiv preprint arXiv:2208.12037}.

\bibitem[{Li and Hoiem(2018)}]{li2017lwf}
Zhizhong Li and Derek Hoiem. 2018.
\newblock \href {https://doi.org/10.1109/TPAMI.2017.2773081} {Learning without
  forgetting}.
\newblock \emph{IEEE Transactions on Pattern Analysis and Machine
  Intelligence}, 40(12):2935--2947.

\bibitem[{Lin et~al.(2014)Lin, Maire, Belongie, Hays, Perona, Ramanan,
  Doll{\'a}r, and Zitnick}]{lin2014coco}
Tsung-Yi Lin, Michael Maire, Serge Belongie, James Hays, Pietro Perona, Deva
  Ramanan, Piotr Doll{\'a}r, and C.~Lawrence Zitnick. 2014.
\newblock \href
  {https://link.springer.com/chapter/10.1007/978-3-319-10602-1_48} {Microsoft
  coco: Common objects in context}.
\newblock In \emph{Computer Vision -- ECCV 2014}, pages 740--755, Cham.
  Springer International Publishing.

\bibitem[{Lomonaco and Maltoni(2017)}]{lomonaco2017core50}
Vincenzo Lomonaco and Davide Maltoni. 2017.
\newblock \href {https://proceedings.mlr.press/v78/lomonaco17a.html} {Core50: a
  new dataset and benchmark for continuous object recognition}.
\newblock In \emph{Proceedings of the 1st Annual Conference on Robot Learning},
  volume~78 of \emph{Proceedings of Machine Learning Research}, pages 17--26.
  PMLR.

\bibitem[{Lopez-Paz and Ranzato(2017)}]{lopez2017GEM}
David Lopez-Paz and Marc'Aurelio Ranzato. 2017.
\newblock \href
  {http://papers.nips.cc/paper/7225-gradient-episodic-memory-for-continual-learning}
  {Gradient episodic memory for continual learning}.
\newblock In \emph{Advances in Neural Information Processing Systems},
  volume~30, pages 6470--6479.

\bibitem[{Lu et~al.(2020)Lu, Goswami, Rohrbach, Parikh, and Lee}]{lu202012}
Jiasen Lu, Vedanuj Goswami, Marcus Rohrbach, Devi Parikh, and Stefan Lee. 2020.
\newblock \href
  {https://openaccess.thecvf.com/content_CVPR_2020/html/Lu_12-in-1_Multi-Task_Vision_and_Language_Representation_Learning_CVPR_2020_paper.html}
  {12-in-1: Multi-task vision and language representation learning}.
\newblock In \emph{Proceedings of the IEEE/CVF Conference on Computer Vision
  and Pattern Recognition (CVPR)}.

\bibitem[{McCloskey and Cohen(1989)}]{mccloskey1989catastrophic}
Michael McCloskey and Neal~J Cohen. 1989.
\newblock \href
  {https://www.sciencedirect.com/science/article/abs/pii/S0079742108605368}
  {Catastrophic interference in connectionist networks: The sequential learning
  problem}.
\newblock In \emph{Psychology of learning and motivation}, volume~24, pages
  109--165. Elsevier.

\bibitem[{Mehta et~al.(2021)Mehta, Patil, Chandar, and
  Strubell}]{mehta2021empirical}
Sanket~Vaibhav Mehta, Darshan Patil, Sarath Chandar, and Emma Strubell. 2021.
\newblock \href {https://arxiv.org/abs/2112.09153} {An empirical investigation
  of the role of pre-training in lifelong learning}.
\newblock \emph{arXiv preprint arXiv:2112.09153}.

\bibitem[{Mi et~al.(2020)Mi, Kong, Lin, Yu, and Faltings}]{mi2020generalized}
Fei Mi, Lingjing Kong, Tao Lin, Kaicheng Yu, and Boi Faltings. 2020.
\newblock Generalized class incremental learning.
\newblock In \emph{Proceedings of the IEEE/CVF Conference on Computer Vision
  and Pattern Recognition Workshops}, pages 240--241.

\bibitem[{Mirzadeh et~al.(2020)Mirzadeh, Farajtabar, Pascanu, and
  Ghasemzadeh}]{mirzadeh2020understanding}
Seyed~Iman Mirzadeh, Mehrdad Farajtabar, Razvan Pascanu, and Hassan
  Ghasemzadeh. 2020.
\newblock \href
  {https://proceedings.neurips.cc/paper/2020/file/518a38cc9a0173d0b2dc088166981cf8-Paper.pdf}
  {Understanding the role of training regimes in continual learning}.
\newblock In \emph{Advances in Neural Information Processing Systems},
  volume~33, pages 7308--7320. Curran Associates, Inc.

\bibitem[{Nguyen et~al.(2019{\natexlab{a}})Nguyen, Achille, Lam, Hassner,
  Mahadevan, and Soatto}]{nguyen-2019-understanding}
Cuong~V. Nguyen, Alessandro Achille, Michael Lam, Tal Hassner, Vijay Mahadevan,
  and Stefano Soatto. 2019{\natexlab{a}}.
\newblock \href {http://arxiv.org/abs/1908.01091} {Toward understanding
  catastrophic forgetting in continual learning.}
\newblock \emph{CoRR}, abs/1908.01091.

\bibitem[{Nguyen et~al.(2019{\natexlab{b}})Nguyen, Jun, Tran, Yalew, and
  Kim}]{nguyen2019contcap}
Giang Nguyen, Tae~Joon Jun, Trung Tran, Tolcha Yalew, and Daeyoung Kim.
  2019{\natexlab{b}}.
\newblock \href {https://arxiv.org/abs/1909.08745} {Contcap: A scalable
  framework for continual image captioning}.
\newblock \emph{arXiv preprint arXiv:1909.08745}.

\bibitem[{Pennington et~al.(2014)Pennington, Socher, and
  Manning}]{pennington2014glove}
Jeffrey Pennington, Richard Socher, and Christopher~D. Manning. 2014.
\newblock \href {http://www.aclweb.org/anthology/D14-1162} {Glove: Global
  vectors for word representation}.
\newblock In \emph{Empirical Methods in Natural Language Processing (EMNLP)},
  pages 1532--1543.

\bibitem[{Ramasesh et~al.(2021)Ramasesh, Dyer, and Raghu}]{ramasesh2020anatomy}
Vinay~Venkatesh Ramasesh, Ethan Dyer, and Maithra Raghu. 2021.
\newblock \href {https://openreview.net/forum?id=LhY8QdUGSuw} {Anatomy of
  catastrophic forgetting: Hidden representations and task semantics}.
\newblock In \emph{International Conference on Learning Representations}.

\bibitem[{Ratcliff(1990)}]{ratcliff1990connectionist}
Roger Ratcliff. 1990.
\newblock \href {https://psycnet.apa.org/fulltext/1990-18992-001.html}
  {Connectionist models of recognition memory: constraints imposed by learning
  and forgetting functions.}
\newblock \emph{Psychological review}, 97(2):285.

\bibitem[{Rebuffi et~al.(2017)Rebuffi, Kolesnikov, Sperl, and
  Lampert}]{rebuffi2017icarl}
Sylvestre-Alvise Rebuffi, Alexander Kolesnikov, Georg Sperl, and Christoph~H
  Lampert. 2017.
\newblock \href
  {https://openaccess.thecvf.com/content_cvpr_2017/html/Rebuffi_iCaRL_Incremental_Classifier_CVPR_2017_paper.html}
  {icarl: Incremental classifier and representation learning}.
\newblock In \emph{Proceedings of the IEEE conference on Computer Vision and
  Pattern Recognition}, pages 2001--2010.

\bibitem[{Reimers and Gurevych(2019)}]{reimers-2019-sentence-bert}
Nils Reimers and Iryna Gurevych. 2019.
\newblock \href {https://arxiv.org/abs/1908.10084} {Sentence-bert: Sentence
  embeddings using siamese bert-networks}.
\newblock In \emph{Proceedings of the 2019 Conference on Empirical Methods in
  Natural Language Processing}. Association for Computational Linguistics.

\bibitem[{Riemer et~al.(2019)Riemer, Cases, Ajemian, Liu, Rish, Tu, , and
  Tesauro}]{riemer2018learning}
Matthew Riemer, Ignacio Cases, Robert Ajemian, Miao Liu, Irina Rish, Yuhai Tu,
  , and Gerald Tesauro. 2019.
\newblock \href {https://openreview.net/forum?id=B1gTShAct7} {Learning to learn
  without forgetting by maximizing transfer and minimizing interference}.
\newblock In \emph{International Conference on Learning Representations}.

\bibitem[{Ruder and Plank(2017)}]{ruder-plank-2017-learning}
Sebastian Ruder and Barbara Plank. 2017.
\newblock \href {https://doi.org/10.18653/v1/D17-1038} {Learning to select data
  for transfer learning with {B}ayesian optimization}.
\newblock In \emph{Proceedings of the 2017 Conference on Empirical Methods in
  Natural Language Processing}, pages 372--382, Copenhagen, Denmark.
  Association for Computational Linguistics.

\bibitem[{Srinivasan et~al.(2022)Srinivasan, Chang, Alva, Chochlakis, Rostami,
  and Thomason}]{srinivasan2022climb}
Tejas Srinivasan, Ting-Yun Chang, Leticia Leonor~Pinto Alva, Georgios
  Chochlakis, Mohammad Rostami, and Jesse Thomason. 2022.
\newblock Climb: A continual learning benchmark for vision-and-language tasks.
\newblock \emph{arXiv preprint arXiv:2206.09059}.

\bibitem[{Standley et~al.(2020)Standley, Zamir, Chen, Guibas, Malik, and
  Savarese}]{pmlr-v119-standley20a}
Trevor Standley, Amir Zamir, Dawn Chen, Leonidas Guibas, Jitendra Malik, and
  Silvio Savarese. 2020.
\newblock \href {https://proceedings.mlr.press/v119/standley20a.html} {Which
  tasks should be learned together in multi-task learning?}
\newblock In \emph{Proceedings of the 37th International Conference on Machine
  Learning}, volume 119 of \emph{Proceedings of Machine Learning Research},
  pages 9120--9132. PMLR.

\bibitem[{Strubell et~al.(2019)Strubell, Ganesh, and
  McCallum}]{strubell-etal-2019-energy}
Emma Strubell, Ananya Ganesh, and Andrew McCallum. 2019.
\newblock \href {https://doi.org/10.18653/v1/P19-1355} {Energy and policy
  considerations for deep learning in {NLP}}.
\newblock In \emph{Proceedings of the 57th Annual Meeting of the Association
  for Computational Linguistics}, pages 3645--3650, Florence, Italy.
  Association for Computational Linguistics.

\bibitem[{Suglia et~al.(2020)Suglia, Konstas, Vanzo, Bastianelli, Elliott,
  Frank, and Lemon}]{suglia-etal-2020-compguesswhat}
Alessandro Suglia, Ioannis Konstas, Andrea Vanzo, Emanuele Bastianelli, Desmond
  Elliott, Stella Frank, and Oliver Lemon. 2020.
\newblock \href {https://doi.org/10.18653/v1/2020.acl-main.682}
  {{C}omp{G}uess{W}hat?!: A multi-task evaluation framework for grounded
  language learning}.
\newblock In \emph{Proceedings of the 58th Annual Meeting of the Association
  for Computational Linguistics}, pages 7625--7641, Online. Association for
  Computational Linguistics.

\bibitem[{Tarvainen and Valpola(2017)}]{mean-teachers}
Antti Tarvainen and Harri Valpola. 2017.
\newblock \href {https://openreview.net/forum?id=ry8u21rtl} {Mean teachers are
  better role models: Weight-averaged consistency targets improve
  semi-supervised deep learning results}.
\newblock In \emph{Proceedings of the 31st International Conference on Neural
  Information Processing Systems}, NEURIPS'17, page 1195–1204, Red Hook, NY,
  USA. Curran Associates Inc.

\bibitem[{Van~de Ven and Tolias(2018)}]{van2018generative}
Gido~M Van~de Ven and Andreas~S Tolias. 2018.
\newblock Generative replay with feedback connections as a general strategy for
  continual learning.
\newblock \emph{arXiv preprint arXiv:1809.10635}.

\bibitem[{Van~de Ven and Tolias(2019)}]{van2019three}
Gido~M Van~de Ven and Andreas~S Tolias. 2019.
\newblock \href {https://arxiv.org/abs/1904.07734} {Three scenarios for
  continual learning}.
\newblock \emph{arXiv preprint arXiv:1904.07734}.

\bibitem[{Whitehead et~al.(2021)Whitehead, Wu, Ji, Feris, and
  Saenko}]{whitehead2021separating}
Spencer Whitehead, Hui Wu, Heng Ji, Rogerio Feris, and Kate Saenko. 2021.
\newblock \href
  {https://openaccess.thecvf.com/content/CVPR2021/html/Whitehead_Separating_Skills_and_Concepts_for_Novel_Visual_Question_Answering_CVPR_2021_paper.html}
  {Separating skills and concepts for novel visual question answering}.
\newblock In \emph{Proceedings of the IEEE/CVF Conference on Computer Vision
  and Pattern Recognition (CVPR)}, pages 5632--5641.

\bibitem[{Wu et~al.(2022)Wu, Caccia, Li, Li, Qi, and
  Haffari}]{wu2022pretrainedLM}
Tongtong Wu, Massimo Caccia, Zhuang Li, Yuan-Fang Li, Guilin Qi, and Gholamreza
  Haffari. 2022.
\newblock \href {https://openreview.net/forum?id=figzpGMrdD} {Pretrained
  language model in continual learning: A comparative study}.
\newblock In \emph{International Conference on Learning Representations}.

\bibitem[{Yoon et~al.(2020)Yoon, Kim, Yang, and Hwang}]{Yoon2020Scalable}
Jaehong Yoon, Saehoon Kim, Eunho Yang, and Sung~Ju Hwang. 2020.
\newblock \href {https://openreview.net/forum?id=r1gdj2EKPB} {Scalable and
  order-robust continual learning with additive parameter decomposition}.
\newblock In \emph{International Conference on Learning Representations}.

\bibitem[{Zamir et~al.(2018)Zamir, Sax, Shen, Guibas, Malik, and
  Savarese}]{zamir2018taskonomy}
Amir~R. Zamir, Alexander Sax, William Shen, Leonidas~J. Guibas, Jitendra Malik,
  and Silvio Savarese. 2018.
\newblock \href
  {https://openaccess.thecvf.com/content_cvpr_2018/html/Zamir_Taskonomy_Disentangling_Task_CVPR_2018_paper.html}
  {Taskonomy: Disentangling task transfer learning}.
\newblock In \emph{Proceedings of the IEEE Conference on Computer Vision and
  Pattern Recognition (CVPR)}.

\end{thebibliography}

\newpage
\appendix
\section{Data Details} \label{appendix:dataset}
We investigate three continual learning settings based on the VQA-v2 dataset~\cite{vqav2}, a collection of visual question annotations in English.
Tasks in the Diverse Domains setting are created by grouping $10$ objects from COCO annotations~\cite{lin2014coco} as follows:
\begin{itemize}[leftmargin=*]
    \item Group 1: bird, car, keyboard, motorcycle, orange, pizza, sink, sports ball, toilet, zebra
    \item Group 2: airplane, baseball glove, bed, bus, cow, donut, giraffe, horse, mouse, sheep
    \item Group 3: boat, broccoli, hot dog, kite, oven, sandwich, snowboard, surfboard, tennis racket, TV
    \item Group 4: apple, baseball bat, bear, bicycle, cake, laptop, microwave, potted plant, remote, train
    \item Group 5: banana, carrot, cell phone, chair, couch, elephant, refrigerator, skateboard, toaster, truck
\end{itemize}

We also provide a few example questions for each task in Question Types:
\begin{itemize}[leftmargin=*]
    \item Action: What is the cat doing?, Is the man catching the ball?, What is this sport?
    \item Color: What color is the ground?, What color is the right top umbrella?
    \item Count: How many skaters are there?, How many elephants?, How many rooms do you see?
    \item Scene: Is the picture taken inside?, Is this photo black and white?, What is the weather like?
    \item Subcategory: What type of vehicle is this?, What utensil is on the plate?, What kind of car is it?
\end{itemize}

\begin{figure*}[ht!]
    \centering
     \begin{subfigure}[b]{0.48\textwidth}
         \centering
         \includegraphics[width=\textwidth]{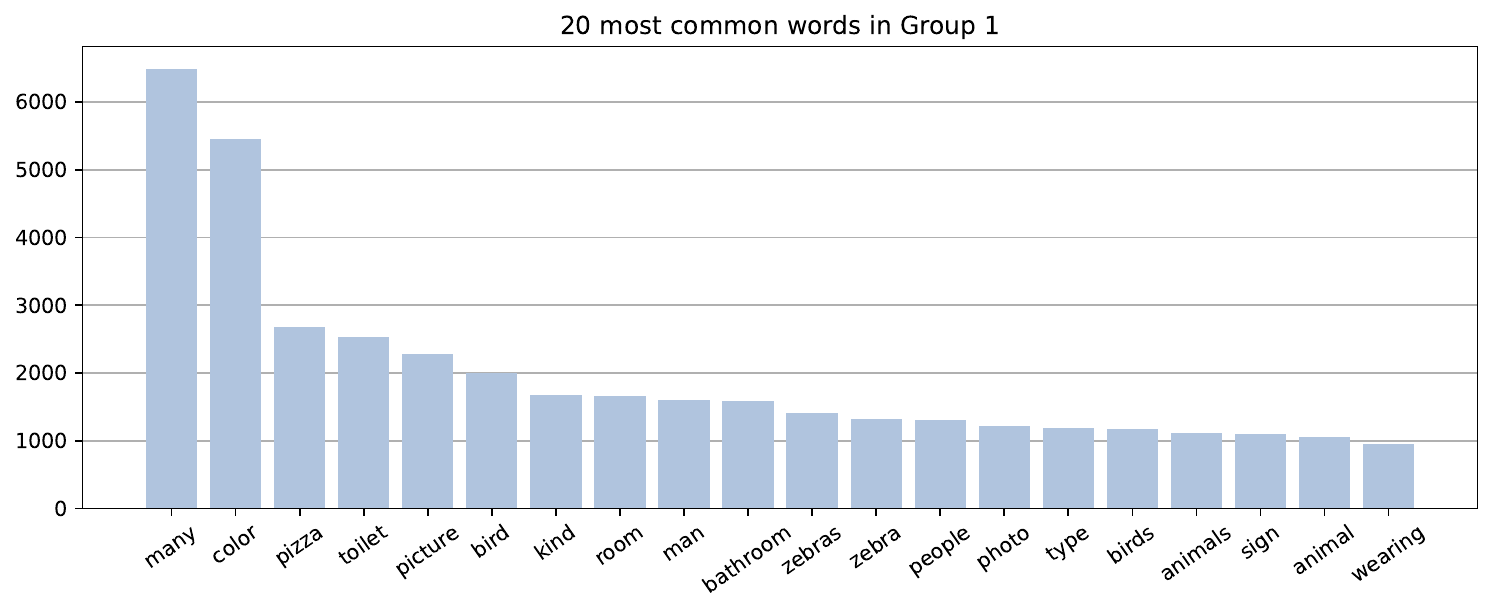}
     \end{subfigure}
     \hfill
     \begin{subfigure}[b]{0.48\textwidth}
         \centering
         \includegraphics[width=\textwidth]{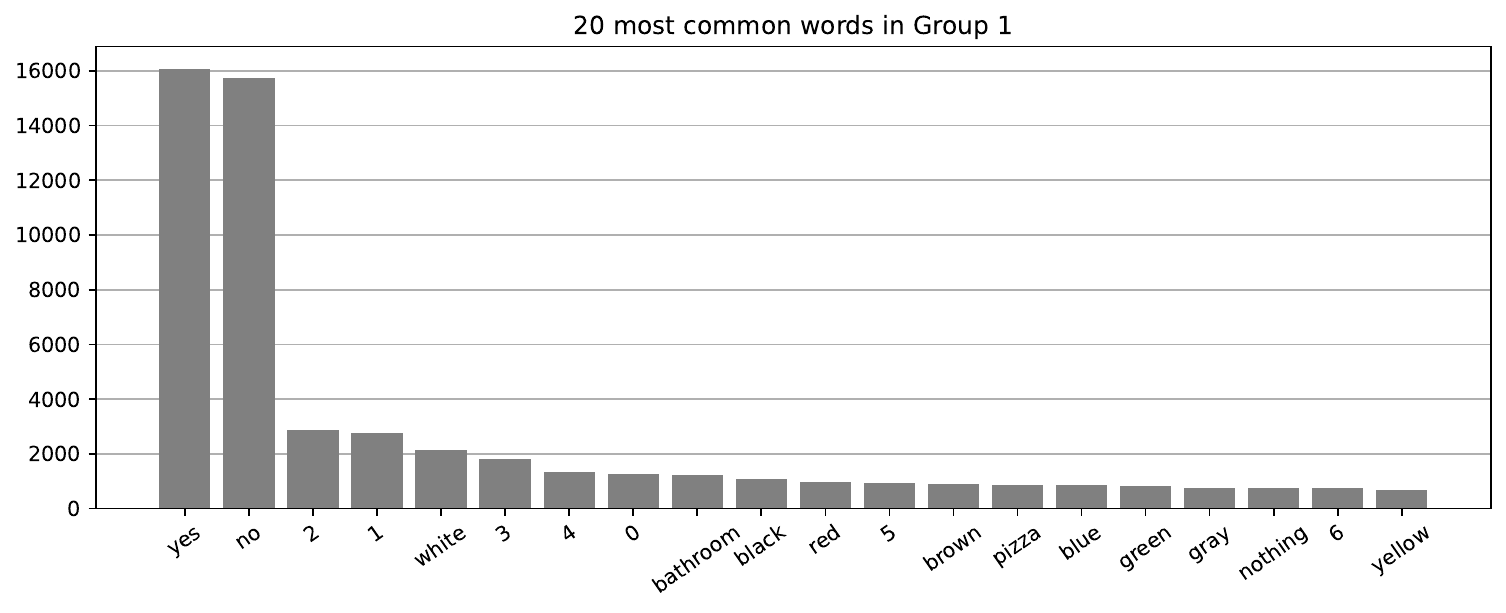}
     \end{subfigure}
     \hfill
     \begin{subfigure}[b]{0.48\textwidth}
         \centering
         \includegraphics[width=\textwidth]{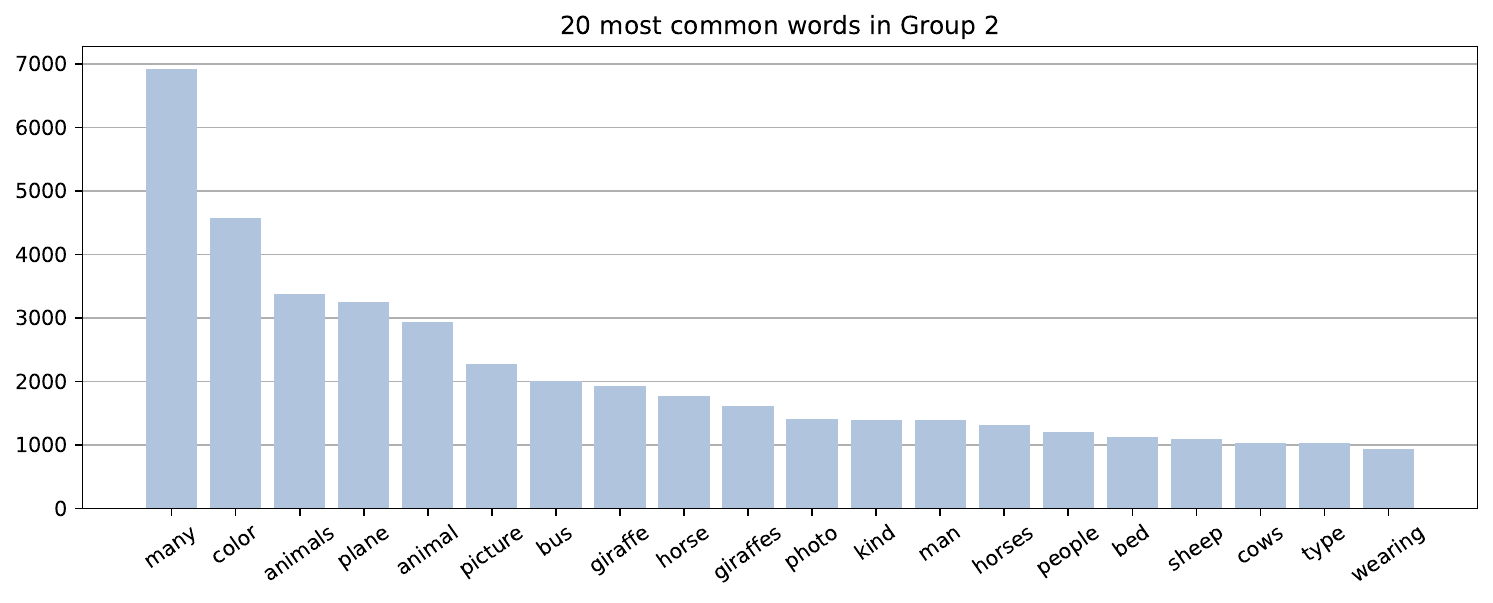}
     \end{subfigure}
     \hfill
     \begin{subfigure}[b]{0.48\textwidth}
         \centering
         \includegraphics[width=\textwidth]{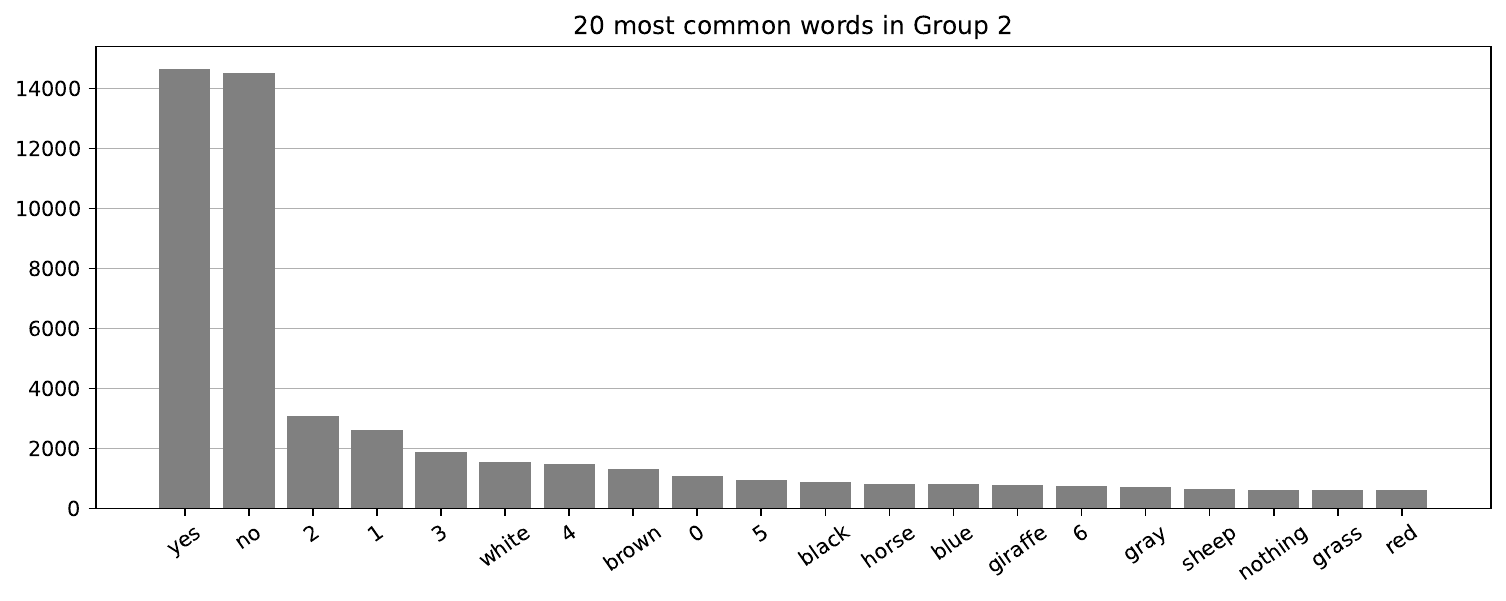}
     \end{subfigure}
     \hfill
     \begin{subfigure}[b]{0.48\textwidth}
         \centering
         \includegraphics[width=\textwidth]{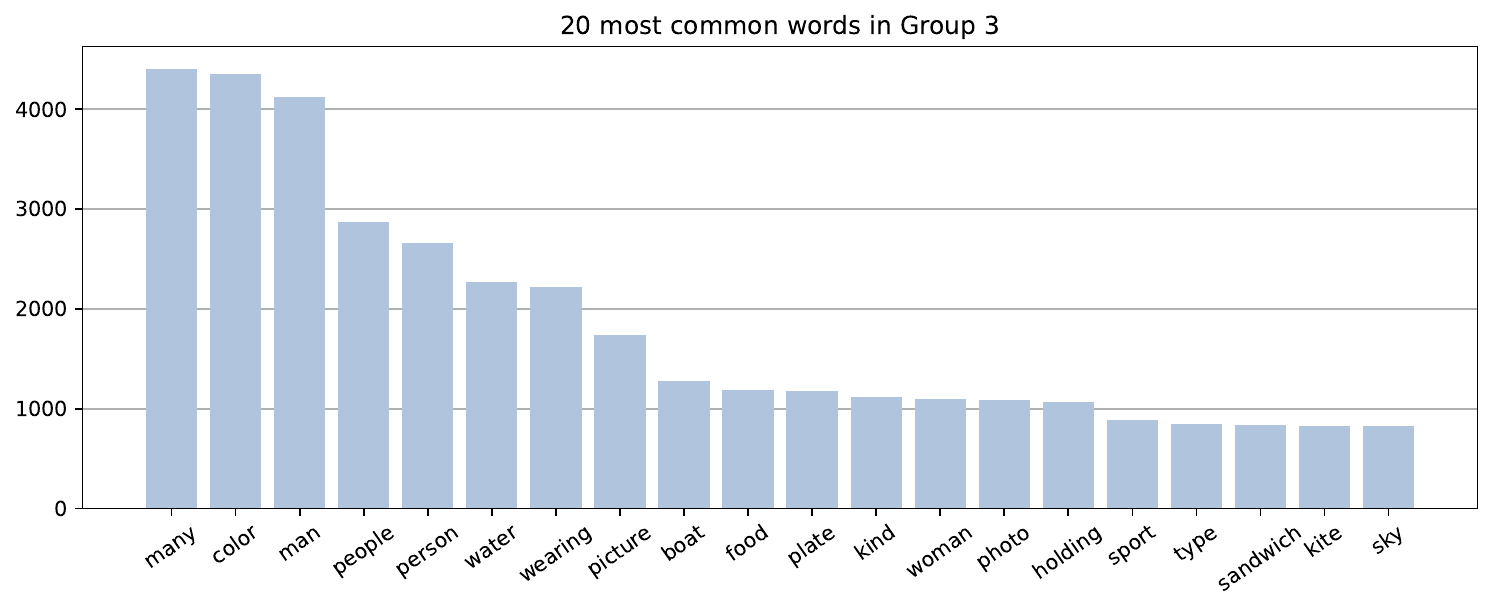}
     \end{subfigure}
     \hfill
      \begin{subfigure}[b]{0.48\textwidth}
         \centering
         \includegraphics[width=\textwidth]{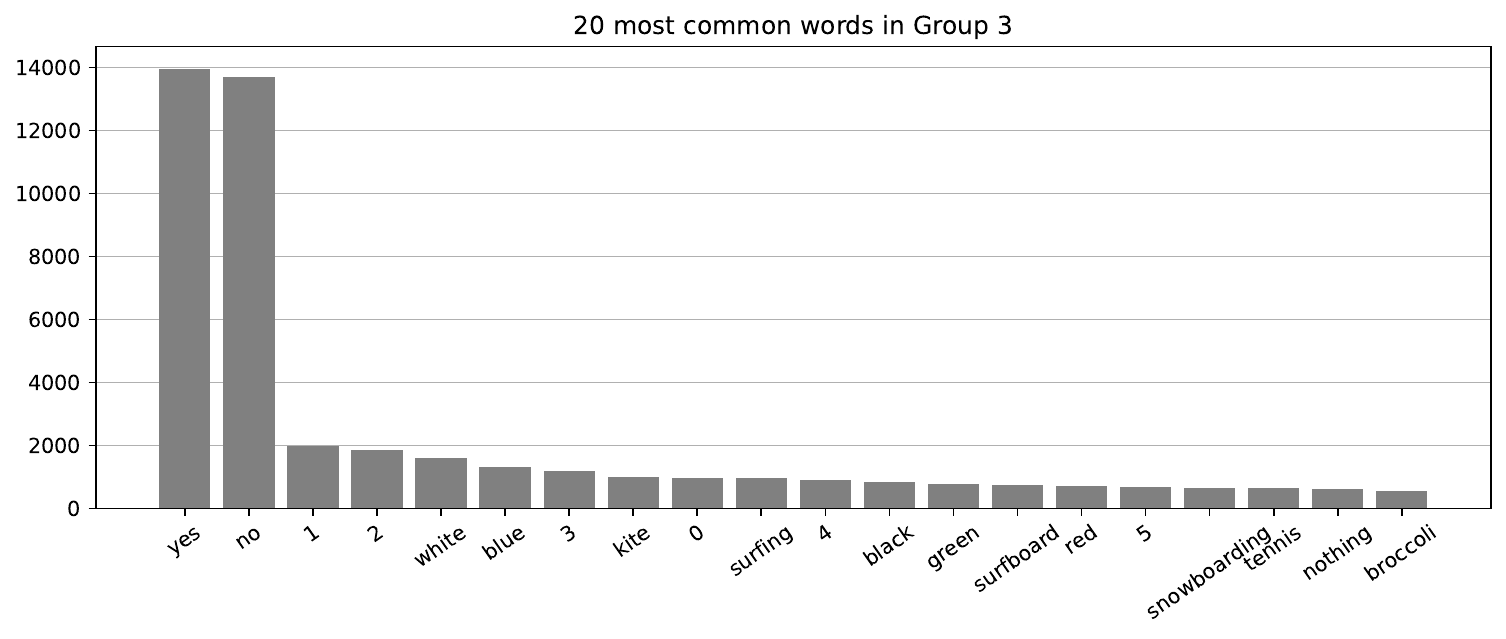}
     \end{subfigure}
     \hfill
     \begin{subfigure}[b]{0.48\textwidth}
         \centering
         \includegraphics[width=\textwidth]{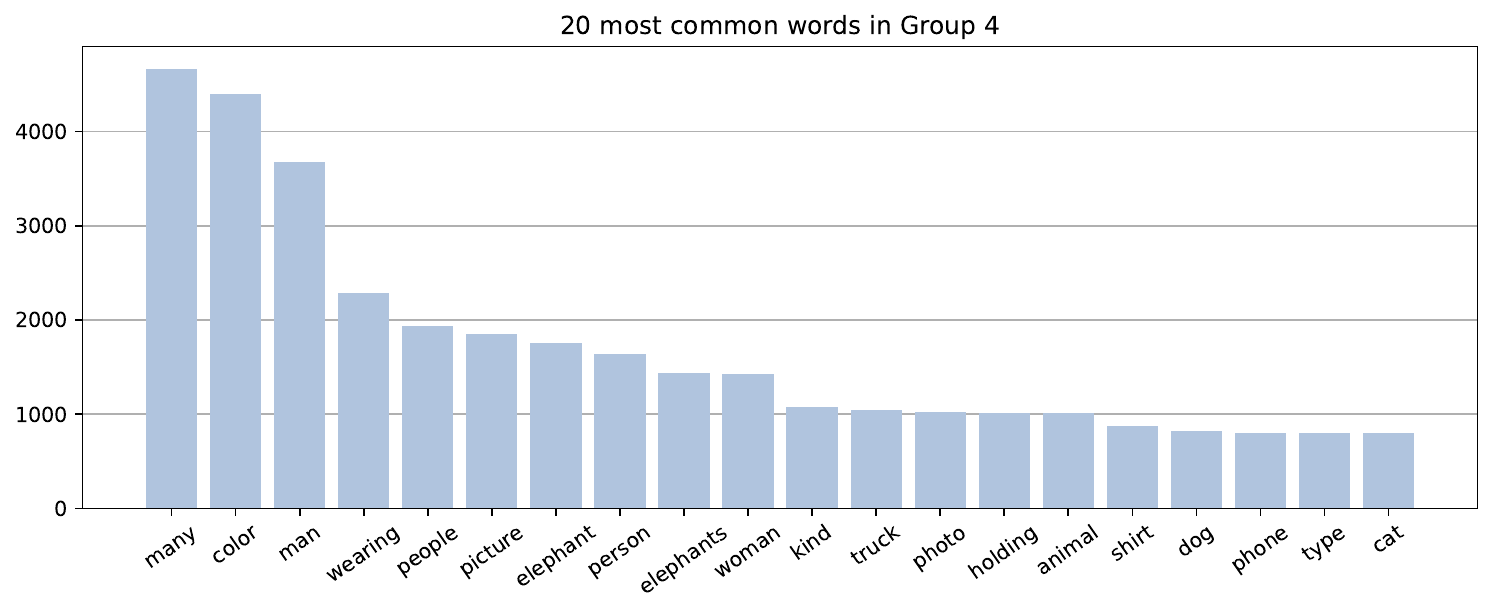}
     \end{subfigure}
     \hfill
     \begin{subfigure}[b]{0.48\textwidth}
         \centering
         \includegraphics[width=\textwidth]{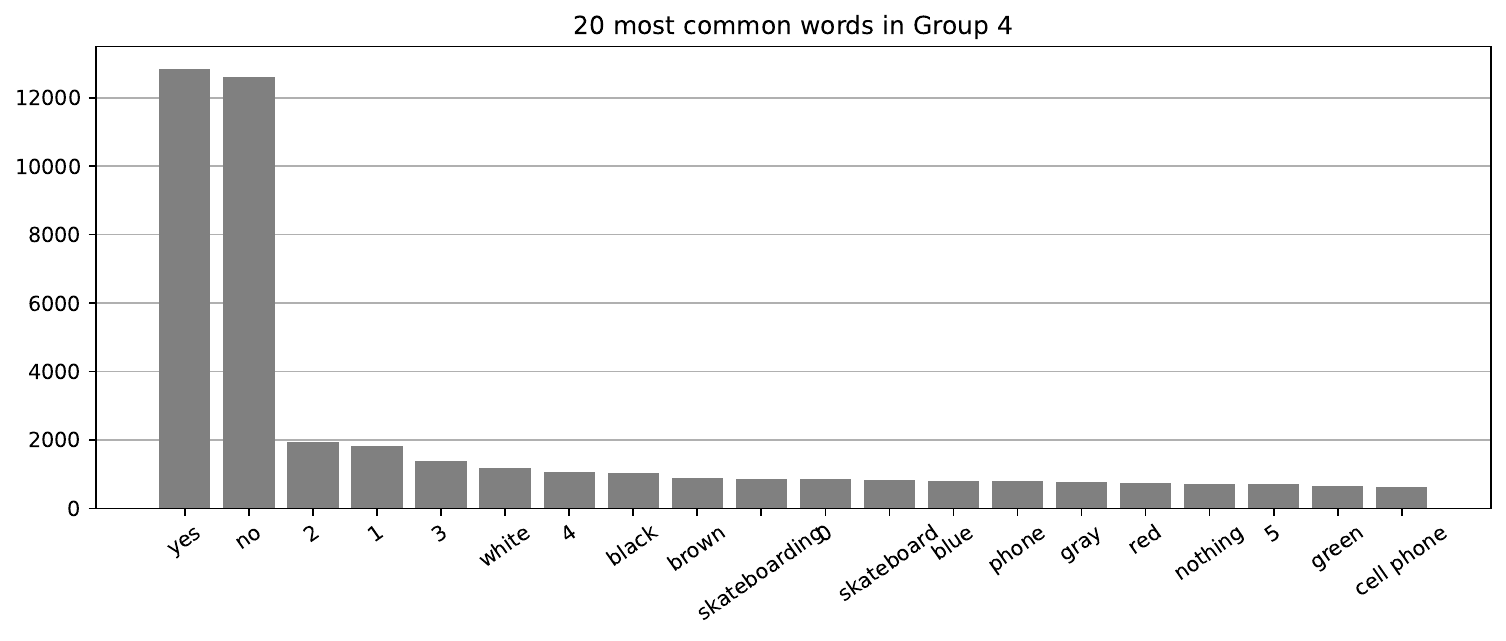}
     \end{subfigure}
     \hfill
     \begin{subfigure}[b]{0.48\textwidth}
         \centering
         \includegraphics[width=\textwidth]{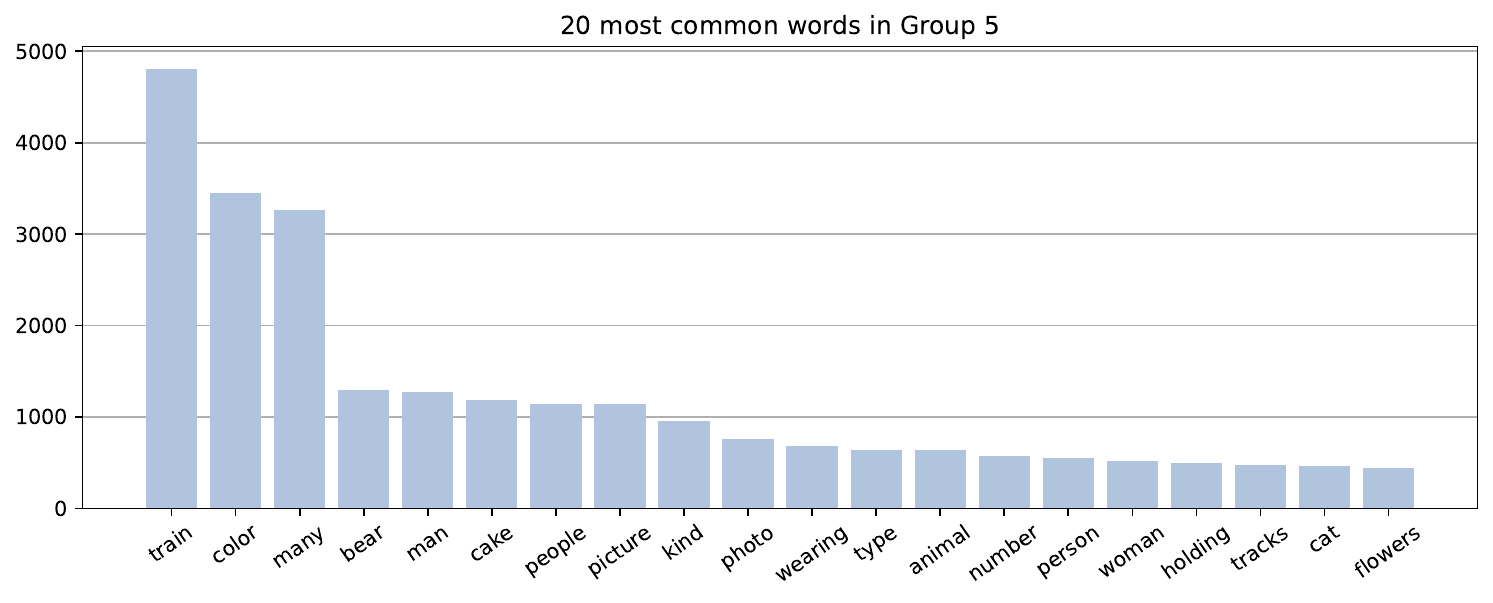}
     \end{subfigure}
     \hfill
     \begin{subfigure}[b]{0.48\textwidth}
         \centering
         \includegraphics[width=\textwidth]{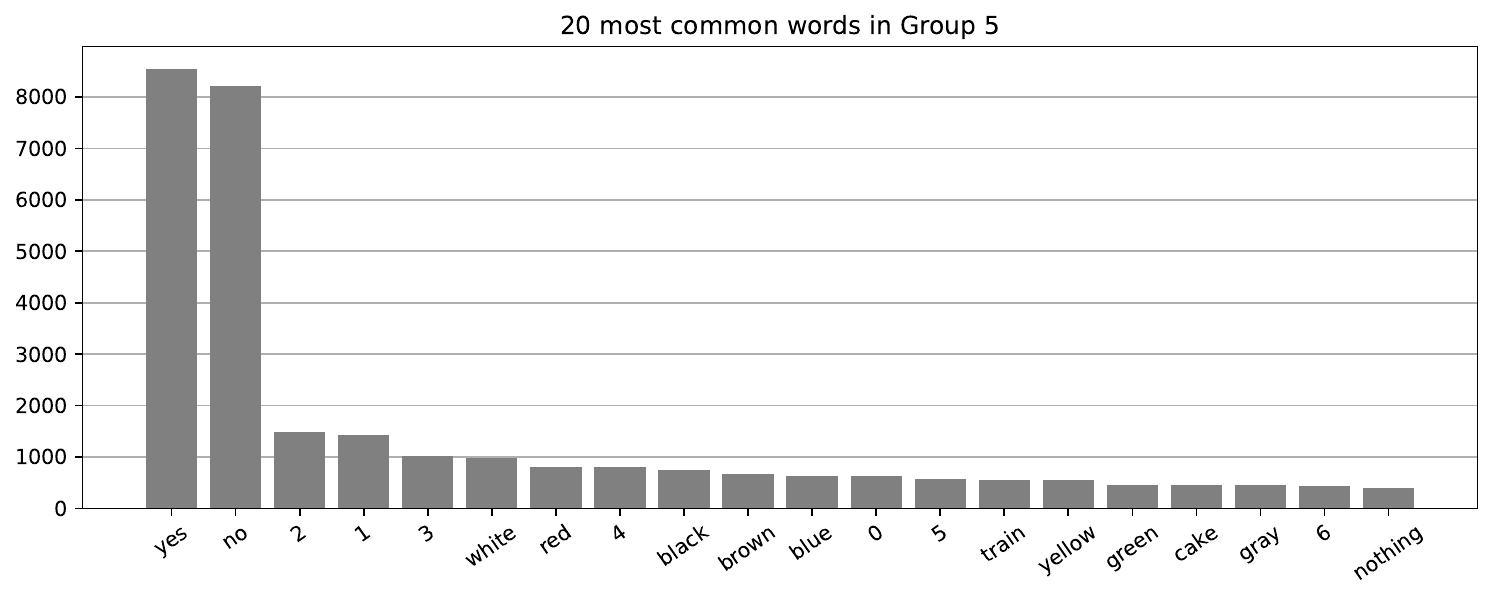}
     \end{subfigure}
    \caption{Most common words (left) and answers (right) per task Diverse Domains.}
    \label{fig:top_ans_d}
\end{figure*}

\begin{figure*}[ht!]
    \centering
     \begin{subfigure}[b]{0.48\textwidth}
         \centering
         \includegraphics[width=\textwidth]{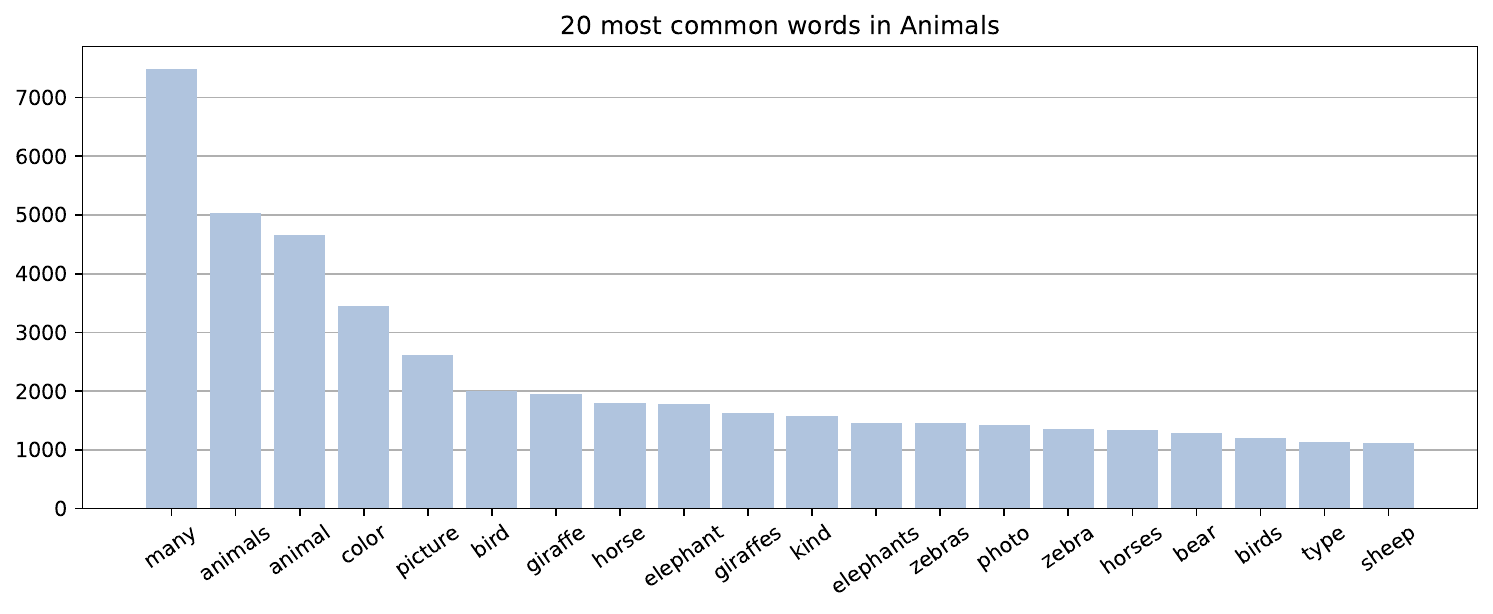}
     \end{subfigure}
     \hfill
     \begin{subfigure}[b]{0.48\textwidth}
         \centering
         \includegraphics[width=\textwidth]{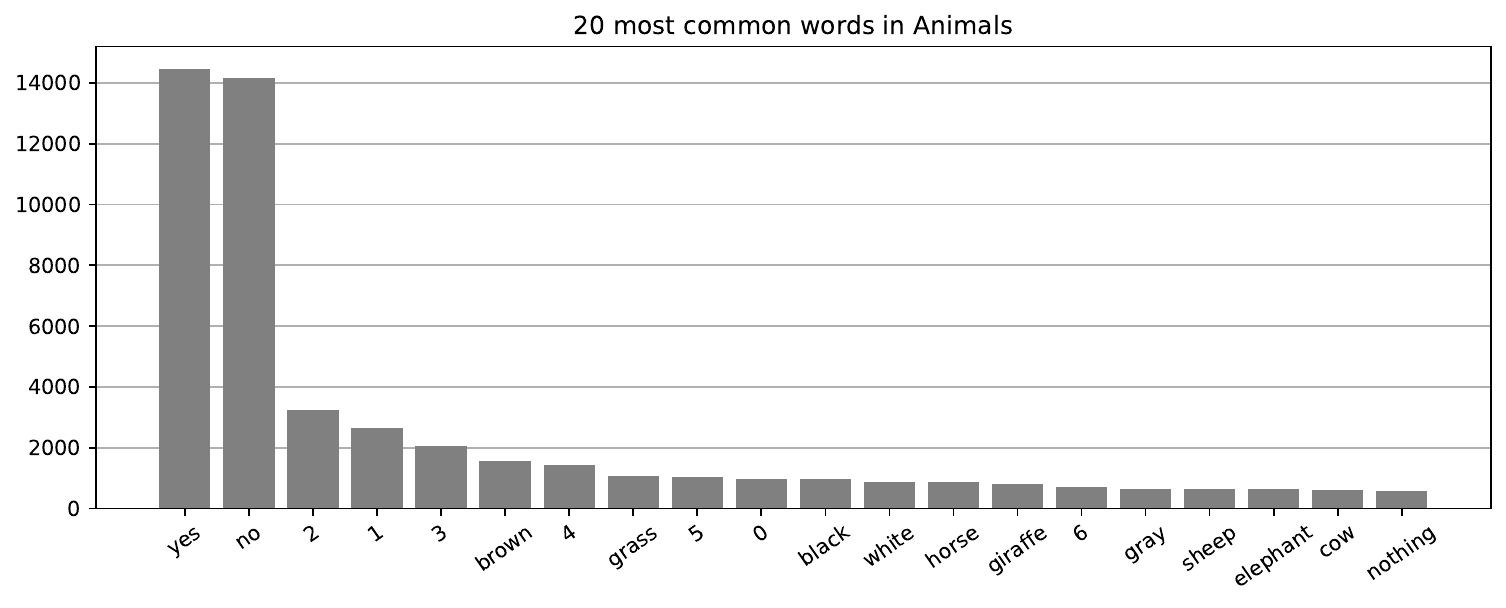}
     \end{subfigure}
     \hfill
     \begin{subfigure}[b]{0.48\textwidth}
         \centering
         \includegraphics[width=\textwidth]{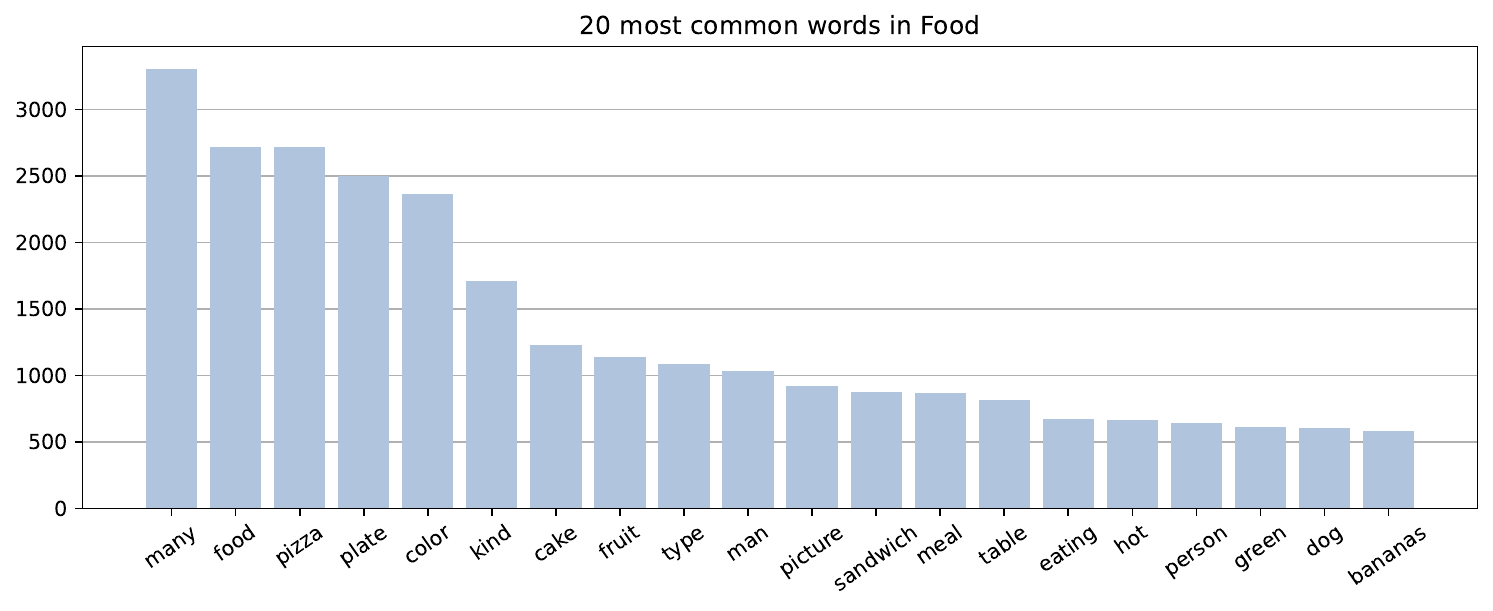}
     \end{subfigure}
     \hfill
     \begin{subfigure}[b]{0.48\textwidth}
         \centering
         \includegraphics[width=\textwidth]{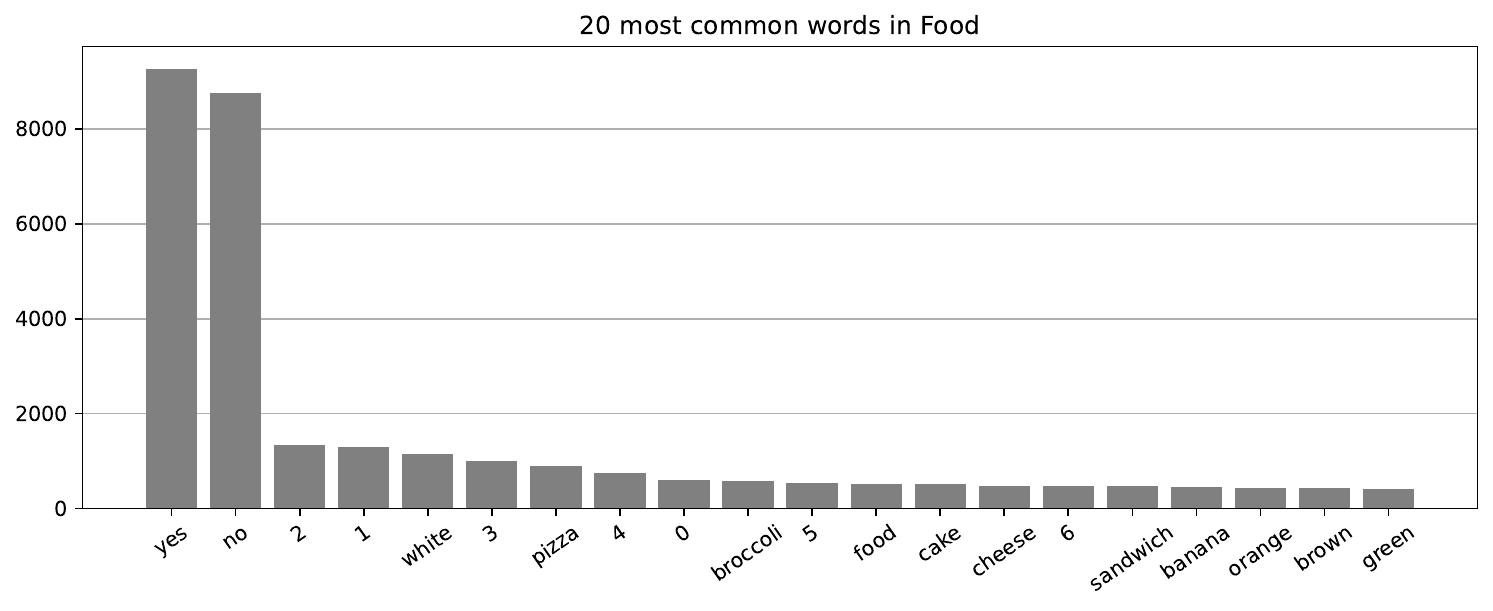}
     \end{subfigure}
     \hfill
     \begin{subfigure}[b]{0.48\textwidth}
         \centering
         \includegraphics[width=\textwidth]{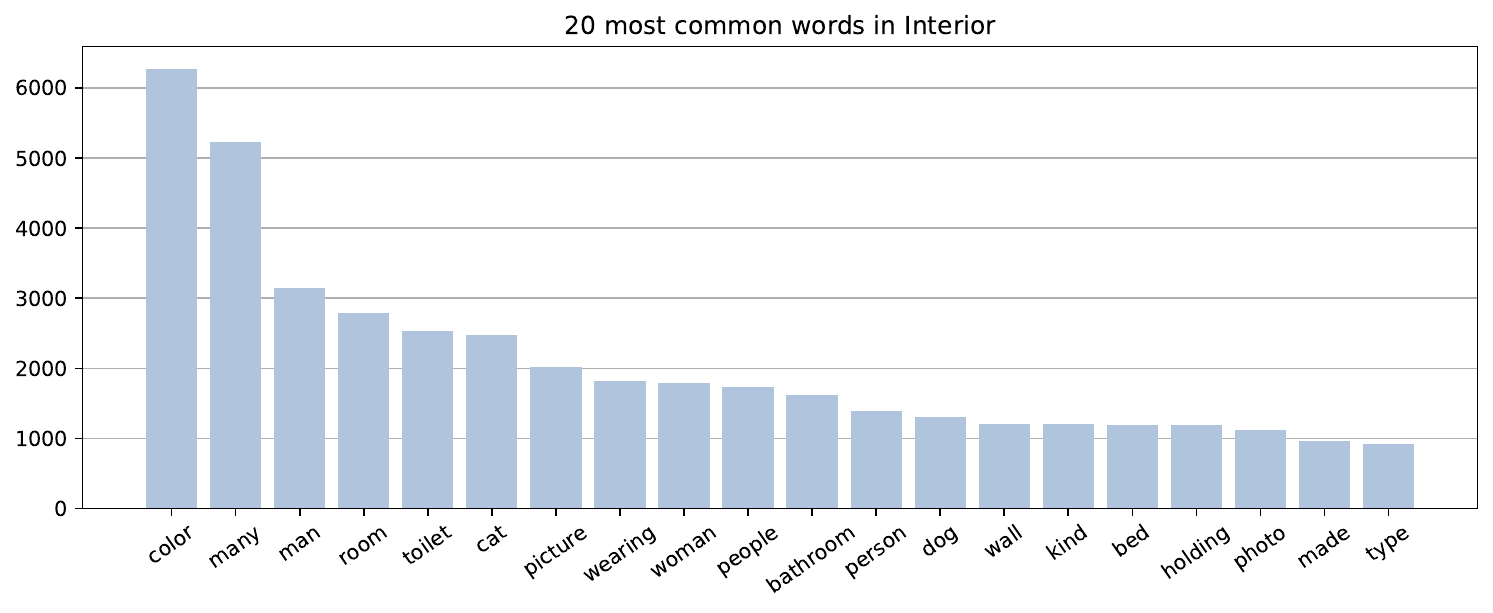}
     \end{subfigure}
     \hfill
      \begin{subfigure}[b]{0.48\textwidth}
         \centering
         \includegraphics[width=\textwidth]{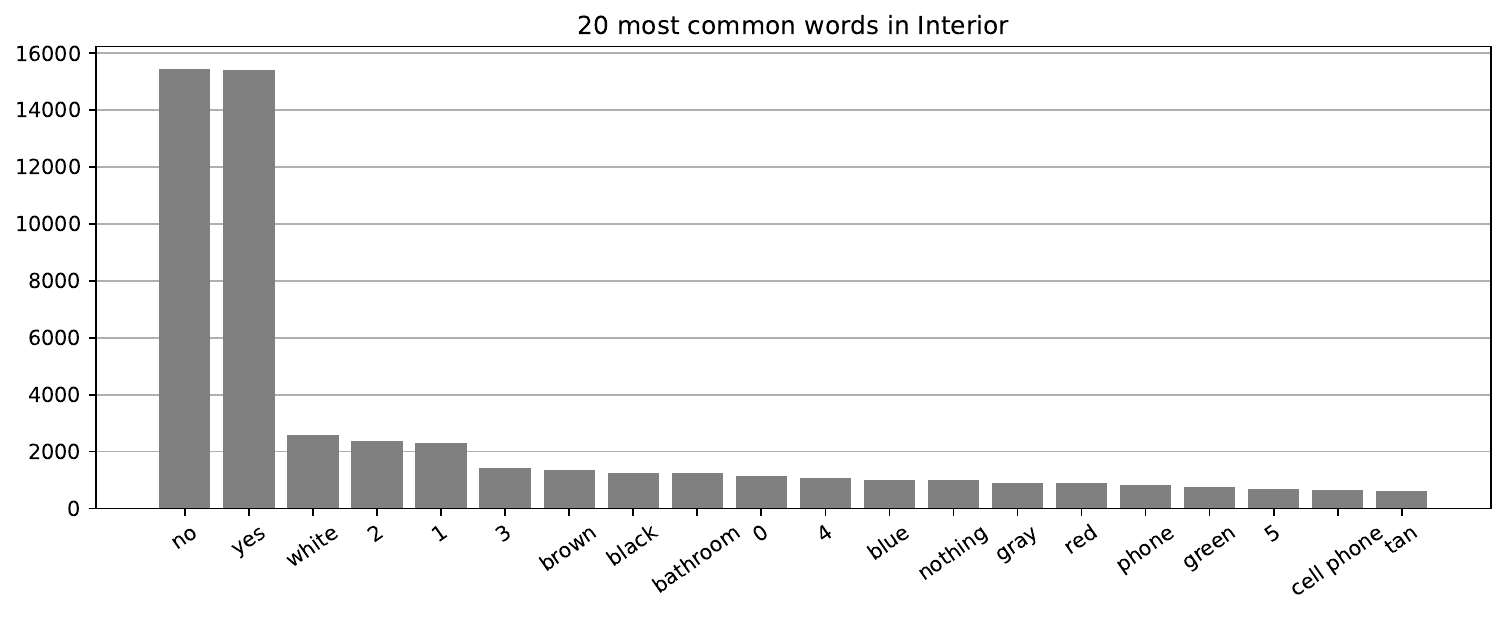}
     \end{subfigure}
     \hfill
     \begin{subfigure}[b]{0.48\textwidth}
         \centering
         \includegraphics[width=\textwidth]{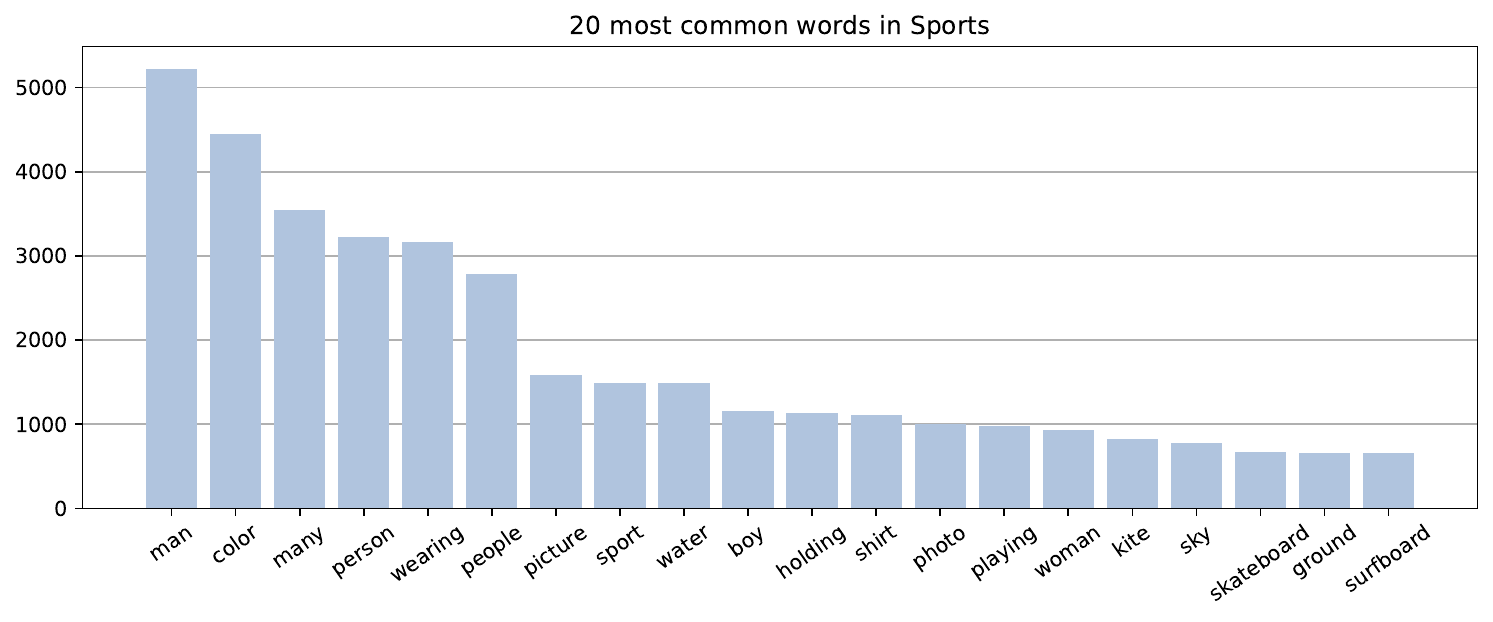}
     \end{subfigure}
     \hfill
     \begin{subfigure}[b]{0.48\textwidth}
         \centering
         \includegraphics[width=\textwidth]{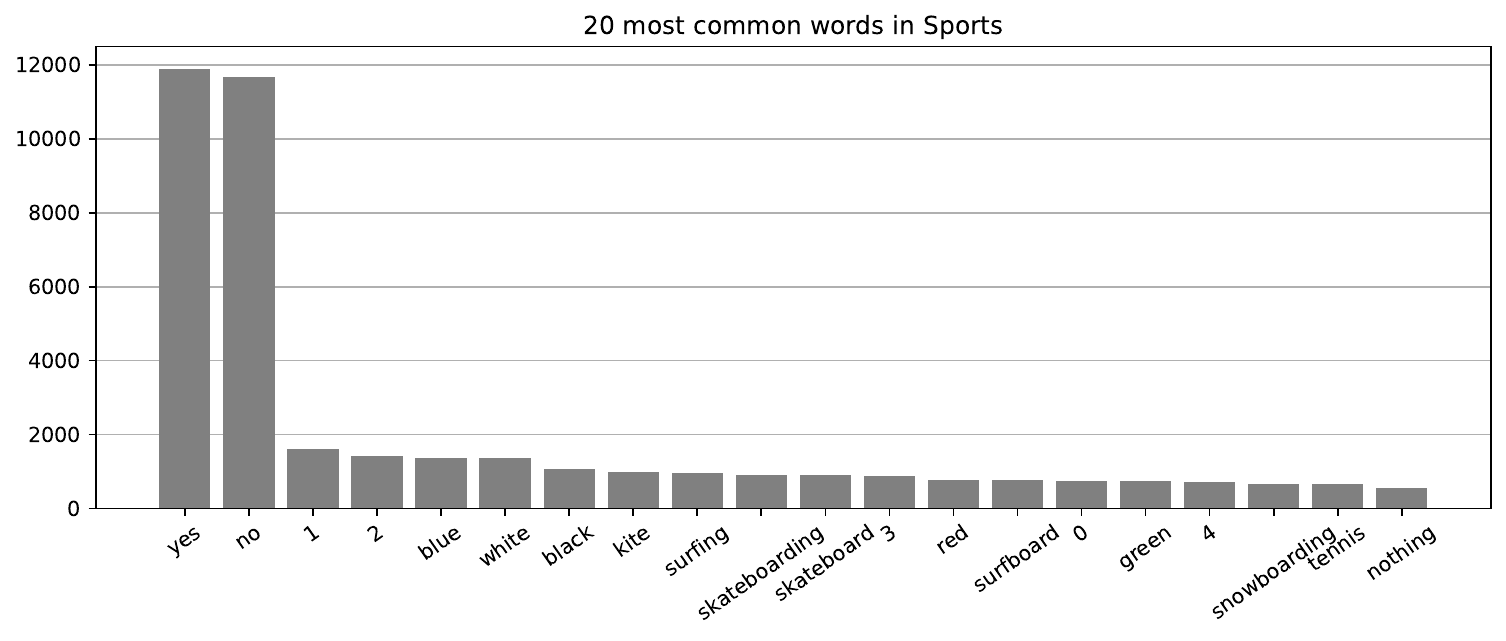}
     \end{subfigure}
     \hfill
     \begin{subfigure}[b]{0.48\textwidth}
         \centering
         \includegraphics[width=\textwidth]{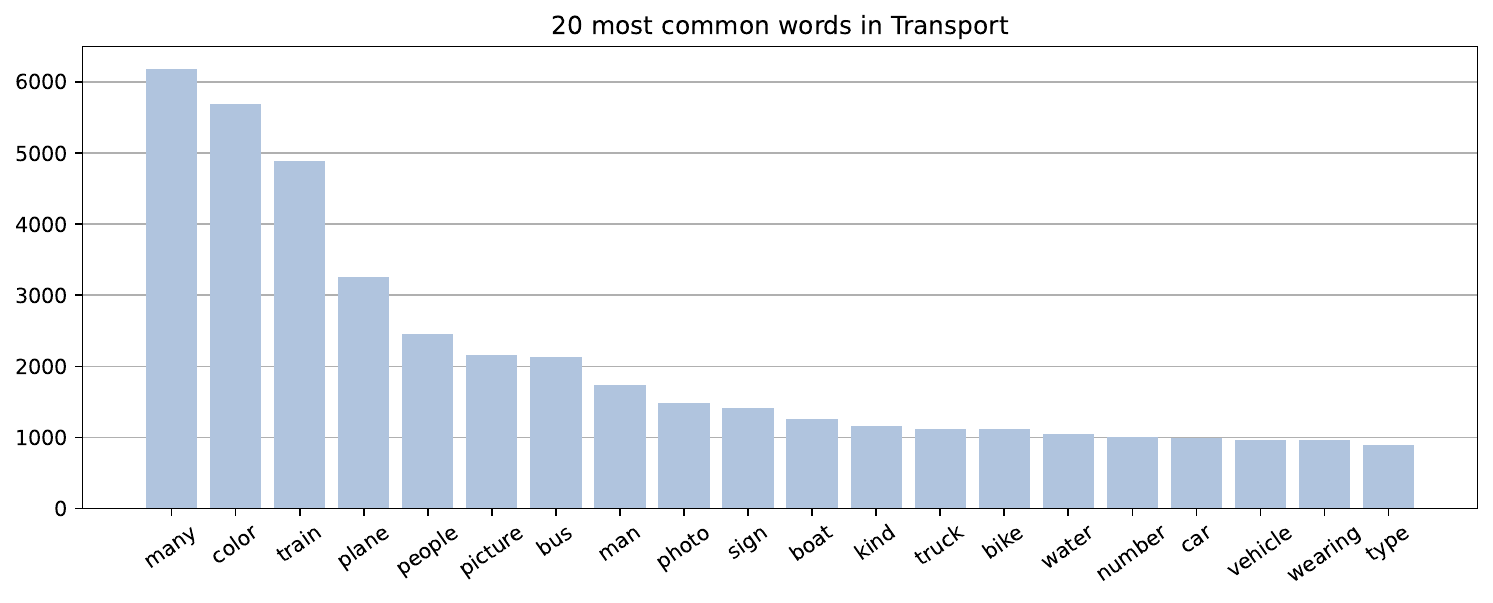}
     \end{subfigure}
     \hfill
     \begin{subfigure}[b]{0.48\textwidth}
         \centering
         \includegraphics[width=\textwidth]{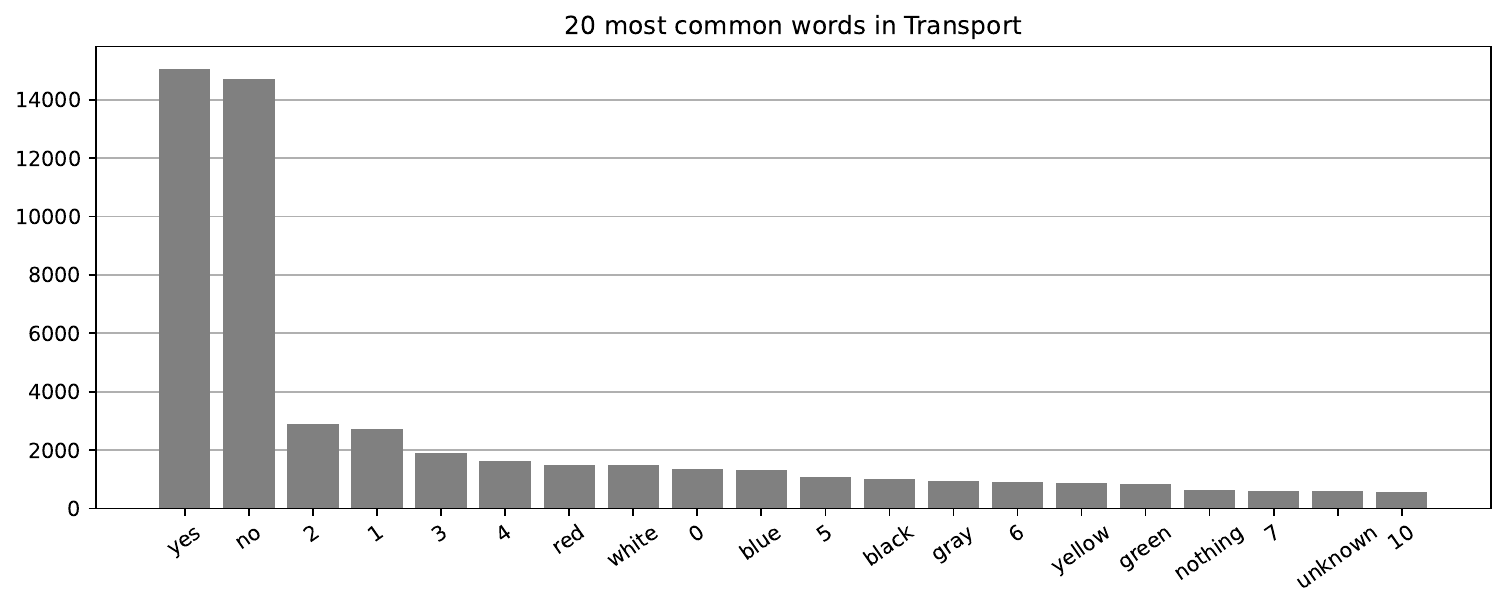}
     \end{subfigure}
    \caption{Most common words (left) and answers (right) per task Taxonomy Domains.}
    \label{fig:top_ans_t}
\end{figure*}

\begin{figure*}[ht!]
    \centering
     \begin{subfigure}[b]{0.48\textwidth}
         \centering
         \includegraphics[width=\textwidth]{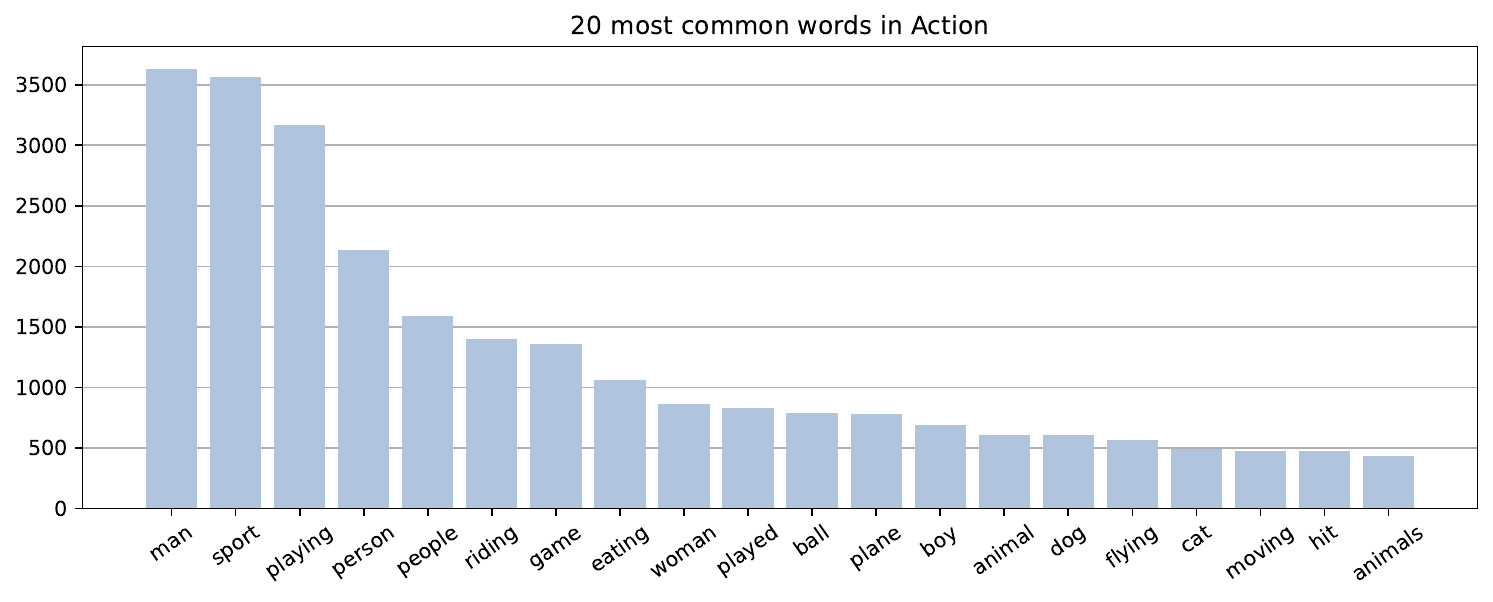}
     \end{subfigure}
     \hfill
     \begin{subfigure}[b]{0.48\textwidth}
         \centering
         \includegraphics[width=\textwidth]{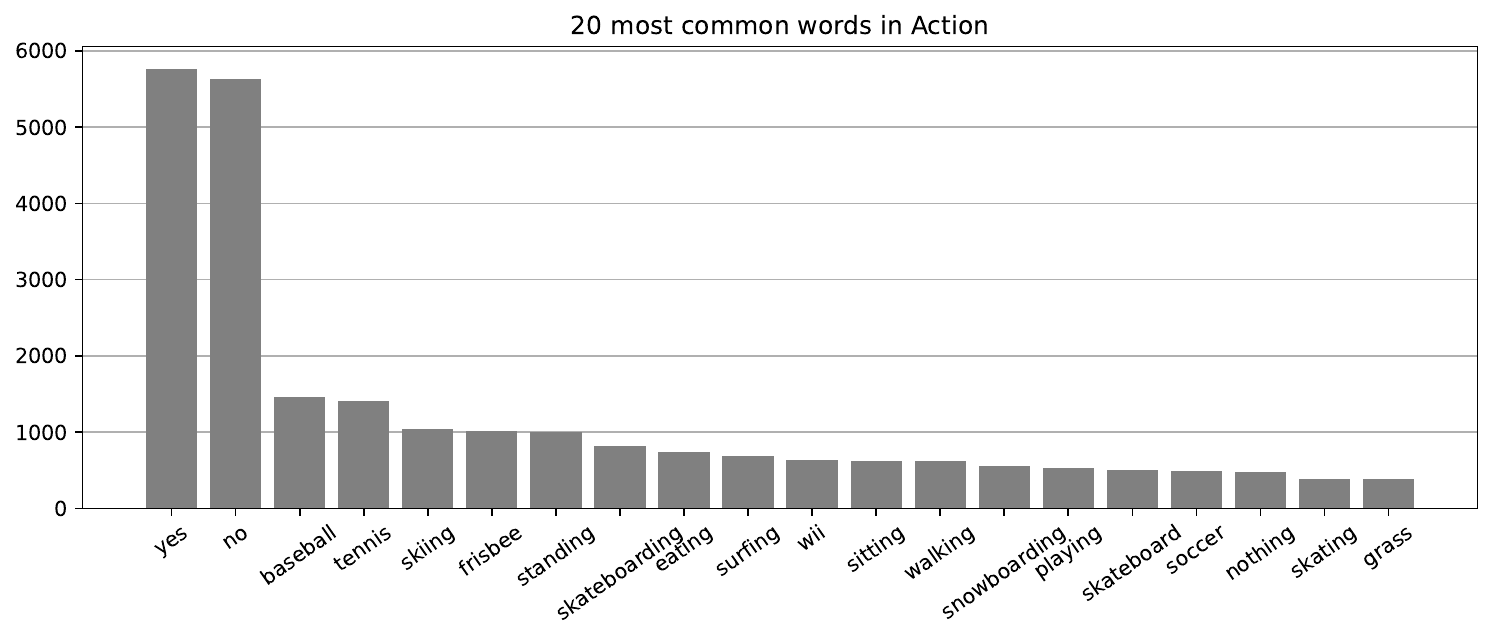}
     \end{subfigure}
     \hfill
     \begin{subfigure}[b]{0.48\textwidth}
         \centering
         \includegraphics[width=\textwidth]{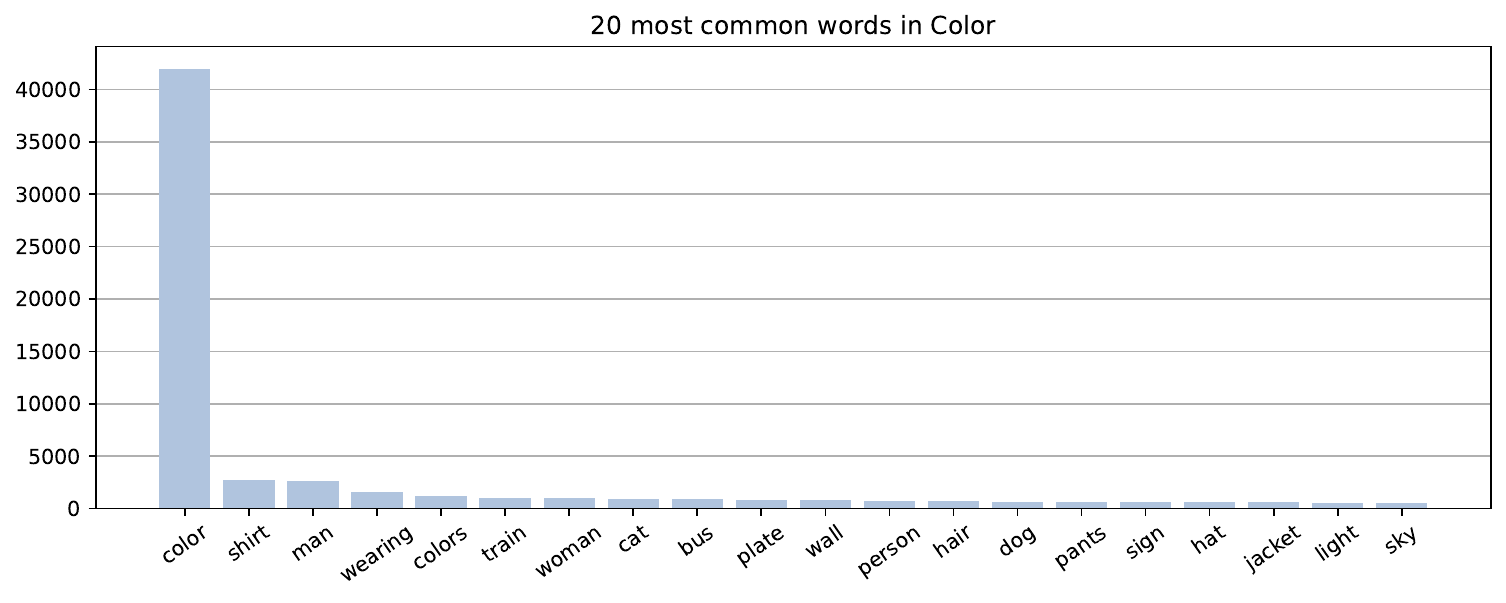}
     \end{subfigure}
     \hfill
     \begin{subfigure}[b]{0.48\textwidth}
         \centering
         \includegraphics[width=\textwidth]{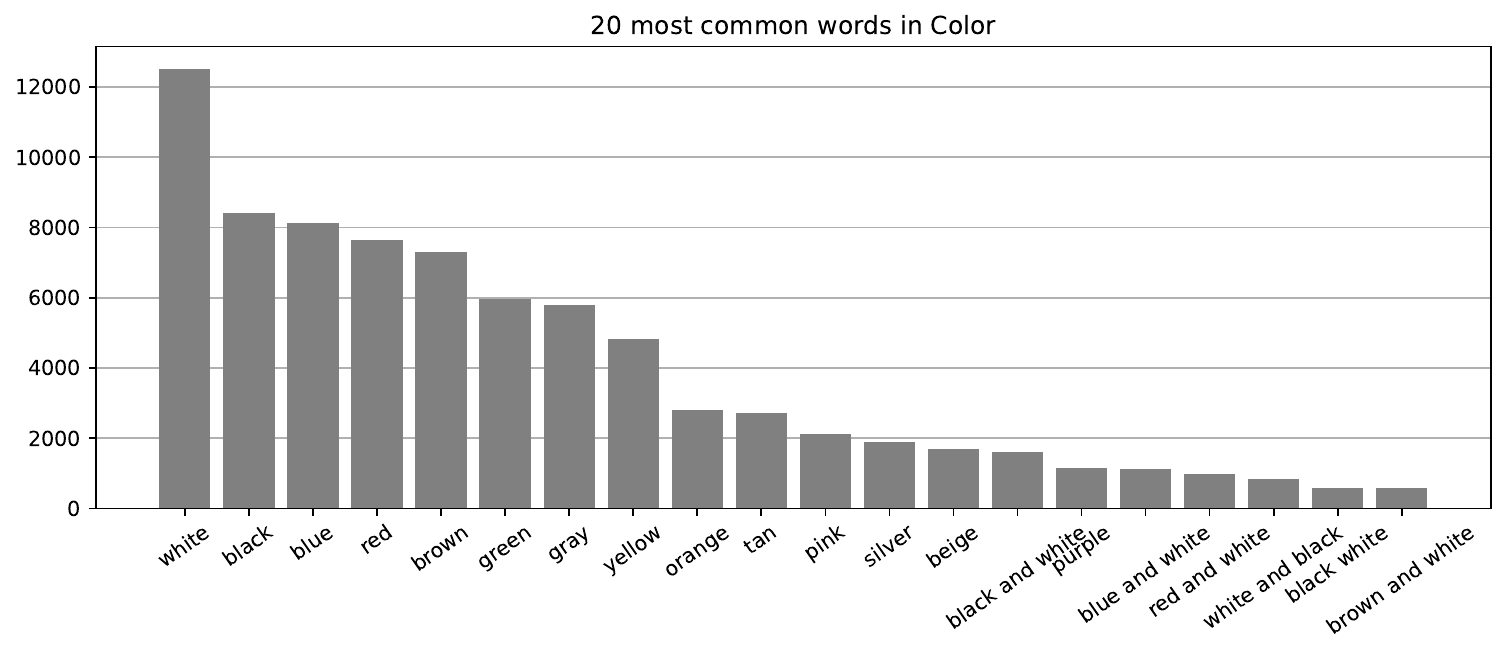}
     \end{subfigure}
     \hfill
     \begin{subfigure}[b]{0.48\textwidth}
         \centering
         \includegraphics[width=\textwidth]{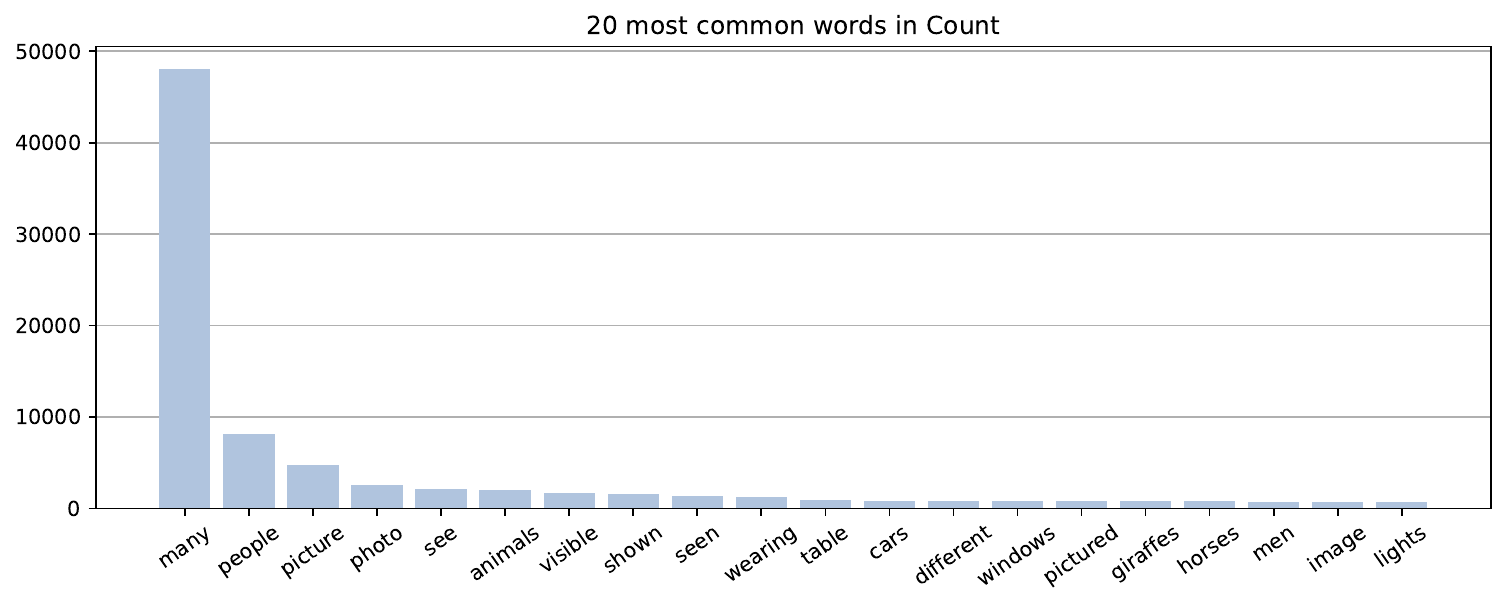}
     \end{subfigure}
     \hfill
     \begin{subfigure}[b]{0.48\textwidth}
         \centering
         \includegraphics[width=\textwidth]{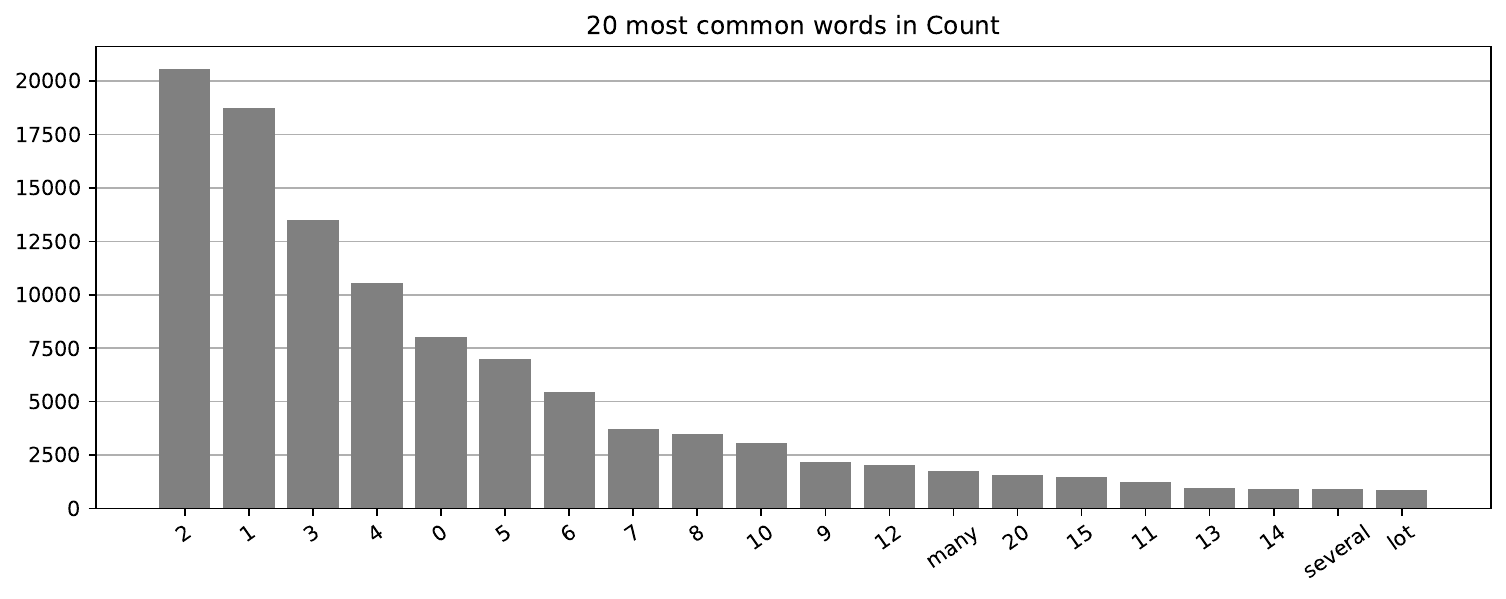}
     \end{subfigure}
     \hfill
     \begin{subfigure}[b]{0.48\textwidth}
         \centering
         \includegraphics[width=\textwidth]{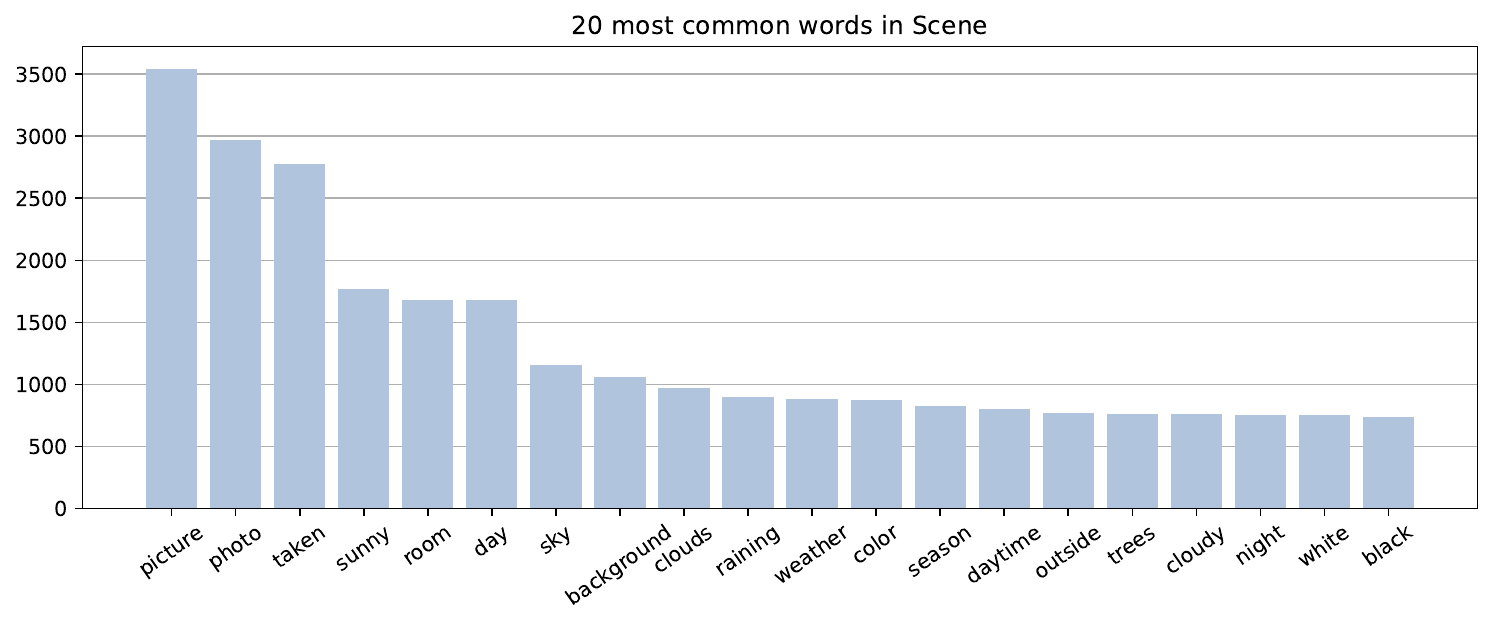}
     \end{subfigure}
     \hfill
     \begin{subfigure}[b]{0.48\textwidth}
         \centering
         \includegraphics[width=\textwidth]{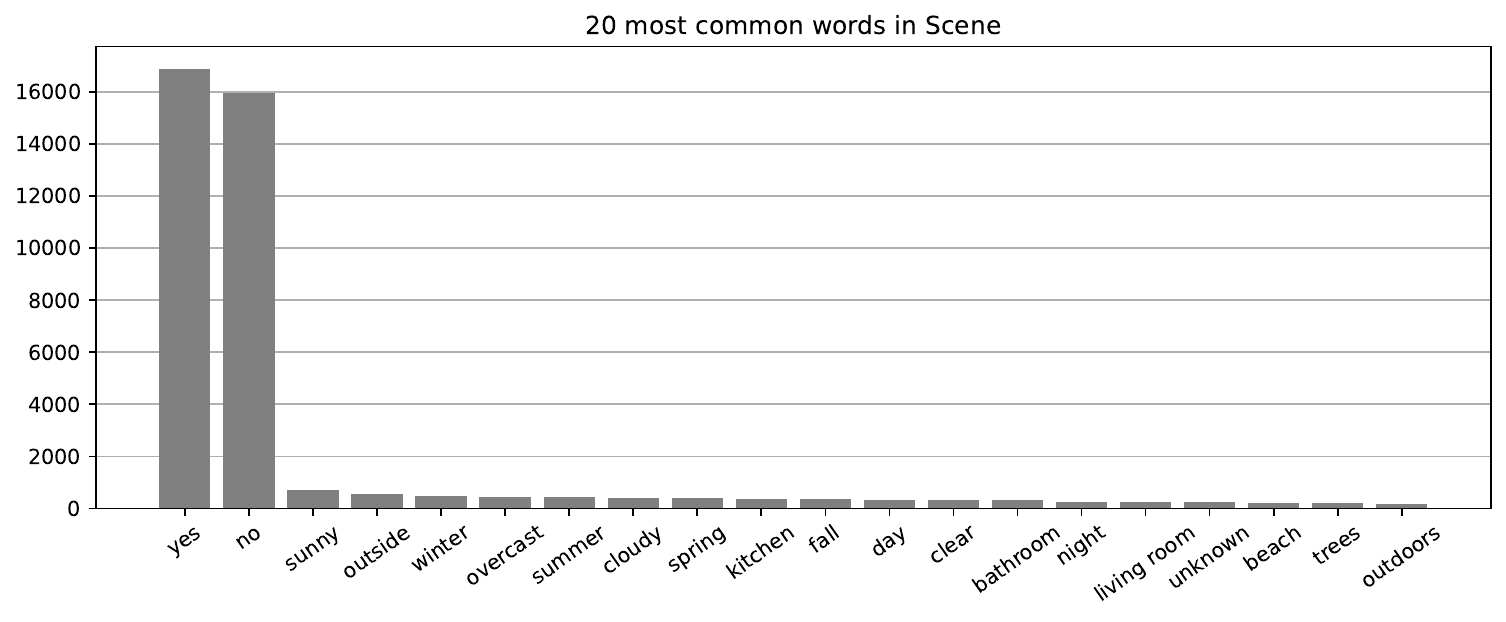}
     \end{subfigure}
     \hfill
     \begin{subfigure}[b]{0.48\textwidth}
         \centering
         \includegraphics[width=\textwidth]{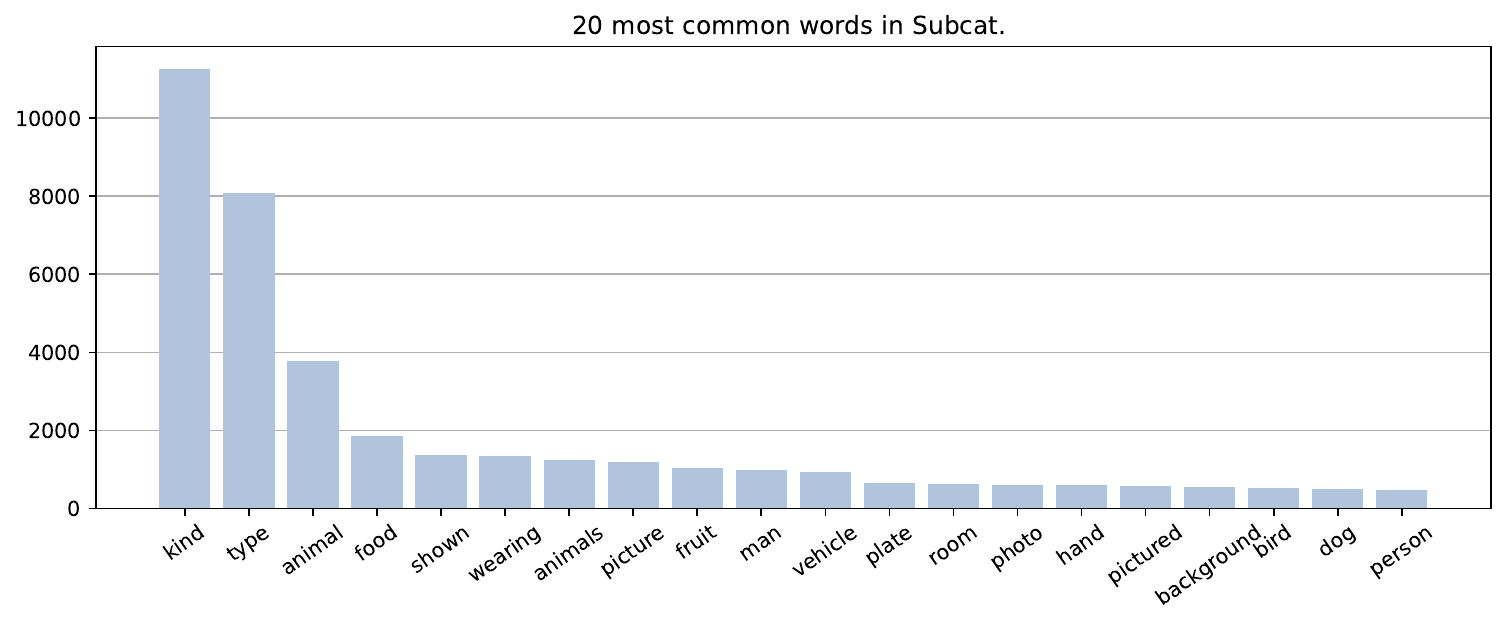}
     \end{subfigure}
     \hfill
     \begin{subfigure}[b]{0.48\textwidth}
         \centering
         \includegraphics[width=\textwidth]{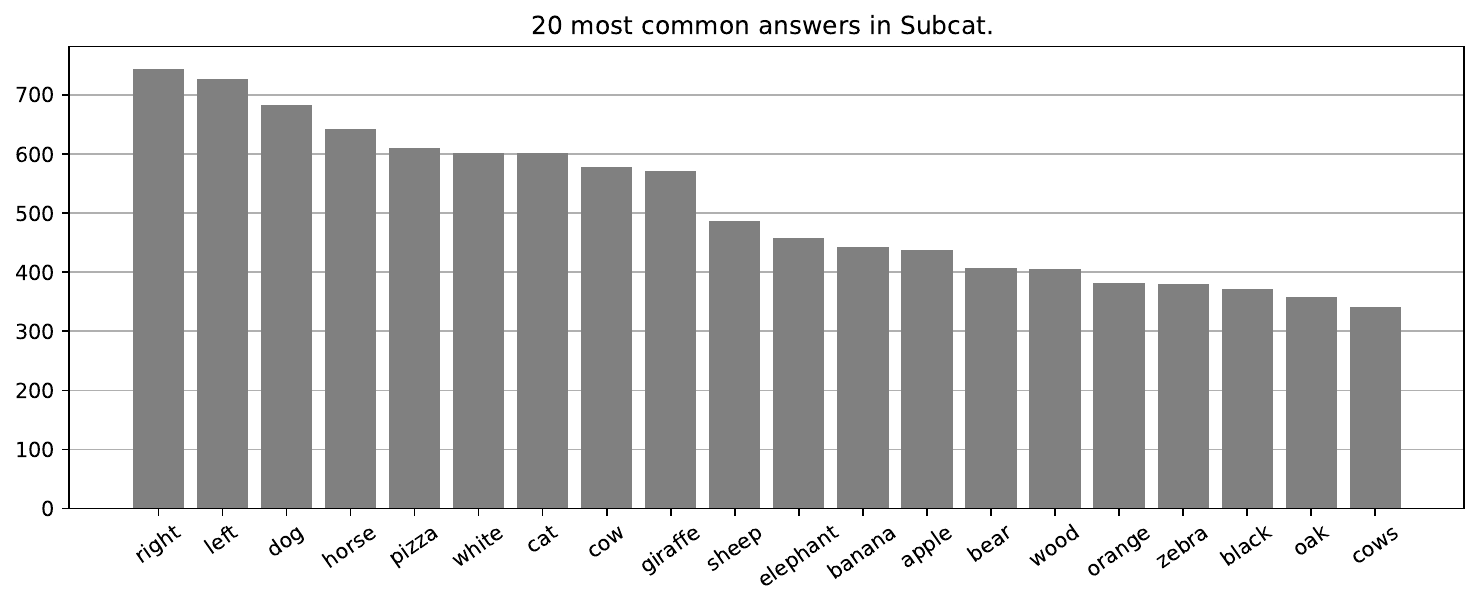}
     \end{subfigure}
    \caption{Most common words (left) and answers (right) per task in Question Types.}
    \label{fig:top_ans_q}
\end{figure*}

Figures~\ref{fig:top_ans_d}-\ref{fig:top_ans_q} show the distribution of the 20 most common question words and answers for each task. The counts are computed on the combined train and validation data, excluding stopwords from the question vocabulary.
These plots support our general findings about the characteristics of each task and the relationships between them. For example, answers in Diverse Domains are highly similar across tasks, while the most considerable difference of common answers is observed in Question Types.
In addition, frequent nouns in Diverse and Taxonomy Domains reflect the typical objects from the image annotations of each task.
Common words in Question Types also  follow the definition of each task. For example, top words in Scene such as `sunny', `room', `outside' refer to the entire image, while Action words such as `sport', `playing', `moving' refer to activities shown in the image. 

\begin{table}[]
    \centering
    \resizebox{\columnwidth}{!}{%

    \begin{tabular}{lcccccc}\hline
     \textbf{Dissimilarity} & \textbf{Diverse} & \textbf{Taxonomy} & \textbf{Questions}  \\\hline
        Answers  & \underline{0.567} (0.009) & \underline{0.791} (0.000) & \underline{0.795} (0.000) \\
        Image embed. &  0.248 (0.293) & \underline{0.492} (0.028) & \underline{-0.640} (0.002)) \\
        Question embed. & 0.184 (0.437) & \underline{0.531} (0.016) & \underline{0.631} (0.003) \\
        Joint embed. & 0.220 (0.350) &   \underline{0.622} (0.003) & -0.223 (0.344) \\\hline
    \end{tabular}}
    \caption{Spearman correlation of pairwise performance drop and and different dissimilarity heuristics. In addition to the results in Table~\ref{tab:spearman}, we show in parentheses the corresponding p-values. We underline statistically significant results (p < 0.05).}
    \label{tab:spearman_appendix}
\end{table}

\begin{figure*}[ht!]
    \centering
         \begin{subfigure}[b]{0.9\textwidth}
         \includegraphics[width=\textwidth]{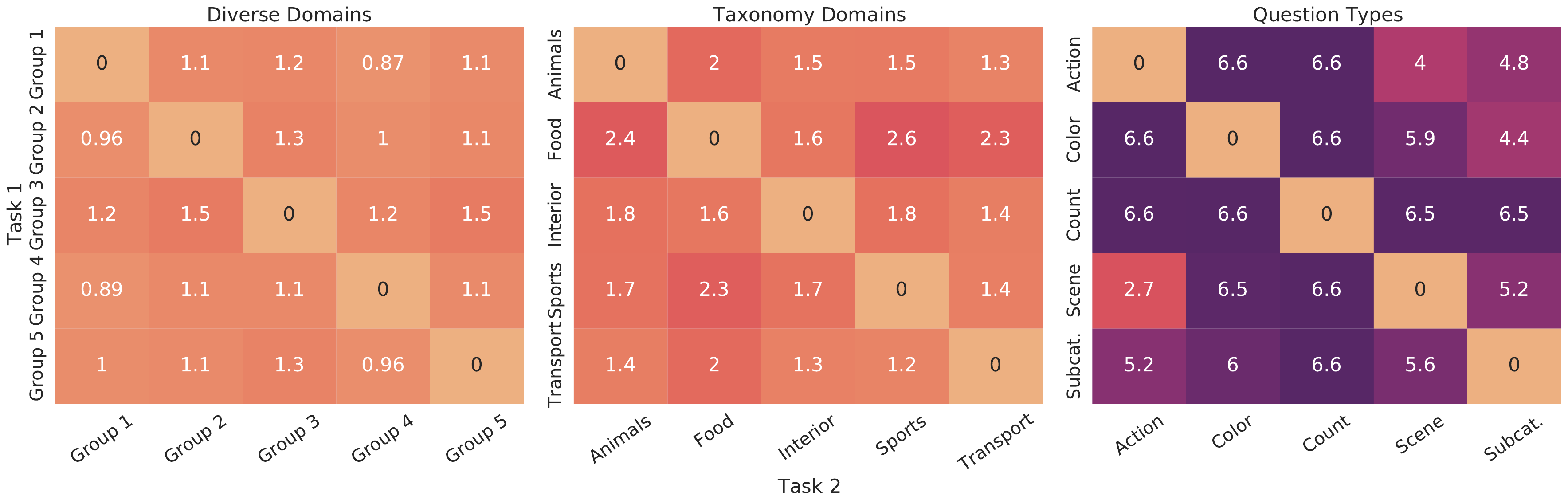}
         \caption{Divergence of answer distributions.}
    \end{subfigure}
    \\[15pt]
    \begin{subfigure}[b]{0.9\textwidth}         \includegraphics[width=\textwidth]{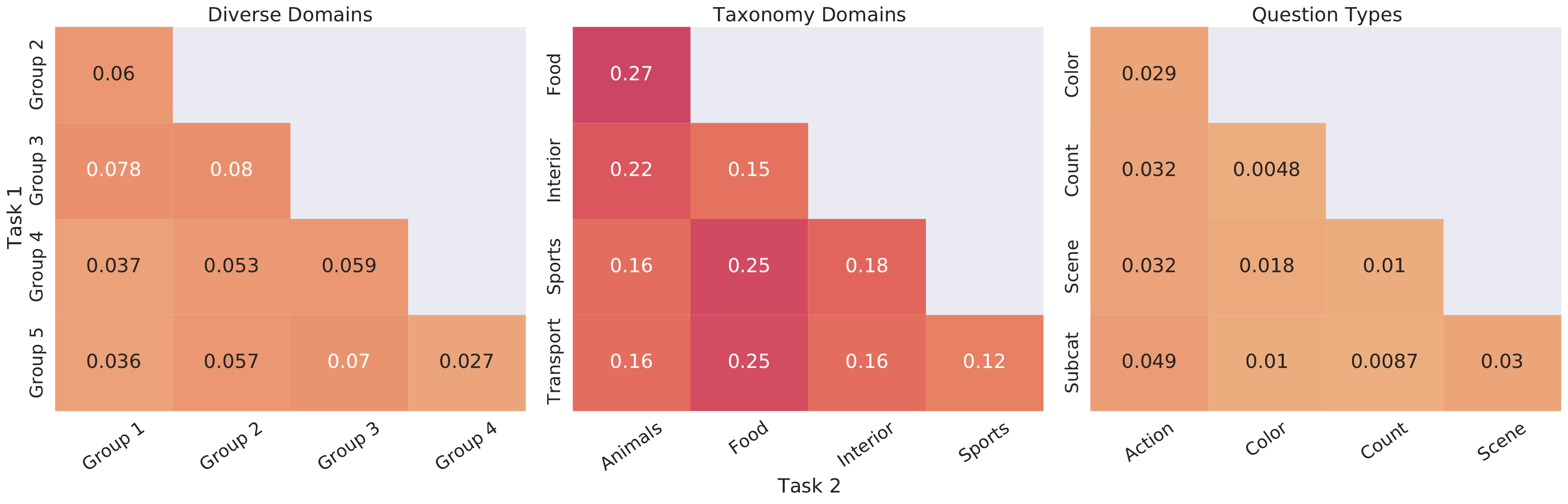}
     \caption{Cosine distance of image embeddings.}
    \end{subfigure}
     \hfill    \\[15pt]

     \begin{subfigure}[b]{0.9\textwidth}
         \includegraphics[width=\textwidth]{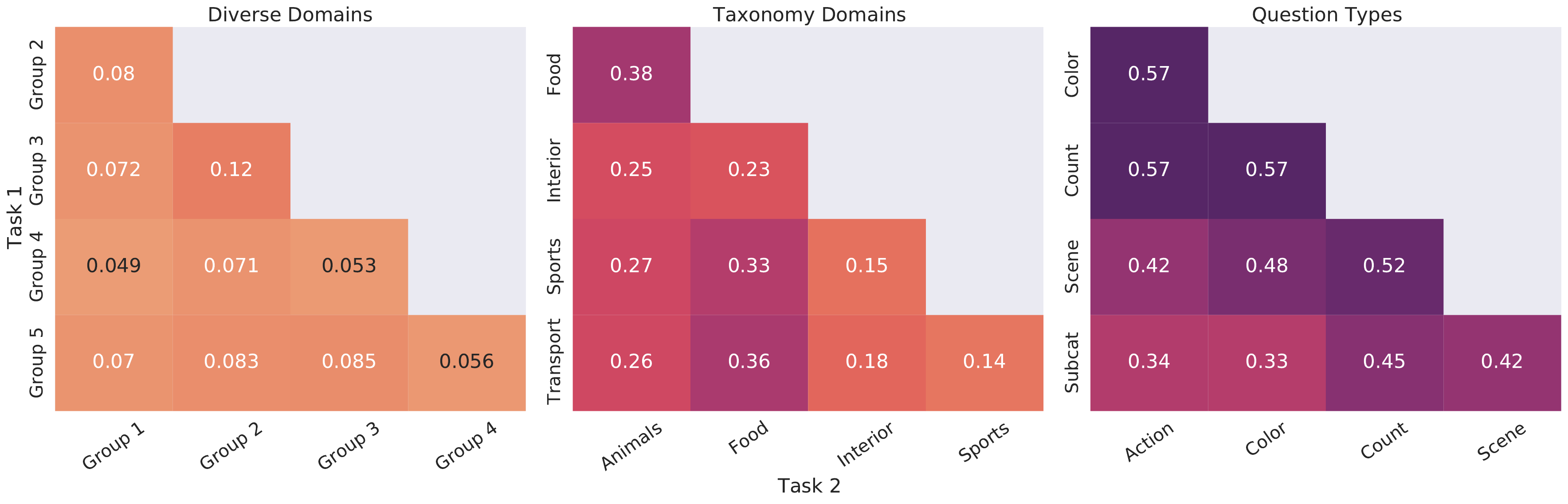}
         \caption{Cosine distance of question embeddings.}
    \end{subfigure}
    \hfill    \\[15pt]

        \begin{subfigure}[b]{0.9\textwidth}
         \includegraphics[width=\textwidth]{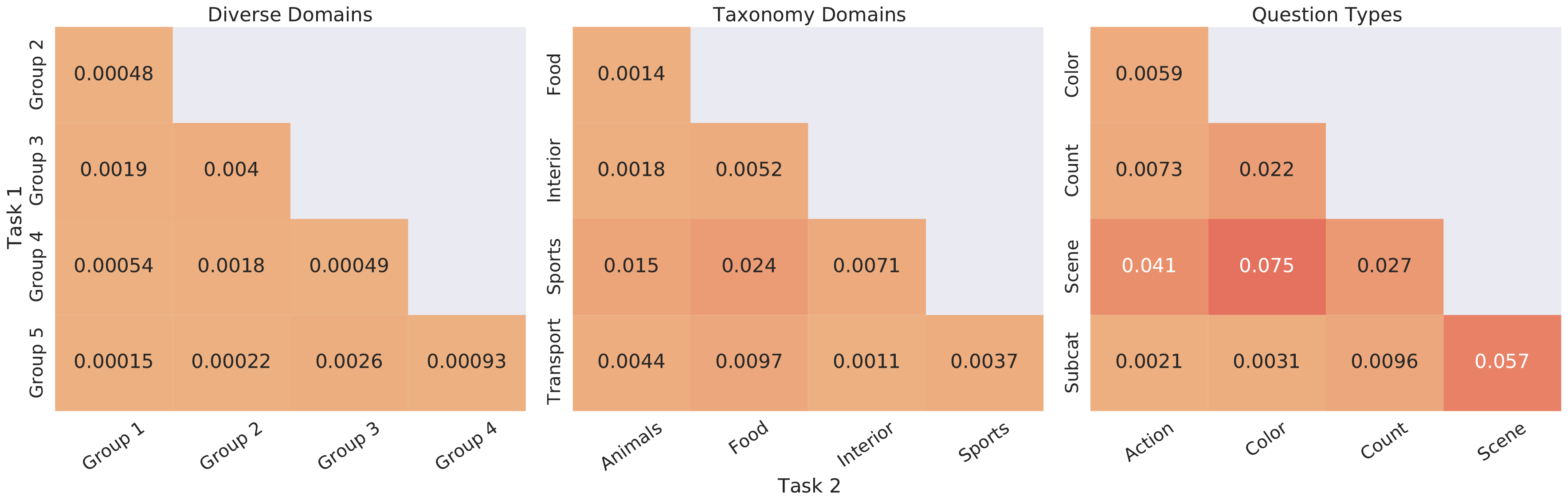}
        \caption{Cosine distance of joint embeddings.}
    \end{subfigure}
    \caption{Dissimilarity measures between task pairs.}
    \label{fig:dissimilarity}
\end{figure*}

\section{Implementation Details} \label{sec:implementation_details}
Our implementation is based on the publicly available PyTorch codebase of UNITER (\url{https://github.com/ChenRocks/UNITER}).
For the continual learning experiments, we train a UNITER-base model (86M parameters) on a cluster of NVIDIA V100 GPUs using a single node with 4 GPUs.
Training on a sequence of 5 tasks requires on average $\sim$ 5 GPU hours. The main experiments (Table~\ref{tab:results}) require approximately a total of 200 GPU hours.

We first tune the batch size and learning rate with naive finetuning.
Following previous work on finetuning V+L models, we downscale the learning rate of the pretrained backbone by 10x. 
Keeping these hyperparameters fixed, we then tune the continual learning hyperparameters (EWC, LwF $\lambda$).
All hyperparameters are selected through grid search based on the maximum final accuracy as shown in Table ~\ref{tab:hyperparams}.
Initial results with a pretrained model on Taxonomy Domains showed that best performance is achieved with a mixing ratio of 3:1 of new and old data per batch. We keep this ratio constant for all experiments.

\begin{table}[]
    \centering
   \resizebox{\columnwidth}{!}{
 
    \begin{tabular}{llcccc}\hline
 & \textbf{Setting} & \textbf{Batch Size} & \textbf{Learning Rate} & \textbf{LwF $\lambda$} & \textbf{EWC $\lambda$} \\\hline
\multirow{6}{1em}{\rotatebox[origin=c]{90}{UNITER}}   &  Diverse & 512 & 8e-5 & 1 & 400 \\
  &  Diverse+PT & 1024 & 8e-5  & 0.7 & 500 \\
  &  Taxonomy & 512 & 8e-5 & 1 & 600 \\
  &  Taxonomy+PT & 1024 & 5e-5 & 0.5 & 500 \\
  &  Questions & 1024 & 1e-4 & 0.9 & 50K  \\
  &  Questions+PT & 512 & 5e-5 & 0.4 & 20K \\\hline
\multirow{3}{1em}{\rotatebox[origin=c]{90}{ViLT}}  & Diverse+PT  & 1024 & 1e-5 & - & 500\\
  & Taxnomy+PT & 1024 & 1e-5 & - & 700\\
  & Questions+PT & 512 & 8e-5 & - & 10K \\ \hline
   \end{tabular}}
   \caption{Best hyperparameters for all settings. PT: Initialization from pretrained checkpoint.}
    \label{tab:hyperparams}
\end{table}

Each experiment is repeated five times with a different random seed and task order. The task orders used in our experiments are the following:

\begin{itemize}[leftmargin=*]
 \setlength\itemsep{1pt}
    \item \textbf{Diverse Domains}
    \item group 5, group 3, group 2, group 4, group 1
    \item group 1, group 2, group 5, group 3, group 4
    \item group 4, group 3, group 5, group 1, group 2
    \item group 3, group 1, group 4, group 2, group 5
    \item group 2, group 5, group 1, group 4, group 3
    \item \textbf{Taxonomy Domains}
    \item food, animals, sports, interior, transport
    \item transport, sports, food, animals, interior
    \item interior, animals, food, transport, sports
    \item animals, food, interior, sports, transport
    \item sports, interior, transport, animals, food
    \item \textbf{Question types}
    \item action, count, subcategory, scene, color
    \item color, subcategory, action, count, scene
    \item scene, count, action, color, subcategory
    \item subcategory, color, scene, action, count
    \item count, scene, color, subcategory, action
\end{itemize}

\section{ViLT Results}
Table~\ref{tab:results_vilt} shows the detailed results for ViLT across the three settings.
\begin{table*}[ht]
\centering
\small
\resizebox{0.6\textwidth}{!}{%
\begin{tabular}{clcccc}
\hline 
 & & & & &  \\[-7pt]
\textbf{Split} & \textbf{Method} & \textbf{Accuracy} & \textbf{LA} & \textbf{BWT} & \textbf{SBWT}\\ \hline
 & & & & &  \\[-7pt]
\multirow{4}{1em}{\rotatebox[origin=c]{90}{Diverse}} & Fixed Model & 51.64 \tiny{\textcolor{gray}{$\pm$ 3.09}} & - & - & - \\
& Finetuning & 61.07 \tiny{\textcolor{gray}{$\pm$ 0.41}} & 65.03 \tiny{\textcolor{gray}{$\pm$ 1.06}} & -5.01 \tiny{\textcolor{gray}{$\pm$ 1.02}} & -2.80 \tiny{\textcolor{gray}{$\pm$ 0.68}}\\
& EWC & 61.80 \tiny{\textcolor{gray}{$\pm$ 0.96}} & 63.64 \tiny{\textcolor{gray}{$\pm$ 1.33}} & -2.30 \tiny{\textcolor{gray}{$\pm$ 0.62}} & -1.14 \tiny{\textcolor{gray}{$\pm$ 0.38}} \\
& ER & 64.22 \tiny{\textcolor{gray}{$\pm$ 0.10}} & 64.74 \tiny{\textcolor{gray}{$\pm$ 0.84}} & -0.98 \tiny{\textcolor{gray}{$\pm$ 0.52}} & -0.25 \tiny{\textcolor{gray}{$\pm$ 0.71}} \\
\rowcolor{gray!30}
& Joint &  67.51 \tiny{\textcolor{black}{$\pm$ 0.10}} & - & - & - \\
\hline
 & & & & &\\[-7pt]
\multirow{4}{1em}{\rotatebox[origin=c]{90}{Taxonomy}} & Fixed Model &  50.74 \tiny{\textcolor{gray}{$\pm$ 1.09}} & - & - & - \\
& Finetuning & 61.25 \tiny{\textcolor{gray}{$\pm$ 0.50}} & 66.51 \tiny{\textcolor{gray}{$\pm$ 0.27}} & -6.57 \tiny{\textcolor{gray}{$\pm$ 0.75}} & -4.09 \tiny{\textcolor{gray}{$\pm$ 0.41}}\\
& EWC & 63.69 \tiny{\textcolor{gray}{$\pm$ 0.46}} & 64.86 \tiny{\textcolor{gray}{$\pm$ 0.29}} & -1.46 \tiny{\textcolor{gray}{$\pm$ 0.40}} & -0.92 \tiny{\textcolor{gray}{$\pm$ 0.24}} \\
& ER & 63.52 \tiny{\textcolor{gray}{$\pm$ 0.20}} & 65.59 \tiny{\textcolor{gray}{$\pm$ 0.22}} & -2.59 \tiny{\textcolor{gray}{$\pm$ 0.30}} & -1.46 \tiny{\textcolor{gray}{$\pm$ 0.20}} \\
\rowcolor{gray!30}
& Joint &  67.84 \tiny{\textcolor{black}{$\pm$ 0.09}} & - & - & - \\
\hline
 & & & & &\\[-7pt]

\multirow{4}{1em}{\rotatebox[origin=c]{90}{Questions}} & Fixed Model & 23.84 \tiny{\textcolor{gray}{$\pm$ 8.20}} & - & - & -  \\
& Finetuning & 36.95 \tiny{\textcolor{gray}{$\pm$ 11.09}} & 71.06 \tiny{\textcolor{gray}{$\pm$ 0.11}} & -42.64 \tiny{\textcolor{gray}{$\pm$ 13.93}} & -32.86 \tiny{\textcolor{gray}{$\pm$ 14.25}} \\
& EWC & 60.25 \tiny{\textcolor{gray}{$\pm$ 2.86}} & 68.60 \tiny{\textcolor{gray}{$\pm$ 0.33}} & -10.45 \tiny{\textcolor{gray}{$\pm$ 3.52}} & -8.19 \tiny{\textcolor{gray}{$\pm$ 2.81}} \\
& ER & 65.61 \tiny{\textcolor{gray}{$\pm$ 0.76}} & 70.77 \tiny{\textcolor{gray}{$\pm$ 0.18}} & -6.45 \tiny{\textcolor{gray}{$\pm$ 1.17}} & -2.86 \tiny{\textcolor{gray}{$\pm$ 0.62}} \\
\rowcolor{gray!30}
& Joint &  72.41 \tiny{\textcolor{black}{$\pm$ 0.12}} & - & - & - \\
\hline
\end{tabular}}
\caption{\label{vilt} ViLT Results from VQA Incremental
Learning. We report the average and standard deviation over five random task orders. LA: Learned Accuracy, BWT: Backward Transfer, SBWT: Semantic Backward Transfer.}
\label{tab:results_vilt}
\end{table*}

\section{Qualitative Results}\label{appendix:qualitative}
Table~\ref{tab:examples} shows examples of predicted answers with different approaches. The two top examples are from two different task orders in Question Types, and the two bottom examples are from Taxonomy Domains. 
The model trained from scratch (column w/o PT) fails to retain knowledge from the corresponding training task.
The pretrained model (column PT) is more resistant to forgetting and we observe that for the first and third images, it even manages to recover the correct answer during the training sequence.  
However, relying only on pretraining is insufficient, as the model still tends to change the predicted answer based on the most recent training task. 
Both EWC and ER combined with pretraining successfully retain previous knowledge.

\begin{table*}[!ht]
    \centering
    \begin{tabular}{cc}\hline
    \\[-10pt]
    \includegraphics[height=3cm, width=2cm]{Images/example1.jpg} &    
    \resizebox{0.45\textwidth}{!}{
    \begin{tabular}[b]{lcccc}
    \multicolumn{5}{c}{\large{What is the horse doing?}} \\[5pt]\hline
      Task & w/o PT & PT & PT+EWC & PT+ER \\ \hline
      \textbf{Action} & jumping & jumping & jumping & jumping \\
      Count & two & one & jumping & jumping \\
      Subcat. & riding & jump & jumping & jumping \\
      Scene & cold & jumping & jumping & jumping \\ 
      Color & black & black & jumping & jumping \\
    \end{tabular}} \\\hline
        \\[-10pt]
    \includegraphics[height=3cm, width=2.3cm]{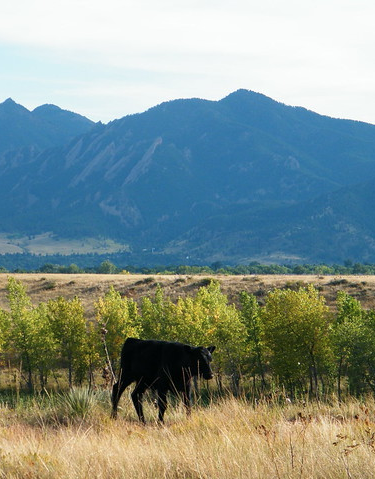} &    
    \resizebox{0.45\textwidth}{!}{
    \begin{tabular}[b]{lcccc}        
        \multicolumn{5}{c}{\large{What color is the cow?}} \\[5pt]\hline
      Task & w/o PT & PT & PT+EWC & PT+ER \\ \hline
      \textbf{Color} & black & black & black & black \\
      Subcat & black & black & black & black \\
      Action & zero & yes & cow & black \\
      Count & one & one & black & black \\ 
      Scene & green & green & black & black \\
    \end{tabular}} \\\hline
        \\[-10pt]
    \includegraphics[height=3cm, width=2.3cm]{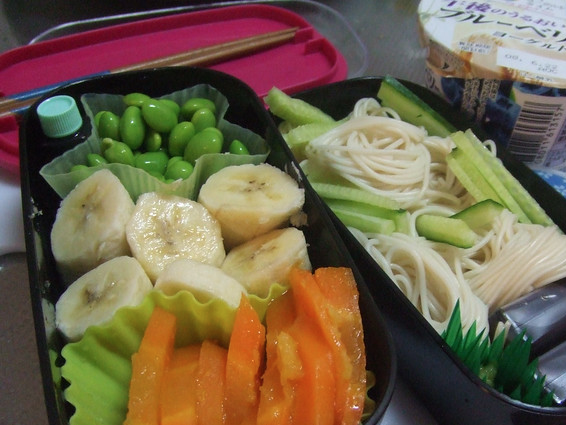} &    
    \resizebox{0.45\textwidth}{!}{
    \begin{tabular}[b]{lcccc}
    \multicolumn{5}{c}{\large{What is orange?}}\\[5pt]  \hline
      Task & w/o PT & PT & PT+EWC & PT+ER \\ \hline
    \textbf{Food} & carrots & carrots & carrots & carrots \\      
      Animals & birds & carrots & carrots &  carrots \\
      Sports & nothing & kites & carrots & carrots \\
     Interior & chair & carrots & carrots & carrots  \\
      Transport & nothing & tomato & carrots & carrots \\ 
    \end{tabular}} \\\hline
    \\[-10pt]
    \includegraphics[height=3cm, width=2.3cm]{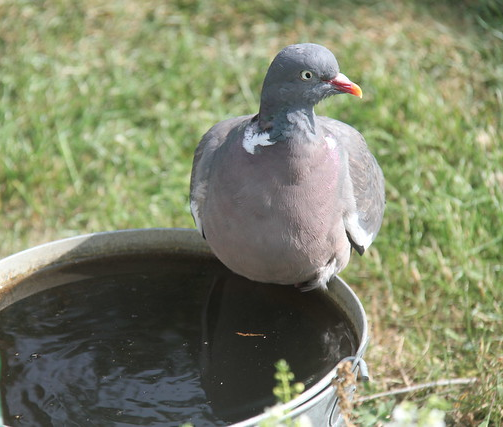} &    
    \resizebox{0.45\textwidth}{!}{
    \begin{tabular}[b]{lcccc}
    \multicolumn{5}{c}{\large{What type of bird is this?}}\\[5pt]  \hline
      Task & w/o PT & PT & PT+EWC & PT+ER \\ \hline
      Interior & dog & owl & owl & owl \\
      \textbf{Animals} & pigeon & pigeon & pigeon & pigeon \\
      Food & turkey & pigeon & pigeon & pigeon \\
      Transport & not sure & duck & pigeon & seagull \\ 
      Sports & zero & seagull & pigeon & seagull \\
    \end{tabular}} \\\hline
    \end{tabular}

    \caption{Examples of the evolution of predicted answers with different approaches combined with UNITER. Column Task shows the order of the training tasks. The bold task corresponds to the task of the sample.}
    \label{tab:examples}
  \end{table*}

\begin{table*}[]
    \centering
\resizebox{0.8\textwidth}{!}{%
    \begin{tabular}{lccccccc}
\hline
\multicolumn{2}{c}{\textbf{Reference}} & \multicolumn{3}{c}{\textbf{Compared Answer 1}} & \multicolumn{3}{c}{\textbf{Compared Answer 2}} \\
\textbf{Answer}	& \textbf{Acc}	& \textbf{Answer} &	\textbf{Acc} & \textbf{SBWT} & \textbf{Answer} & \textbf{Acc} & \textbf{SBWT} \\\hline
skateboarding & 1 & skateboard & 0 & -0.164 & black & 0 & -0.836 \\
snowboarding  & 1 & skiing & 0 & -0.134 & winter & 0 & -0.529 \\
breakfast & 1 & sandwich  & 0 & -0.340 & one & 0 & -0.855 \\
food & 1 & meat & 0 & -0.320 & toothbrush & 0 & -0.832 \\
skateboarding & 1 & skateboard  & 0.3 & -0.115 & skateboard & 0 & -0.164 \\
carrots & 1 & carrot & 0.3 & -0.093 & three & 0 & -0.818 \\
sheep & 1 & goat & 0.3 & -0.197 & white & 0 & -0.676 \\
cloudy & 1 & overcast & 0.3 & -0.151 & gray & 0 & -0.577 \\
black & 0 & black and white & 1 & 0.136 & brown & 1 & 0.269 \\
\hline
    \end{tabular}}
    \caption{Comparison of the SBWT metric of two answers with respect to the same reference answer. We verify that semantically more similar answers have higher SBWT.}
    \label{tab:sbwt_examples}
\end{table*}

Table~\ref{tab:sbwt_examples} presents examples of the SBWT metric. Specifically, it compares SBWT for two pairs of predicted answers with the same initial reference answer. When the initial prediction (reference answer) is correct, and both compared answers are wrong, we observe that SBWT penalizes similar answers less than unrelated ones (see the first four rows of  Table~\ref{tab:sbwt_examples}). 
Similarly, when one of the compared answers is partially correct (rows 5-8) according to the VQA accuracy metric, SBWT is less punishing compared to BWT, which in our examples would be $-0.7$.
Finally, the last row shows an example of corrected compared answers, where the accuracy improvement is weighted with the semantic distance of reference and compared answers.

\end{document}